\documentclass[journal]{IEEEtran}
\usepackage{graphicx}
\usepackage{hyperref}
\usepackage{amsmath,amssymb,amsfonts}
\usepackage{algorithm}
\usepackage{algpseudocode}
\usepackage{amsmath}
\usepackage{subfigure}
\usepackage{booktabs}
\usepackage{tabu}
\usepackage{multirow}
\usepackage{color}
\usepackage{booktabs}
\usepackage{threeparttable}
\ifCLASSINFOpdf
\else
\fi
\begin{document}
\title{SiamCorners: Siamese Corner Networks for Visual Tracking}

\author{Kai Yang,
        Zhenyu He\textsuperscript{*},~\IEEEmembership{Senior Member, IEEE,}
        Wenjie Pei\textsuperscript{*}, 
        Zikun Zhou,
        Xin Li,
        Di Yuan
        and~Haijun Zhang
        \thanks{K. Yang, W. Pei, Z. Zhou, D. Yuan and H. Zhang are with the School of Computer Science and Technology, Harbin Institute of Technology, Shenzhen, 518055, China. 
        
        X. Li is with Peng Cheng Laboratory, Shenzhen, China.
        
        Z. He is with the School of Computer Science and Technology, Harbin Institute of Technology, Shenzhen, 518055, China, and also with the Shenzhen Institute of Artificial Intelligence and Robotics for Society, China, e-mail: (zhenyuhe@hit.edu.cn).
        
        \textsuperscript{*}Zhenyu He and Wenjie Pei are corresponding authors.
        
        }

        }

\maketitle

\begin{abstract}
The current Siamese network based on region proposal network (RPN) has attracted great attention in visual tracking due to its excellent accuracy and high efficiency. However, the design of the RPN involves the selection of the number, scale, and aspect ratios of anchor boxes, which will affect the applicability and convenience of the model. Furthermore, these anchor boxes require complicated calculations, such as calculating their intersection-over-union (IoU) with ground truth bounding boxes. Due to the problems related to anchor boxes, we propose a simple yet effective anchor-free tracker (named Siamese corner networks, SiamCorners), which is end-to-end trained offline on large-scale image pairs. Specifically, we introduce a modified corner pooling layer to convert the bounding box estimate of the target into a pair of corner predictions (the bottom-right and the top-left corners). By tracking a target as a pair of corners, we avoid the need to design the anchor boxes. This will make the entire tracking algorithm more flexible and simple than anchor-based trackers. In our network design, we further introduce a layer-wise feature aggregation strategy that enables the corner pooling module to predict multiple corners for a tracking target in deep networks. We then introduce a new penalty term that is used to select an optimal tracking box in these candidate corners. Finally, SiamCorners achieves experimental results that are comparable to the state-of-art tracker while maintaining a high running speed. In particular, SiamCorners achieves a 53.7\% AUC on NFS30 and a 61.4\% AUC on UAV123, while still running at 42 frames per second (FPS).
\end{abstract}

\begin{IEEEkeywords}
Visual tracking, Siamese network, top-left corner, bottom-right corner.
\end{IEEEkeywords}

% \fi
\IEEEpeerreviewmaketitle

\section{Introduction}
\IEEEPARstart{V}{isual} trackers based on Siamese networks \cite{li2019siamrpn++, wang2019fast, zhang2019deeper, siamban}  have achieved state-of-the-art results on various tracking benchmark datasets \cite{wu2015object, fan2019lasot, mueller2016benchmark, kiani2017need, muller2018trackingnet}. A key component of state-of-the-art trackers \cite{li2019siamrpn++, zhu2018distractor, li2018high} is based on anchor boxes, which are boxes of different sizes and aspect ratios that select as tracking candidates. Anchor boxes are widely used in visual tracking \cite{li2019siamrpn++, zhang2019deeper, zhu2018distractor, li2018high}, and can achieve a good trade-off in speed-accuracy. These methods \cite{li2019siamrpn++, zhang2019deeper, zhu2018distractor, li2018high} first generate anchor boxes densely over the image, 
they then find the candidate regions of the target by learning a strong classifier, and they finally refine their coordinates through a regression network.

\begin{figure}[!t]
\centering
\setlength{\abovecaptionskip}{0.cm}
\includegraphics[width=8.8 cm, scale=.5]{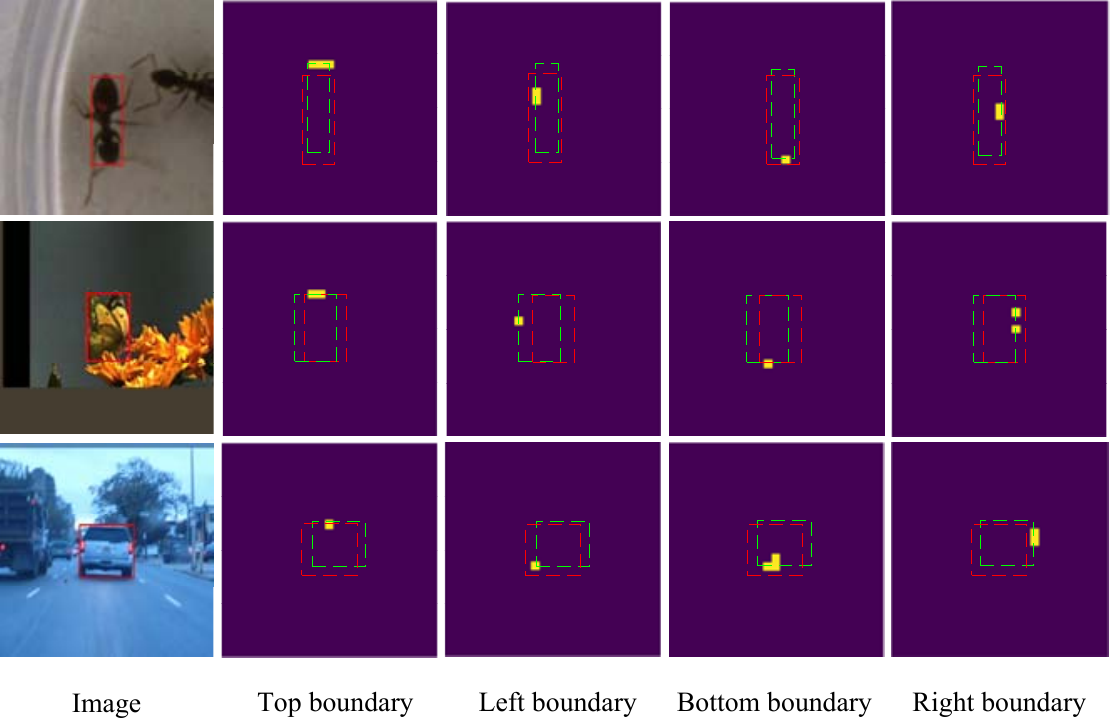}
\caption{Visualization results of feature maps (keep only the highest score for clarity) of the four boundaries (green box) predicted by Siamese network. The feature maps of the four boundaries are fed to a modified corner pooling layer to predict the corner heatmaps and offsets of the green box. Specifically, we combine the offset prediction and penalty strategies to make the green box close to the ground truth box (red box).}
\label{heatmap}
\end{figure}
\vspace{-0.1cm}

However, trackers based on anchor boxes have two drawbacks. First, these trackers need very large-scale anchor boxes, e.g. more than 2k in SiamRPN \cite{li2018high} and more than 3k in SiamRPN++ \cite{li2019siamrpn++}. It is because that the anchor-based tracker is trained to classify whether each anchor box contains a target region, and a large number of anchor boxes are required to ensure that they have sufficient overlapping regions with the ground truth boxes. In training, the positive and negative samples are defined by calculating the IoU between anchor boxes and ground truth boxes \cite{tian2019fcos}. They \cite{li2019siamrpn++, zhang2019deeper, zhu2018distractor, li2018high} simply think that the anchor boxes are positive samples if the IoU is larger than a hand-crafted threshold and are negative samples if the IoU is less than a hand-crafted threshold. However, recent research \cite{zhang2020bridging} indicates that this fixed hand-craft threshold will affect the performance of anchor-based algorithms. As a result, a tiny part of the anchor boxes can ensure overlap with ground truth boxes. In this design, this will produce a huge imbalance between positive samples and negative samples \cite{li2019siamrpn++} and slows down training \cite{law2018cornernet, lin2017focal}. Furthermore, the imbalance of positive and negative samples will have a huge impact on a model's performance \cite{zhang2020bridging, paa-eccv2020}. For example, the training for the classification model is inefficient as most locations are easy negative samples, which will lead to a degraded classification model \cite{lin2017focal}. If the classifier cannot produce a correct candidate region for the target location, then the regression network will not be accurate \cite{2017Faster, Ocean_2020}.

Second, trackers with hand-crafted anchor boxes will introduce many design options. This includes how to select the number, size, and aspect ratio of anchor boxes. This design is mostly inspired by ad-hoc heuristics and it can become more complex when a network predicts multiple-level feature maps, each of which uses different features and has specific anchor boxes \cite{li2019siamrpn++,law2018cornernet, lin2017focal,liu2016ssd}. These parameters in anchor-based trackers need to be tuned carefully. Even with careful design, it is difficult for trackers to deal with large changes of target shapes, especially for small targets, due to the fixed scales and aspect ratios of the anchor boxes.

In contrast, the location and boundary of the target that humans can identify are not obtained by predefined candidate boxes \cite{siamban}. Therefore, we can design a tracking algorithm without any predefined anchor boxes. Motivated by these analyses, we introduce a SiamCorners method for visual tracking, which makes our tracking algorithm more flexible and general by discarding anchor boxes. Our proposed method consists of four steps. First, we crop four boundary images containing target boundary information from an initial frame as the template images of the SiamCorners.  Specifically, the four boundary images regard the top, left, bottom and right boundary of the bounding box of the initial frame as the initial locations, respectively, and it then crops at a specific ratio of the width or height of the box. The SiamCorners then predicts the feature maps (see figure \ref{heatmap}) of the four boundaries of the bounding box based on the input of template images and search images. Second, inspired by the recently proposed CornerNet \cite{law2018cornernet, law2019cornernet}, which is used to predict a bounding box as a pair of corners in object detection. However, the original CornerNet is class-specific, and hence cannot be used directly for visual tracking. Consequently, we introduce a modified corner pooling layer to help the network integrate the target-specific feature in the Siamese framework. The entire corner pooling layer is used to predict the multiple top-left corners and bottom-right corners of the bounding box by the predicted feature maps of the four boundaries. Third, a decoding process is performed to predict the corner locations based on the heatmaps and offset maps. It is worth noting that we chose a Non-maximum-suppression (NMS) operation to remove some abnormal corners. Finally, multi-level features fusion and some penalty strategies are adopted to select the best tracking bounding box. Intuitively, features from earlier convolution outputs of deep networks will contain rich low-level feature information such as shape, color, which is important for tracking localization. Features from the latter convolution outputs of the deep networks will contain rich semantic information that is beneficial to deal with some challenging factors like huge deformation and background clutter. In our design, multi-layer features are fed into the multiple SiamCorners modules individually. We then use a new penalty function to select the best tracking box from the candidate corners generated by the layer-wise aggregation strategy. Specifically, we first employ the location and scale penalty function that considers the changes of width, height, and center of the tracking target compared to the previous frame. This penalty strategy is used to re-rank the corners' score to remove the outliers. We then utilize a penalty term to suppress large changes in the ratio and size of the corners. Subsequently, we select the corners with the largest score as a candidate tracking box. Finally, a linear interpolation is employed to make the tracking box changing smoothly. This design simplifies the output of the whole network and eliminates the complex design of the anchor boxes.

In summary, the proposed SiamCorners mainly includes the following contributions:
	
	\begin{itemize}
	    \item We propose an anchor-free proposal network (SiamCorners) that avoids many hand-crafted configurations during designing the anchor boxes and makes our method more flexible and general.
        \item We present a modified corner pooling layer to integrate target-specific information into the corner prediction.
	    \item We formulate a multi-layer features fusion strategy to help SiamCorners benefit from both low-level information and high-level information.
	    \item We introduce a novel penalty term that considers the changes of the center, width, and height of the target in the adjacent frames, which makes SiamCorners more suitable for tracking tasks.
	
	\end{itemize}

%\begin{figure*}[!t]
%\centering
%\centerline{{\includegraphics[width=5.7in]{pipeline.pdf}}}
%\caption{Overview of our SiamCorners in visual tracking. We transform visual tracking into a pair of bounding box corners detection. In the figure, the SiamCorners network predicts two heatmaps, one for top-left corner, and one for bottom-right corner.
%}
%\label{pipeline}
%\end{figure*}

Based on these contributions, we demonstrate the effectiveness and efficiency of SiamCorners on five tracking benchmark datasets: OTB100 \cite{wu2015object}, UAV123 \cite{mueller2016benchmark}, LaSOT \cite{fan2019lasot}, NFS30 \cite{kiani2017need} and TrackingNet \cite{muller2018trackingnet}.  Our SiamCorners approach achieves state-of-the-art results on the five datasets. Moreover, the wide range of ablation experiments are performed on the UAV123 and the LaSOT datasets.

\section{RELATED WORKS}
Generic object tracking has made great progress in recent years following the development of several new methods \cite{li2019siamrpn++, wang2019fast, zhang2019deeper, siamban, ruan2018multi, liu2017robust, Yuankai2019Learning, liu2020multi-task}. In particular, methods based on Siamese networks play an important role in visual tracking.

\textbf{Siamese Networks:} Recently, trackers based on Siamese networks have received much attention because of their well-balanced in accuracy-efficiency. SiameseFC \cite{bertinetto2016fully} calculates the similarity prediction of template images and search images through a full-convolutional network, while running over 100FPS. CFNet \cite{valmadre2017end} introduces correlation filter layers into the Siamese tracking framework, which can achieve end-to-end representation learning of tracking targets. However, its performance is not significantly improved compared to SiameseFC. SiamRPN \cite{li2018high} develops the tracking task as a one-shot detection task by introducing region proposal subnetwork in the Siamese network, which is trained offline in an end-to-end manner on large-scale image pairs. The region proposal network contains two branches, one for classification branch and  another for regression branch. The classification branch predicts the target location based on the fixed anchor boxes, while the regression branch fine-tunes the predicted target coordinates. Therefore, the design of anchor boxes has a great impact on the final tracking results. DaSiamRPN \cite{zhu2018distractor} further improves the discriminative ability of the model by introducing an effective sampling strategy to control the imbalanced distribution of the training data. However, note that these trackers \cite{bertinetto2016fully, valmadre2017end, li2018high, zhu2018distractor} built their algorithm in shallow 
networks similar to AlexNet \cite{krizhevsky2012imagenet}, the components they proposed did not benefit from low-level and high-level feature information due to relatively shallow networks. While our method uses multiple SiamCorners modules to predict multiple candidate corners according to 
the layer-wise aggregation features of the deep networks. Similarly, SiamRPN++ \cite{li2019siamrpn++} enables SiamRPN tracker to take advantage of deep features by introducing deeper networks, such as ResNet-50 \cite{he2016deep} or deeper networks. Unlike anchor-based trackers \cite{li2019siamrpn++, zhu2018distractor, li2018high}, our SiamCorners tracker is more flexible and general since it eliminates the design of anchor boxes.
\begin{figure*}[!t]
\centering
\centerline{{\includegraphics[width=7.0in]{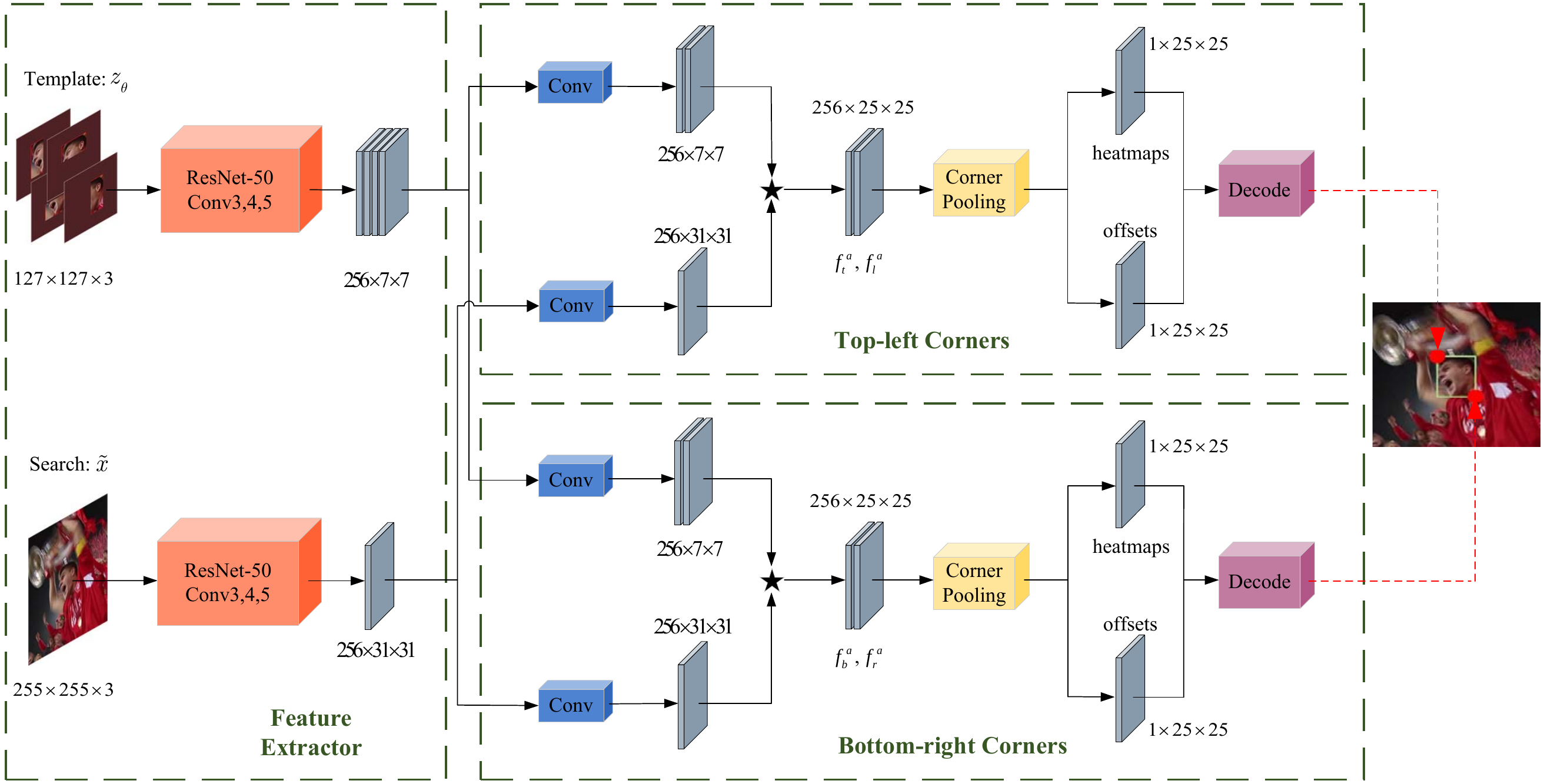}}}
\caption{The overview of our SiamCorners architecture, which includes the Siamese feature extractor followed by the top-left corner and bottom-right corner branches in parallel. $\bigstar$ denotes depth-wise correlation operation. $z_{\theta}(\theta\in \{t,l,b,r\})$ denote the four boundary images cropped from the original image.
}
\label{sys}
\end{figure*}

\textbf{Anchor-free Mechanism:} Anchor-free trackers recently became popular in visual tracking tasks due to their relatively simple and superior performance. FCAF \cite{han2019fully} and Ocean \cite{Ocean_2020} respectively introduce a simple anchor-free tracker by using a variant of the FCOS \cite{tian2019fcos} detector.
They \cite{Ocean_2020, han2019fully} first generate some hand-crafted points instead of a large number of anchor boxes as training samples, and then use a classification network to find some points that may be the locations of a tracking target. Finally, they utilize a regression network to predict a 4D vector, which represents the distance from the target location to the four sides of the bounding box. Unlike these anchor-free trackers, our SiamCorners directly predicts the target box as a pair of corners, which does not include a classification network to determine the coarse location in some hand-crafted points. Therefore, SiamCorners maintains a high-efficiency running speed by eliminating the classification phase of the anchor-free trackers \cite{Ocean_2020, han2019fully}.

Recently, corner-based detectors have attracted widespread attention in the object detection community due to their relatively new paradigms. CornerNet \cite{law2018cornernet} was the first to propose a corner-based detector that converts the bounding box of the target into a pair of corner predictions. CornerNet uses an hourglass network \cite{newell2016stacked} to obtain the heatmaps including the top-left corners and the bottom-right corners of the bounding boxes. It then uses a series of feature embeddings to match these corners. To accurately predict the target, CornerNet introduces a corner pooling to localize the corner. CornerNet-Lite \cite{law2019cornernet} utilizes CornerNet as a baseline to introduce an attention mechanism so that the detector does not require detailed processing for each pixel in the input image, thereby improving the inference efficiency of the entire detector. CenterNet \cite{duan2019centernet:} regards each detection target as a triplet, which predicts the centers of the bounding boxes along with corners. MartixNets \cite{rashwan2020matrixnets} proposes an aspect ratio and scale aware architecture so that CornerNet can eliminate corner pooling operation. Instead of matching corners by feature embeddings in CornerNet, CentripetalNet \cite{CentripetalNet} matches the predicted corners by whether the centripetal shift results of the corners are aligned, where the centripetal shift denotes the spatial shift from the corner locations to the center of the bounding box.

There are two key differences between tracking and detection tasks. First, anchor-free detectors usually aim to detect different categories of targets in an input image, while in tracking, we aim to match specific targets in search images. Therefore, we follow a common Siamese architecture \cite{li2019siamrpn++} that can discriminate the target from the search images. Second, the detector based on anchor-free usually assumes that the target category is predefined. The common dataset MS COCO \cite{lin2014microsoft} of object detection only contains 91 target categories. Thus, independent corner pooling layers are trained in \cite{law2018cornernet, law2019cornernet, duan2019centernet:, rashwan2020matrixnets, CentripetalNet} for each object class. In contrast, the target class is unknown in tracking \cite{li2019siamrpn++, he2017robust, he2016connected}. Furthermore, unlike the object detection tasks, the target does not belong to any pre-defined classes or is represented by existing training datasets. Instead, target-specific corner pooling layers are required for test frames incorporating target information in Siamese framework.

\section{Proposed Method}
\subsection{Overview}

\begin{figure}[!t]
\setlength{\abovecaptionskip}{0.cm}
\setlength{\belowcaptionskip}{-0.cm}
\centering
\makeatletter\def\@captype{figure}\makeatother
{
   \label{crop}
   \includegraphics[width=8.1 cm]{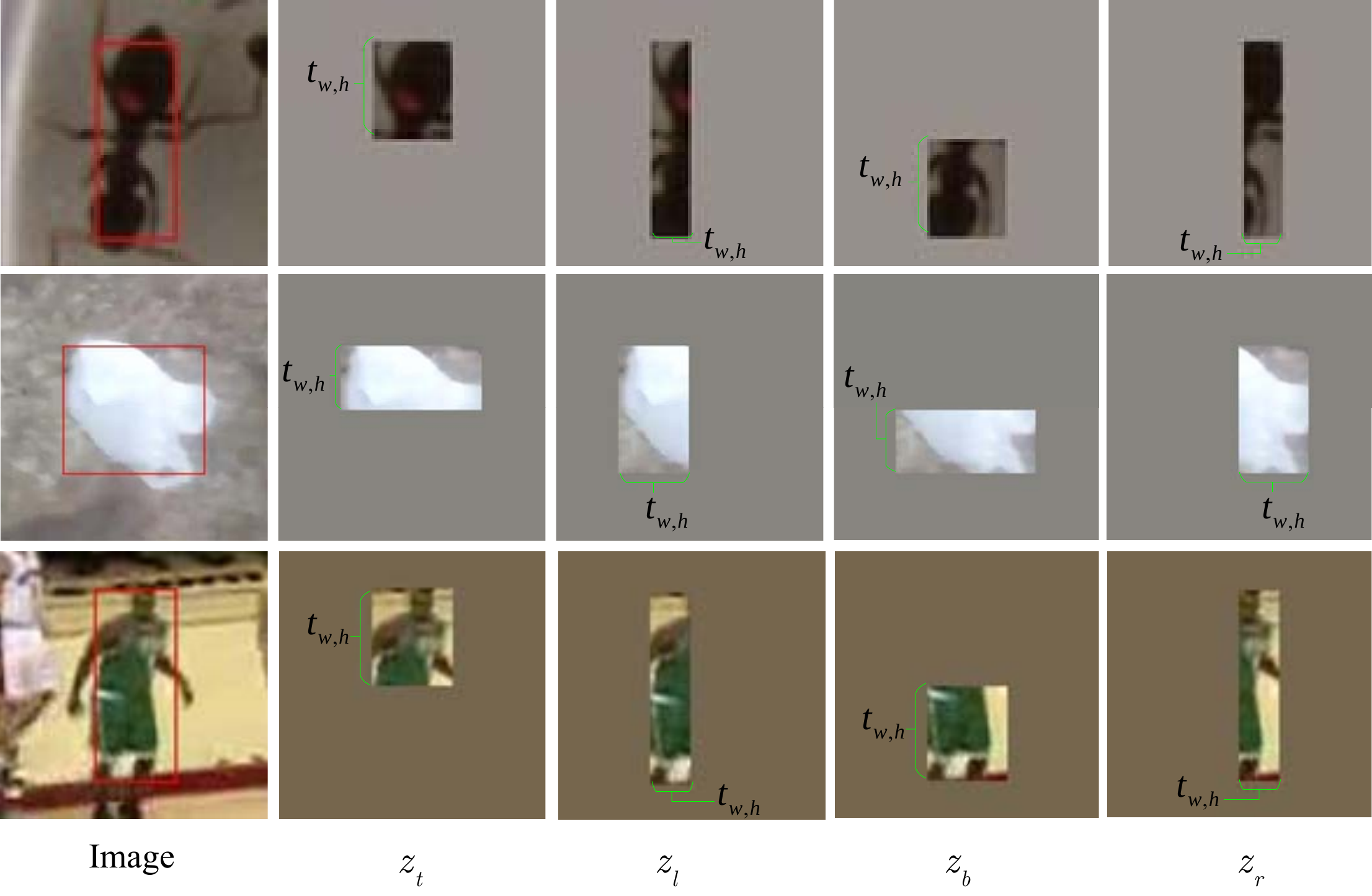}}

\caption{Examples of four boundary images fed to the template branch. The boundary images are cropped from the $t_{w,h}$ width or height of the red box of the input image.}
\label{crop}
\end{figure}

In this work, we propose an anchor-free method, SiamCorners, which detects a target as a pair of corners. A common Siamese network is used to integrate target-specific information into the tracking architecture.

Fig. \ref{sys} provides an overview of SiamCorners. As illustrated in Fig. \ref{sys}, the template branch has four inputs: i) the top-boundary image $z_{t}$, ii) the left-boundary image $z_{l}$, iii) the bottom-boundary image $z_{b}$, iv) the right-boundary image $z_{r}$ (see Fig. \ref{crop}). The target images of the top (left) boundary and the bottom (right) boundary respectively take the four boundaries of the target's bounding box as the initial location, and are cropped from $t_{w,h}$ (we set $t_{w,h}=0.5$ in all experiments) times the width (height) of the bounding box of the original image, respectively. Furthermore, the other  regions of the boundary images are constant padding. Then, we use the same data augmentation strategy as SiamRPN++ \cite{li2019siamrpn++} to resize the four boundary images to a fixed size. Afterwards, the four boundary images of the bounding box are fed to ResNet-50 \cite{he2016deep} network, respectively. The feature maps from different layers, namely $conv3$, $conv4$, and $conv5$ blocks of ResNet-50, are fed to the two branches. Specifically, the feature maps of the top and the left boundary images are fed to the top-left corner prediction branch, and the feature maps of the bottom and the right boundary images are fed to the bottom-right corner prediction branch. The features of different layers in ResNet-50 are followed by a convolution network. Depth-wise correlation operations are further performed to obtain correlation features. The correlation features are then fed to the corner pooling layers to get the corner heatmaps and offsets, respectively (see Section \ref{pooling}). After obtaining the predicted heatmaps and offsets, a simple decoding process is used to obtain the corner candidates (see Section \ref{decoding1}). Finally, the multi-layer feature fusion and the post-processing operation are used to get the final bounding box (see Section \ref{fusion}).

\subsection{Tracking Corners Loss}
We predict two sets of corner heatmaps that are used to calculate the top-left corners and the bottom-right corners of the target, respectively. Let $\tilde{x}\in\mathbb{R}^{\emph{H}\times\emph{W}}$ be a search image, where $\emph{H}$ and $\emph{W}$ are the height and width of the image, respectively. The corner heatmaps of the convolutional network are denoted as $\hat{X}\in \left [0,1\right]^{\frac{H}{s}\times\frac{W}{s}}$, where $\emph{s}$ is a scale factor mapped from the search image to the corner heatmaps.

We train the corner network following Law and Deng \cite{law2019cornernet}. For each heatmap, there is only one ground-truth label for a corner as a positive location and the other locations are negative, which will make the network difficult to learn. For efficient training, we reduce the penalty when the negative locations are still within the radius of the positive location. We determine the radius by the at least \emph{d} IoU (we set \emph{d} = 0.5 in all experiments) between a pair of corners and the ground-truth box. For each ground truth corner $c(c_{x},c_{y})\in\mathbb{R}^{2}$ in the search image, we compute the location of corner heatmap ${\tilde{c}(\tilde{c}_{x},\tilde{c}_{y})=\left \lfloor \frac{c(c_{x},c_{y})}{s} \right \rfloor}$. Given the radius and location ${\tilde{c}}$, the amount of penalty reduction is calculated by 2D Gaussian, $X_{xy}=exp\left(-\frac{(x-\tilde{c}_{x})^2+(y-\tilde{c}_{y})^2}{2\delta_{\tilde{c}}^2} \right)$, where $\delta_{\tilde{c}}$ is 1/3 of the radius. Again, $X_{xy}$ denotes the ``ground truth" heatmap designed with unnormalized Gaussians. Let $\hat{X}_{xy}$ be the prediction score of network at the location $(x,y)$. The training objective is a variant of focal loss \cite{lin2017focal}:
\begin{equation}
\mathcal{L}_{tr} =\frac{-1}{K} \sum_{xy}
\begin{cases}
(1-\hat{X}_{xy})^\alpha log(\hat{X}_{xy})& \mbox{if}X_{xy}=1 \\ \label{1}
(1-X_{xy})^\beta(\hat{X}_{xy})^\alpha log(1-\hat{X}_{xy}) & \mbox{otherwise}
\end{cases}
\end{equation}
where $\alpha$ and $\beta$ denote the weighting factors which control the contribution of each corner (we set $\alpha=2$ and $\beta=4$ in all experiments). $K$ is the number of candidate corners. In eq.(\ref{1}), the $(1-X_{xy})$ term reduces the penalty when the predicted corners are within the radius of the positive location.

However, the output size of the network is usually smaller than the original image as the network becomes deeper. There will be a discretization error when mapping a location $(m,n)$ of the original image to the location $({\left\lfloor\frac{m}{s}\right\rfloor,\left\lfloor\frac{n}{s} \right \rfloor})$ in the heatmaps. Moreover, some precision will be lost in tracking when remapping locations from heatmaps to the original image. To address the discretization error caused by network stride, we introduce an offset network that is used to slightly adjust the corner offset before remapping the heatmap to the original image. The corner offset on the heatmap is calculated as:

\begin{equation}
\emph{\textbf{o$_k$}}=\left(\frac{m_{k}}{s}-\left \lfloor \frac{m_{k}}{s}\right \rfloor,\frac{n_{k}}{s}-\left \lfloor \frac{n_{k}}{s}\right \rfloor \right)
\label{o}
\end{equation}
where $\emph{\textbf{o$_k$}}$ is the corner offset, $m_{k}$ and $n_{k}$ are the $k-th$ corner coordinates of the object in the original image. We employ the smooth L1 Loss \cite{girshick2015fast} at corner locations:
\begin{equation}
\mathcal{L}_{off} =\frac{1}{K}\sum_{k=1}^KSmoothL1Loss(\emph{\textbf{o$_k$}},\hat{\emph{\textbf{o}}}_{k})
\label{off}
\end{equation}
where $\emph{\textbf{o}}_{k}$ and $\hat{\emph{\textbf{o}}}_{k}$ are ``ground truth" offset and offset network prediction, respectively. Therefore, the whole method optimizes the loss function $\mathcal{L}$, which is expressed as:

\begin{equation}
\mathcal{L} = \mathcal{L}_{tr}+\lambda\mathcal{L}_{off} \label{L}
\end{equation}
where $\lambda$ (we set $\lambda=1$ in all experiments)  is a hyper-parameter that controls the balance of the two loss terms.

\subsection{Corner Pooling}
\label{pooling}

\begin{figure*}[!t]
\centering
\centerline{{\includegraphics[width=6.5 in]{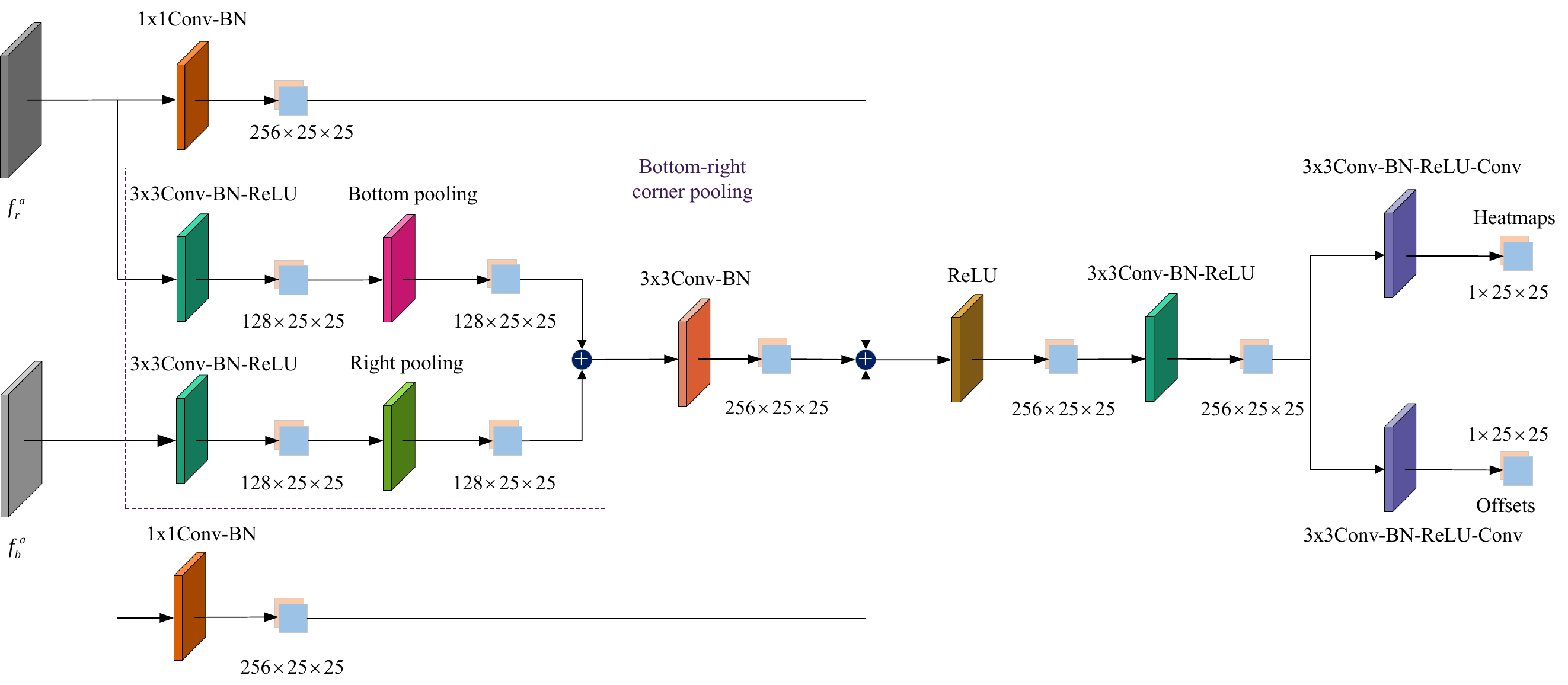}}}
\caption{The corner pooling module starts with the two feature maps, which are modified from the residual block. The bottom pooling and right pooling use the max-pools methods of eq.\ref{5} and \ref{6}, respectively. The entire corner pooling layer has two branches that predict the target's heatmaps and offsets, respectively. In the figure, the $3\times3$ Conv-BN-Relu module consists of a $3\times3$ convolution layer, a Batch Normalization layer, and a ReLU layer. Similarly, the $3\times3$ Conv-BN-ReLU-Conv module consists of the Conv-BN-ReLU module and a $1\times1$ convolution layer.}
\label{poolingmodule}
\end{figure*}

In this section, we describe how to use a modified corner pooling to predict corner heatmaps and offsets in Siamese architecture. The existing corner pooling is originally proposed for object detection, we make modifications to adapt it to object tracking. Given the template images $z_{\theta},\theta \in \{t,l,b,r\}$ and the search image $\tilde{x}$, the feature maps are efficiently denoted by parameterizing ${f^{a}(z_{\theta})={\phi(z_{\theta})}}, a\in \{3,4,5\}$ and ${f^{a}(\tilde{x})={\phi(\tilde{x})}}$, where $\phi$ is a convolutional neural network (CNN). Here, $(f^{3}(z_{\theta}),f^{3}(\tilde{x}))$, $(f^{4}(z_{\theta}),f^{4}(\tilde{x}))$, and $(f^{5}(z_{\theta}),f^{5}(\tilde{x}))$ denote the output features of \emph{conv3}, \emph{conv4}, and \emph{conv5} branches in Fig. \ref{sys}, respectively. Afterward, following the Siamese architecture, we aim to predict the depth-wise correlation feature: ${{f_{\theta}^{a}}={f^{a}(z_{\theta})\bigstar f^{a}(\tilde{x})}}$. Therefore, the specific-target features are integrated into the SiamCorners network.

Here, we will describe in detail how to determine whether a pixel $(i,j)$ is a bottom-right corner in correlation maps. On a $\frac{H}{s}\times\frac{W}{s}$ correlation map, the corner pooling layer max-pools the all feature on $f_{b}^{a}$ to get a feature vector $b_{ij}^{a}$ from $(i,0)$ to $(i,\frac{W}{s})$ , and max-pools the feature map $f_{r}^{a}$ to get a feature vector $r_{ij}^{a}$ from $(0,j)$ to $(\frac{H}{s},j)$. This calculation stage can be expressed as:

\begin{equation}
{b_{ij}^{a}} =
\begin{cases}

max(f_{b_{ij}}^{a}, b_{i(j-1)}^{a}) &\mbox   {if } 0<j\le\frac{W}{s} \\ \label{5}
f_{b_{i0}}^{a}  & \mbox{otherwise}
\end{cases}
\end{equation}

\begin{equation}
{r_{ij}^{a}} =
\begin{cases}
max(f_{r_{ij}}^{a}, r_{(i-1)j}^{a}) &\mbox   {if } 0<i\le\frac{H}{s} \\ \label{6}
f_{r_{0j}}^{a}  & \mbox{otherwise}
\end{cases}
\end{equation}

We then add ${b_{ij}^{a}}$ and ${r_{ij}^{a}}$ to get the final corner pooling result. Similarly, the pooling results ${t_{ij}^{a}}$ and ${l_{ij}^{a}}$ are obtained by max-pooling the feature vectors horizontally from right to left and vertically from bottom to top on the correlation feature maps $f_{t_{ij}}^{a}$ and $f_{l_{ij}}^{a}$, respectively. Finally, the result of top-left corner pooling is obtained by the sum of the two pooling results.
%\begin{figure*}[!t]
%\centering
%\centerline{{\includegraphics[width=5.0in]{CornerPooling.pdf}}}
%\caption{An example of the bottom right pool layer starts with a depth-wise correlation feature. We max-pools the feature vectors horizontally from left to right and vertically from top to bottom on the correlation feature maps, and then add the two pooled results.}
%\label{cor}
%\end{figure*}

%\begin{equation}
%{t_{ij}^{a}} =
%\begin{cases}
%max(f_{t_{ij}}^{a}, t_{(i+1)j}^{a}) &\mbox   {if } i<\frac{H}{s} \\ \label{7}
%f_{t_{\frac{H}{s}j}}^{a}  & \mbox{otherwise}
%\end{cases}
%\end{equation}

%\begin{equation}
%{l_{ij}^{a}} =
%\begin{cases}
%max(f_{l_{ij}}^{a},l_{i(j+1)}^{a}) &\mbox   {if } j<\frac{W}{s} \\ \label{8}
%f_{l_{i\frac{W}{s}}}^{a}  & \mbox{otherwise}
%\end{cases}
%\end{equation}

The architecture of the whole corner pooling is shown in Fig.\ref{poolingmodule}.  The starting part of the corner pooling module corresponding to the two inputs ($f_{b}^{a}$ and $f_{r}^{a}$) is a modified variant of the residual block \cite{he2016deep}, respectively. After the design of the residual block, the corner pooling feature is followed by a $3\times3$ Conv-BN layer and its output is added to the two projection shortcuts. Finally, it predicts the corner heatmaps and offsets by feeding the modified residual feature through a $3\times3$ Conv-BN-ReLU  layer followed by a $3\times3$ Conv-BN-ReLU-Conv layer, respectively. It is worth noting that the modified corner pooling and the original corner pooling in CornerNet \cite{law2018cornernet} have different designs in the network structure and pooling mechanism. As shown in Fig. \ref{poolingmodule}, the modified corner pooling introduces bottom pooling and right pooling in the right and bottom boundary feature maps ($f_{r}^{a}$ and $f_{b}^{a}$) respectively. Both the bottom and right boundary features are followed by projection shortcuts to enhance the corner prediction accuracy. In contrast, the original corner pooling \cite{law2018cornernet}  for object detection task predicts the bottom-right corner by using the bottom-right corner pooling and a single projection shortcut in a unified feature map. 

\subsection{Decoding for Corner Heatmaps and Offsets}
\label{decoding1}
In this section, we aim to predict the tracking bounding box by the given corner heatmaps and offsets. Following Section \ref{pooling}, the final outputs of corner pooling have four parts: i) the top-left corner heatmaps. ii) the bottom-right corner heatmaps. iii) the top-left corner offsets. iv) the bottom-right corner offsets. After corner pooling, the top-left and bottom-right corner heatmaps are first followed by a sigmoid function, respectively. Then, a NMS operation is adopted to remove redundant corner locations. We further obtain the corresponding top-left corner location $
(X{_{T_{l}}^{a}},Y{_{T_{l}}^{a}}) \in\mathbb{R}^{N\times 2}$ and bottom-right corner location $(X{_{B_{r}}^{a}},Y{_{B_{r}}^{a}}) \in\mathbb{R}^{N\times 2}$ by selecting the top $N$ (we set $N=15$ in all experiments) corner scores in top-left and bottom-right corner heatmaps, respctively. To overcome the impact of network stride, we denote $(\Delta X_{T_{l}}^{a}, \Delta Y_{T_{l}}^{a})$ and $(\Delta X_{B_{r}}^{a}, \Delta Y_{B_{r}}^{a})$ as the corresponding offsets of the top-left corners and the bottom-right corners, respectively. Afterward, the refined top $N$ corner sets $COR^{*}={(\hat X{_{T_{l}^{a}},\hat Y{_{T_{l}^{a}}},\hat X{_{B_{r}^{a}}},\hat Y{_{B_{r}^{a}}},S_{cor}^{a})}}$ can be predicted by the following formula:

%\begin{figure*}[!t]%\begin{figure*}[!t]

%\centering
%\centerline{{\includegraphics[width=7.2 in]{Decoding.pdf}}}
%\caption{An example of decoding for heatmap and offset. The values in the blue area of the figure represent potential corner locations or offsets. The #gathering feature block first uses the output of the heatmap to determine the location of the target and then predicts the offset of the corresponding location through offset network.}
%label{decoding}
%end{figure*}

\begin{equation}
\begin{split}
\hat X{_{T_{l}^{a}}}
=X{_{T_{l}}^{a}}+\Delta X_{T_{l}}^{a},\quad \hat Y{_{T_{l}^{a}}}
=Y{_{T_{l}}^{a}}+\Delta Y_{T_{l}}^{a}\\
\hat X{_{B_{r}^{a}}}
=X{_{B_{r}}^{a}}+\Delta X_{B_{r}}^{a},\quad \hat Y{_{B_{r}^{a}}}
=Y{_{B_{r}}^{a}}+\Delta Y_{B_{r}}^{a}\\
S_{cor}^{a}
=\frac{1}{2}(T_{l_{sco}}^{a}+B_{r_{sco}}^{a})
\end{split}
\end{equation}
where $S_{cor}^{a}$ is a pair of corner scores which are the average score of the top-left corner score $T_{l_{sco}}^{a}$ and bottom-right corner score $B_{r_{sco}}^{a}$. Here, $COR^{*}\in\mathbb{R}^{N\times 5}$. After the top $N$ corners $COR^{*}$ are obtained, we use multi-layer features fusion and some penalty strategies to select the optimal tracking corners.

\subsection{Multi-level Features Fusion and Corners Selection}
\label{fusion}
\textbf{Features Fusion:} Intuitively, visual tracking requires rich feature representations that include features from low-level to high-level, and resolutions from fine to coarse and scales from small to large \cite{li2019siamrpn++}. To this end, we introduce a multi-layer features fusion strategy to help SiamCorners benefit from low-level features and high-level features. Following Section \ref{decoding1}, we concatenate $\hat X{_{T_{l}^{a}}}$, $\hat Y{_{T_{l}^{a}}}$, $\hat X{_{B_{r}^{a}}}$ , $\hat Y{_{B_{r}^{a}}}$ and $S_{cor}^{a}$ and use a simple mapping to get the corner locations $X_{T_{l}}$, $Y_{T_{l}}$,  $X_{B_{r}}$, $Y_{B_{r}}$ and $S_{cor}$ corresponding to the original image, respectively. Therefore, the corner sets in the original image can be denoted as a $3N\times5$ dimension term $COR={(X{_{T_{l}},Y{_{T_{l}}},X{_{B_{r}}},Y{_{B_{r}}},S_{cor})}}$.

\textbf{Corners Selection:} After obtaining the corners $COR$, we introduce a post-process strategy to make the SiamCorners more suitable for tracking tasks. In visual tracking, the target in nearby frames almost does not have large variations in size and shape. We calculate the overall scales $c_{sc}$ of the corners by the following formula:

\begin{equation}
c_{sc}^2
=(w+p)(h+p) \label{14}
\end{equation}
where $w=X{_{B_{r}}}-X{_{T_{l}}}$ and $h=Y{_{B_{r}}}-Y{_{T_{l}}}$ denote the width and height of the target, and $p=\frac{1}{2}(w+h)$ denotes the padding. Similarly, the same principle is used to calculate the overall scales $\hat c_{sc}^{2}$ of corners at the last frame. We use a penalty term to suppress large changes in the size and ratio of the target:
\begin{equation}
C_{penalty}
=e^{\eta*max(\frac{c_{r}}{\hat c_{r}},\frac{\hat c_{r}}{c_{r}})*max(\frac{c_{sc}}{\hat c_{sc}},\frac{\hat c_{sc}}{c_{sc}})} \label{15}
\end{equation}
where $c_{r}$ and $\hat c_{r}$  denote the corners' ratio of height and width in the current frame and the last frame, respectively. Here, $\eta$ is a hyper-parameter. Therefore, the score $S_{pen}$ of each corner in the $COR$ is calculated by the weighted penalty as: $S_{pen}^u=C_{penalty}^u*S_{cor}^u$, $u=1,2,...,3N$. Unlike SiamRPN \cite{li2019siamrpn++,li2018high}, we further introduce a novel index term $U(u), u=1,2,...,3N$, which represents the index of the variation degree of each corner in $COR$ from large to small compared with the last frame. The metric of variation degree is determined by the sum of the absolute values of the difference between the center, width, and height of the target between two adjacent frames. For example, when $u=1$, $S_{pen}^{U(1)}$ means that the $U(1)-th$ corner in the corner sets $COR$ has the largest motion change relative to the last frame. Finally, the scores $S_{final}^{U(u)}$ of each corner can be expressed as:

\begin{equation}
S_{final}^{U(u)}
=S_{pen}^{U(u)}*(1-\gamma)+S_{han}^{u}*\gamma \label{16}
\end{equation}
where $\gamma$ is a hyper-parameter that controls the balance between the two score terms. $S_{han}^{u}$ is a variant of the Hanning window since we only take the monotonically increasing part of the Hanning window. After obtaining the final score $S_{final}$ of each corner, the expression of the corner set $COR$ is replaced by $COR_{final}={(X{_{T_{l}},Y{_{T_{l}}},X{_{B_{r}}},Y{_{B_{r}}},S_{final})}}$. Afterwards, we pick the corners with the highest score in the $COR_{final}$ as the final tracking bounding box. Furthermore, a linear interpolation operation is used to update the target size so that the shape can be changed smoothly.

Finally, we summarize the SiamCorners in algorithm {\ref{tracking_all}}.

\begin{algorithm}[!t]
  \caption{Tracking as one-shot}
  \label{tracking_all}
  \begin{algorithmic}[1]
    \Require
       initial template images: $z_{\theta}$;
       search image: $\tilde{x}$;
    \Ensure
     Tracking bounding box $x^{*}$
    \State ${f^{a}(z_{\theta})\gets{\phi(z_{\theta})}}$; \quad {\footnotesize \#  Feed the template images into CNN.}
    \Repeat
      \State ${f^{a}(\tilde{x})\gets{\phi(\tilde{x})}}$;
      \State ${{f_t^{a}},{f_l^{a},f_b^{a}},{f_r^{a}}\gets{f^{a}(z_{\theta})\bigstar f^{a}(\tilde{x})}}$ {\footnotesize \# Depth-wise correlation}
      \State $t^{a},l^{a},b^{a},r^{a}\gets{CornerPooling({{f_t^{a}},{f_l^{a},f_b^{a}},{f_r^{a}}})}$ {\footnotesize \# Section \ref{pooling}}
      \State $COR^{*}\gets {Decoding(t^{a},l^{a},b^{a},r^{a})}$  {\footnotesize \# Decoding process (Section \ref{decoding1})}
       \For{$a=3,4,5$}
            \State $COR\gets{{(\hat X{_{T_{l}^{a}},\hat Y{_{T_{l}^{a}}},\hat X{_{B_{r}^{a}}},\hat Y{_{B_{r}^{a}}},S_{cor}^{a})}}}$ {\footnotesize \# Concatenation}
        \EndFor
        \State $COR_{final}\gets{COR}$\quad {\footnotesize \# Update corner scores using eq.(\ref{14})-(\ref{16})}
        \State $x^{*}\gets{PostProcessing(COR_{final}})$ {\footnotesize {\# Selecting a pair of corners and using a linear interpolation operation} }

    \Until {end of sequence}
  \end{algorithmic}
\end{algorithm}

% or


\section{Experimental Results}
Our method is implemented in Python using PyTorch. On a single NVIDIA RTX 2080 GPU, SiamCorners achieves a tracking speed of 42 FPS by employing ResNet-50 as backbone. To facilitate further research, the completed training and testing code will be released at https://github.com/yangkai12/SiamCorners.
\subsection{Training Details}
 We use ResNet-50 pre-trained on ImageNet \cite{russakovsky2015imagenet} as the backbone network. The training splits of the GOT-10k \cite{huang2019got}, LaSOT \cite{fan2019lasot}, COCO \cite{lin2014microsoft}, YouTube-BoundingBoxes \cite{real2017youtube}, ImageNet VID \cite{russakovsky2015imagenet}, and ImageNet DET datasets are used to train our network to learn how to measure the similarity of input image pairs in visual tracking. In both training and testing phases, we set the sizes of the template image and test image to 127 pixels and 255 pixels, respectively. To reduce overfitting, we implemented some data augmentation strategies in the input image pairs, such as random horizontal flipping, random color jittering, random scaling, random cropping, adjusting the brightness and contrast of an image. We use stochastic gradient descent (SGD) with a batchsize of 58 to optimize the overall training loss from eq. (\ref{L}). We set a learning rate of 0.001 for earlier 5 epochs to train the top-left branch and bottom-right branch. For the latter 15 epochs, the whole network is trained in an end-to-end manner, where the learning rate decays exponentially from 0.005 to 0.0005, while weight decay and momentum are set to 0.0001 and 0.9, respectively. Unlike SiamRPN++ \cite{li2019siamrpn++}, we do not have a finetune backbone operation in the last 10 epochs, which will increase training time exponentially.

\subsection{Testing Metrics}

\begin{figure}[!t]
\centering
\makeatletter\def\@captype{figure}\makeatother
\subfigure{
   \label{fig_Success_Plot}
   \includegraphics[width=4.1 cm]{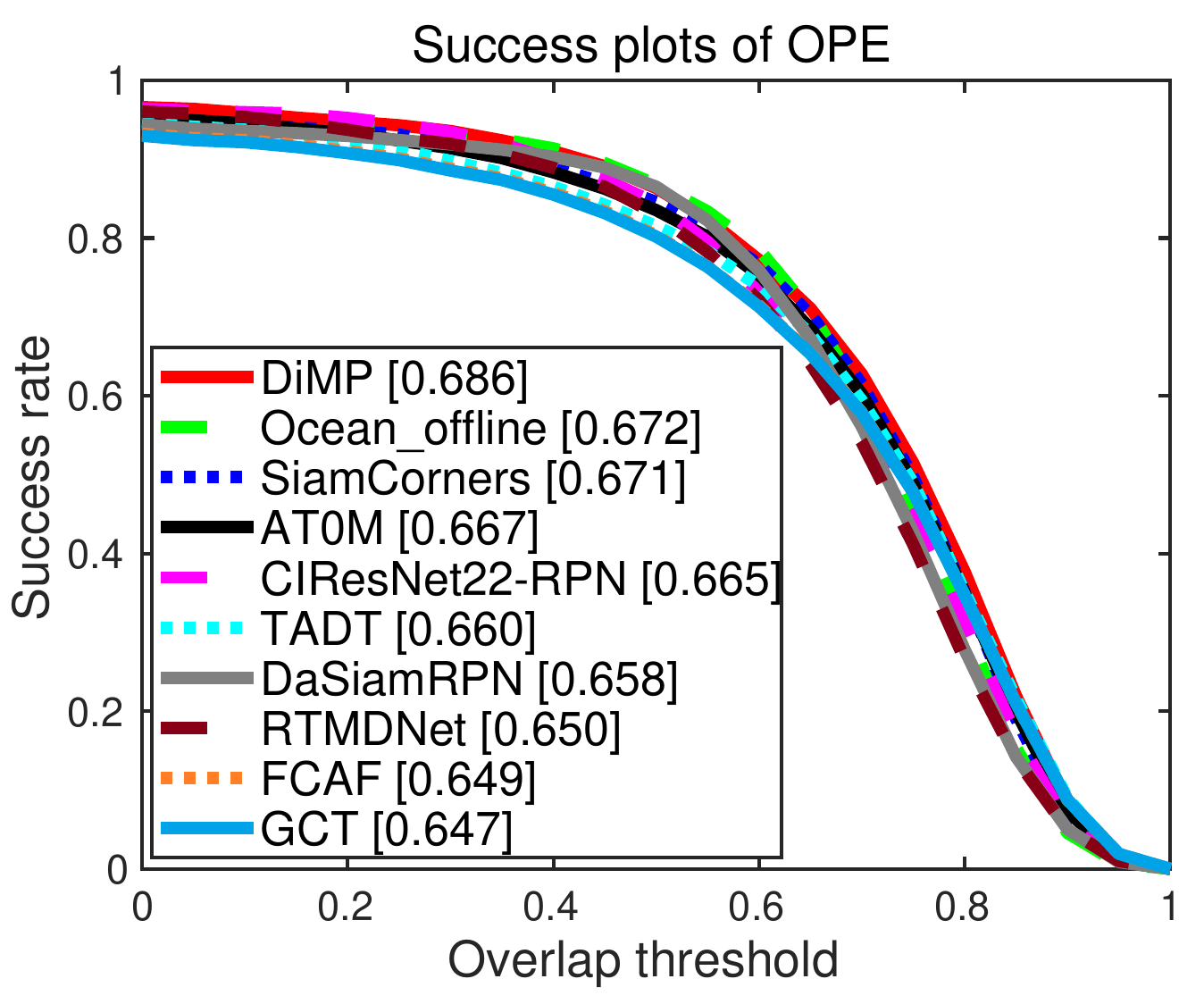}}
\hspace{0.005 in}
\subfigure{
   \label{fig_Precision_Plot}
   \includegraphics[width=4.1 cm]{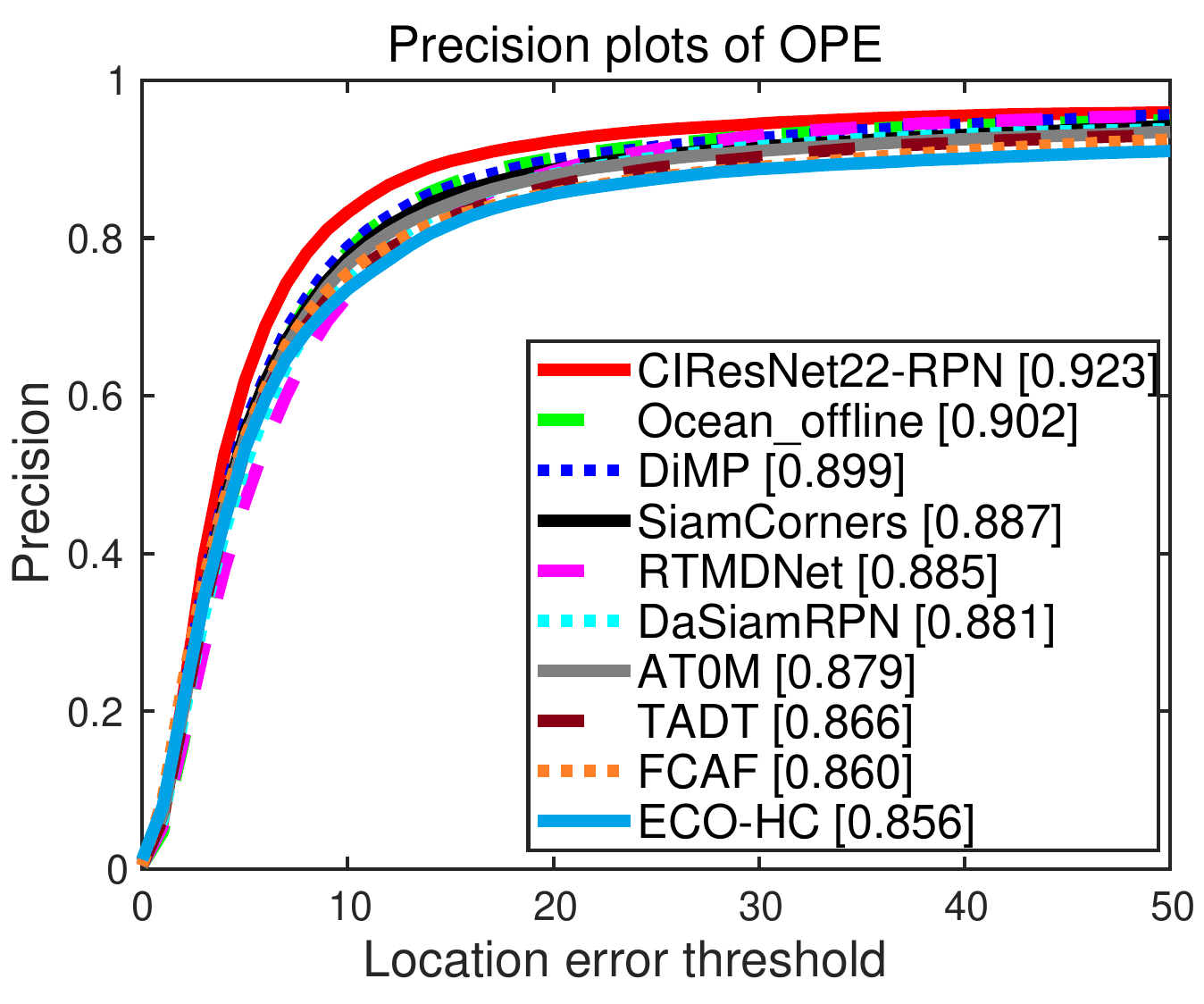}}
\caption{Experimental results of the methods on the OTB100 dataset.}
\label{OTB}
\end{figure}

In this experiment, we evaluate our method on tracking benchmarks OTB100 \cite{wu2015object}, UAV123 \cite{mueller2016benchmark}, LaSOT \cite{fan2019lasot}, NFS30 \cite{kiani2017need}, and TrackingNet \cite{muller2018trackingnet}. We use success scores and precision scores to evaluate the performance of the trackers on OTB100, UAV123 and NFS30. The success scores are defined as the percentage of success frames when the overlap threshold between the predicted result and the ground truth varies from 0 to 1, where a success frame means that the tracking result is larger than a given threshold. For a fair comparison, all trackers are ranked by the area under curve (AUC) of the success
plot. The precision score is computed as the percentage of frames, which is defined that the Euclidean distance between the center of the prediction target and the ground truth is within a given distance threshold. Here, the different distance thresholds are used to evaluate all trackers. For the LaSOT and TrackingNet datasets, we use the success score, precision score, and normalized precision score to evaluate the performance of all trackers, where the definitions of success and precision score are the same as previously described. Meanwhile, the normalized precision is defined as the normalization of precision over the size of the ground truth bounding box. The trackers are then ranked by AUC score with normalized precision between 0 and 0.5. Generally, the AUC score of success plot is mainly used to rank all trackers.

\subsection{State-of-the-art Comparison}

\begin{figure}[!t]
\centering
\makeatletter\def\@captype{figure}\makeatother
\subfigure{
   \label{fig_Success_Plot}
   \includegraphics[width=4.1 cm]{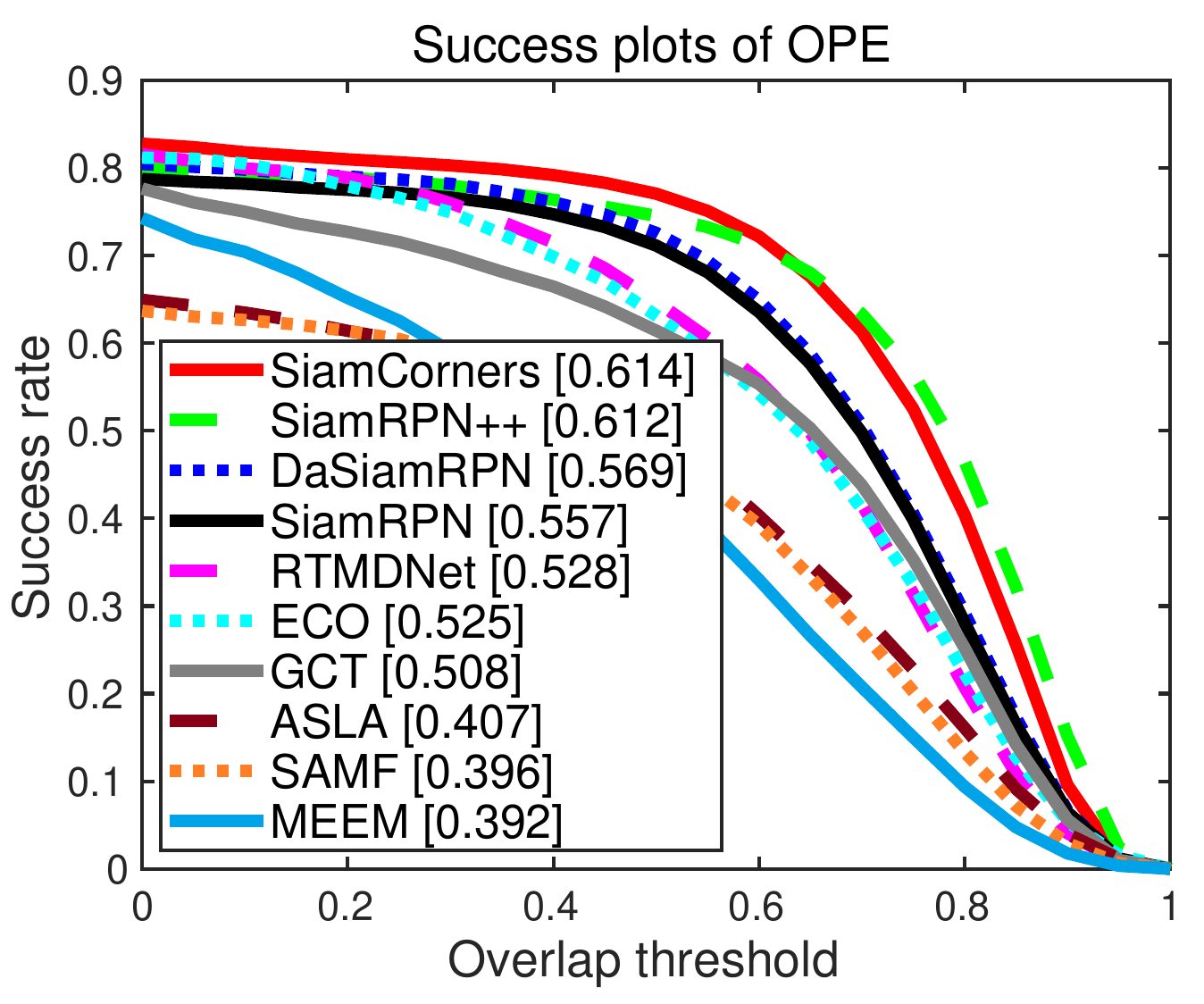}}
\hspace{0.005 in}
\subfigure{
   \label{fig_Precision_Plot}
   \includegraphics[width=4.1 cm]{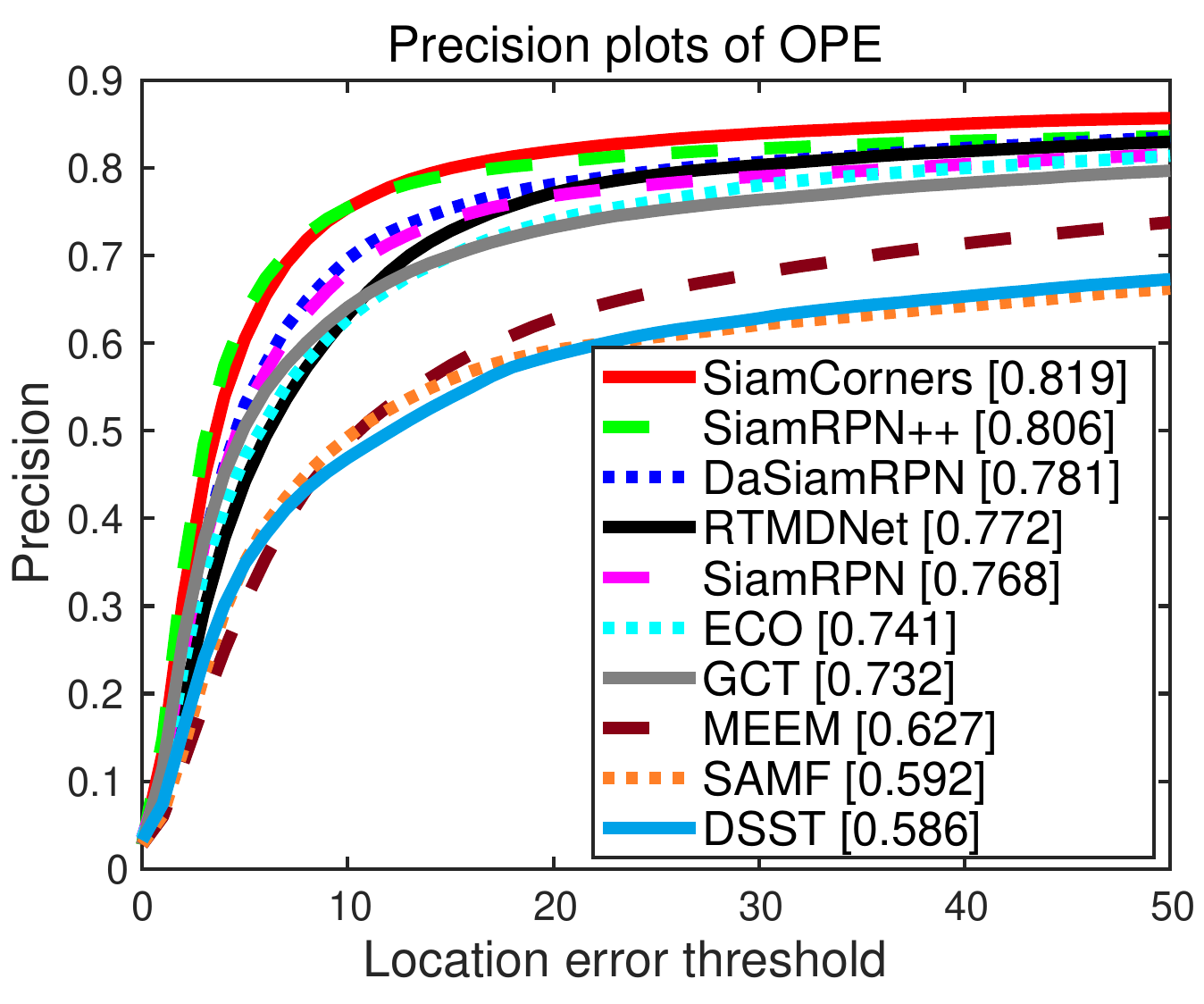}}
\caption{Experimental results of the methods on the UAV123 dataset.}
\label{UAV}
\end{figure}

\begin{figure*}[!t]
\centering
\subfigure[scale variation]{\includegraphics[width=0.245\textwidth]{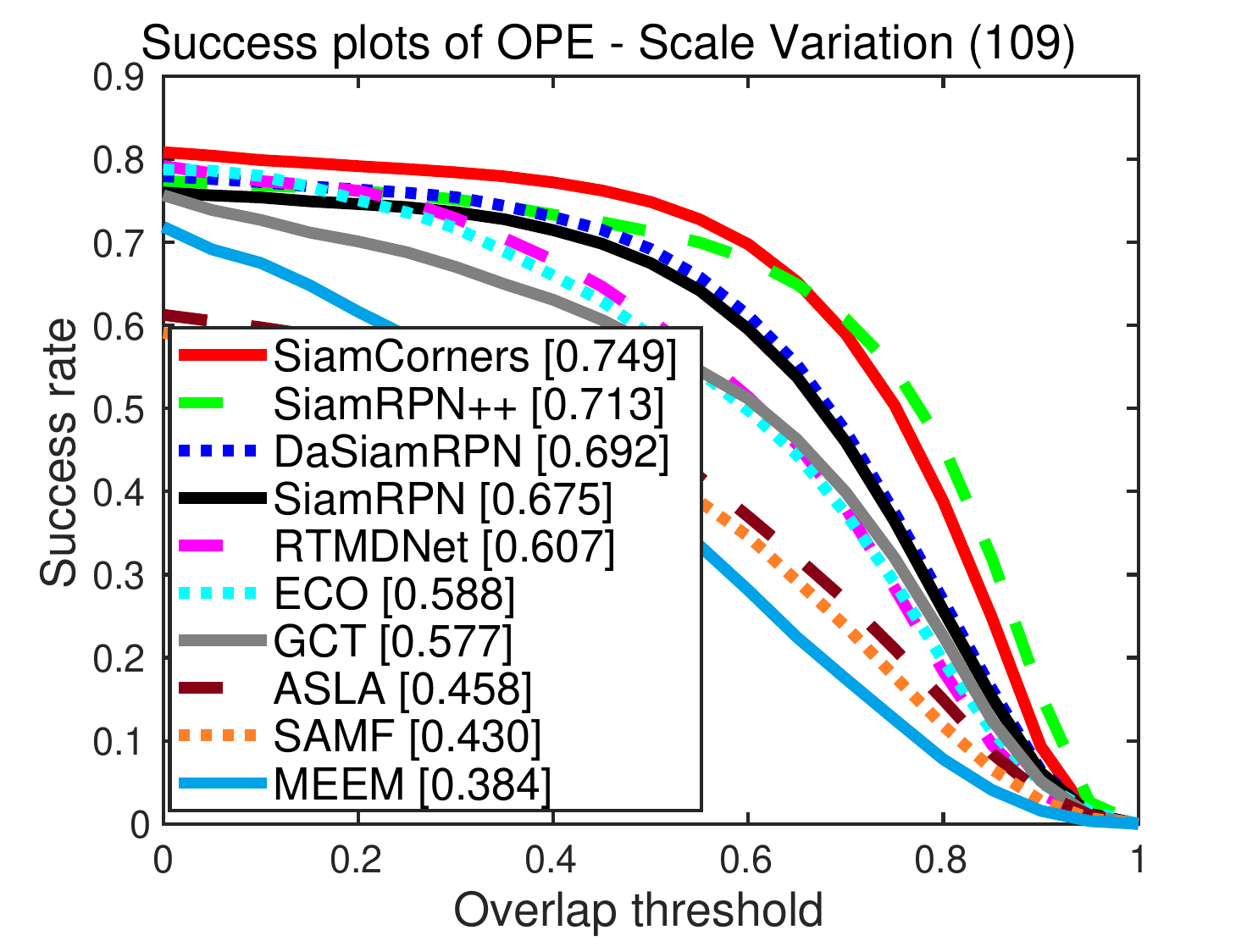}}
\subfigure[aspect ratio change]{\includegraphics[width=0.245\textwidth]{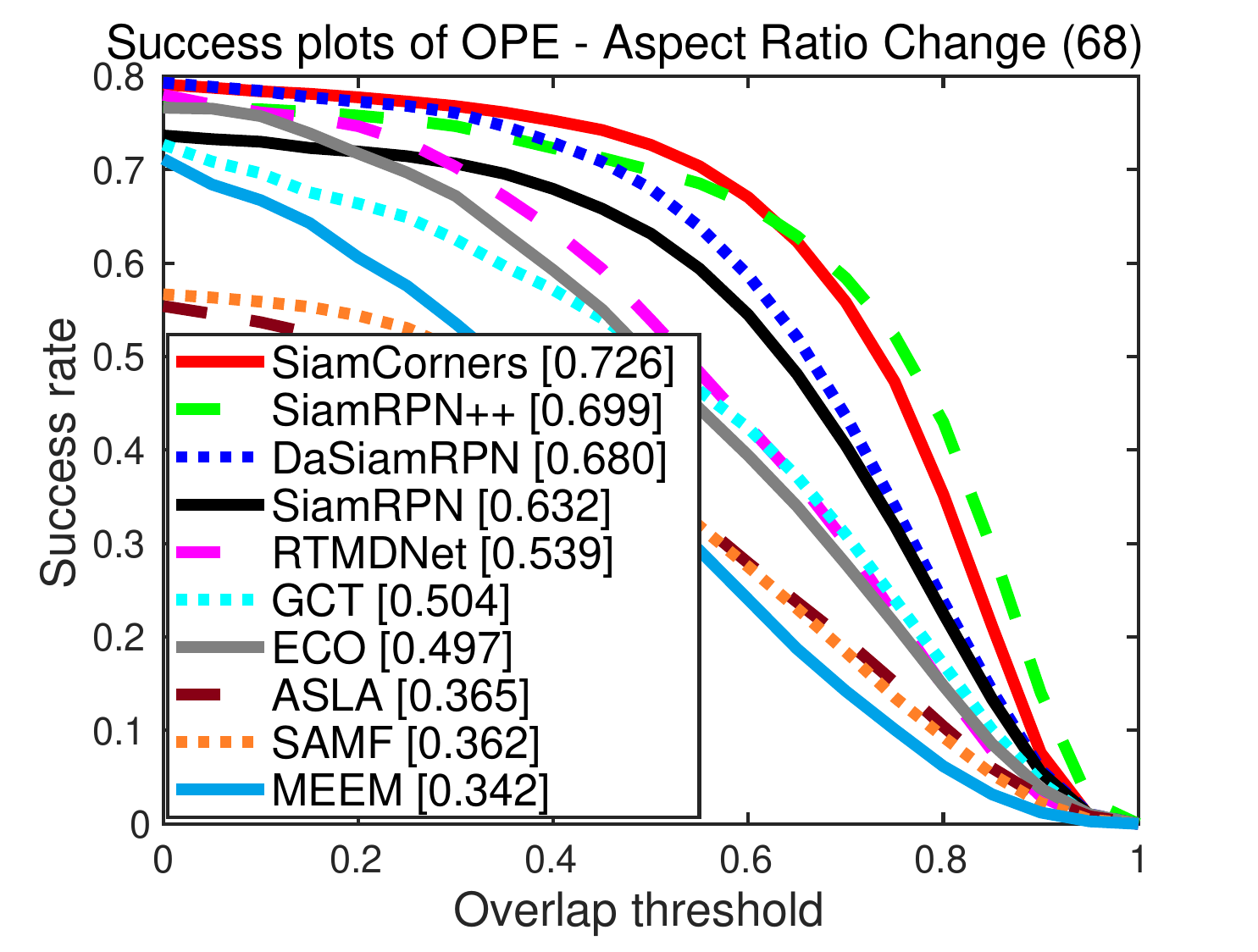}}
\subfigure[low resolution]{\includegraphics[width=0.245\textwidth]{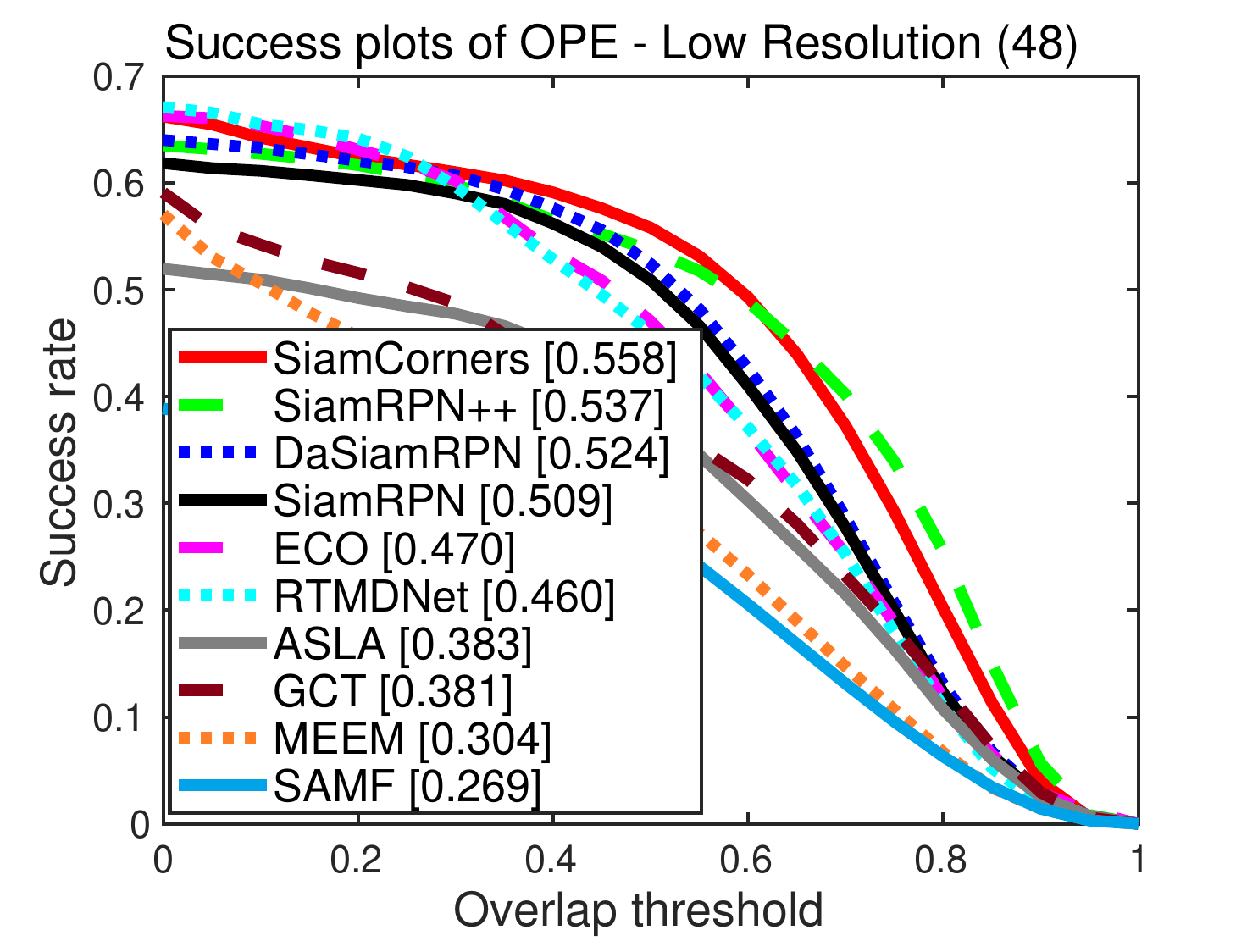}}
\subfigure[fast motion]{\includegraphics[width=0.245\textwidth]{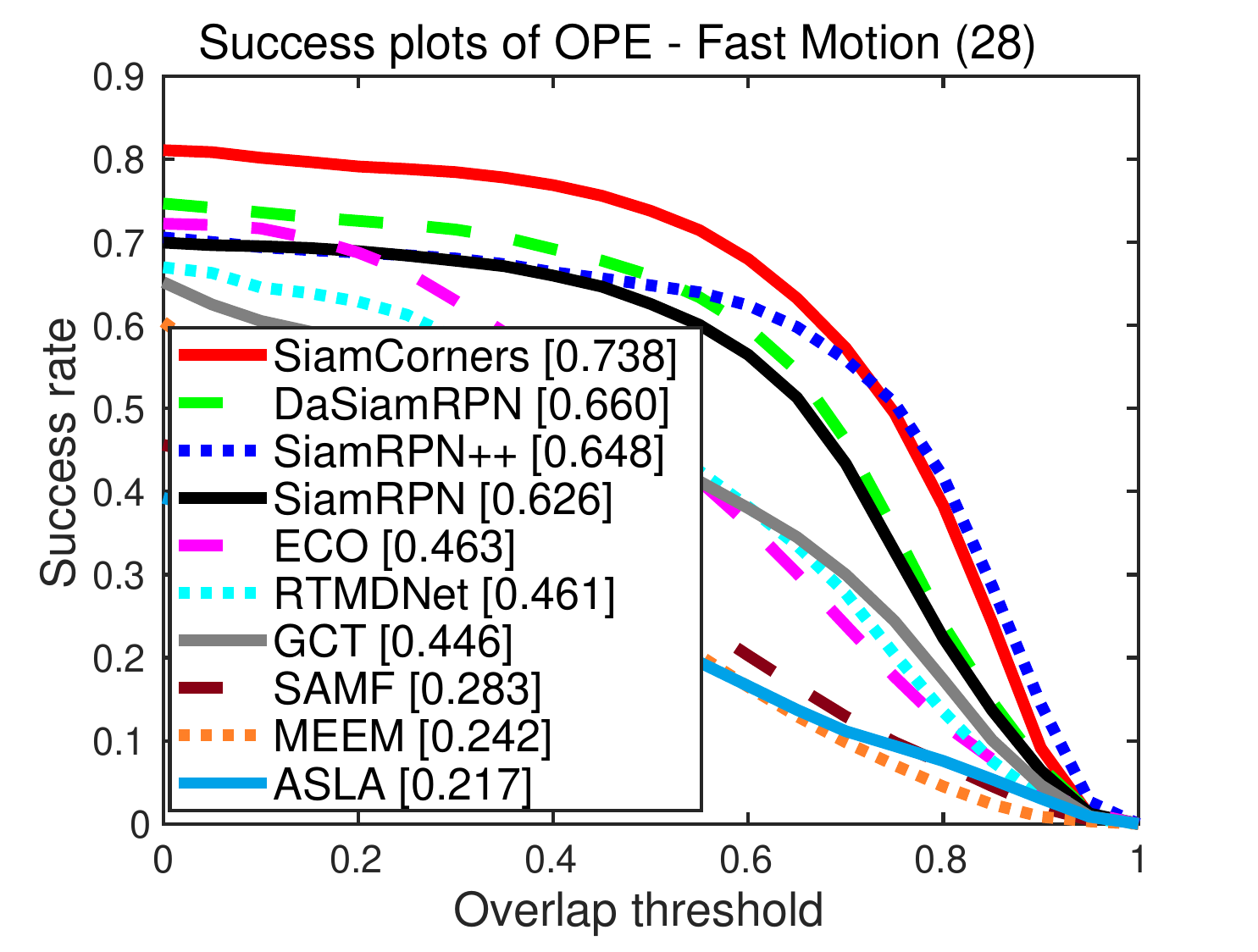}}
\subfigure[full occlusion]{\includegraphics[width=0.245\textwidth]{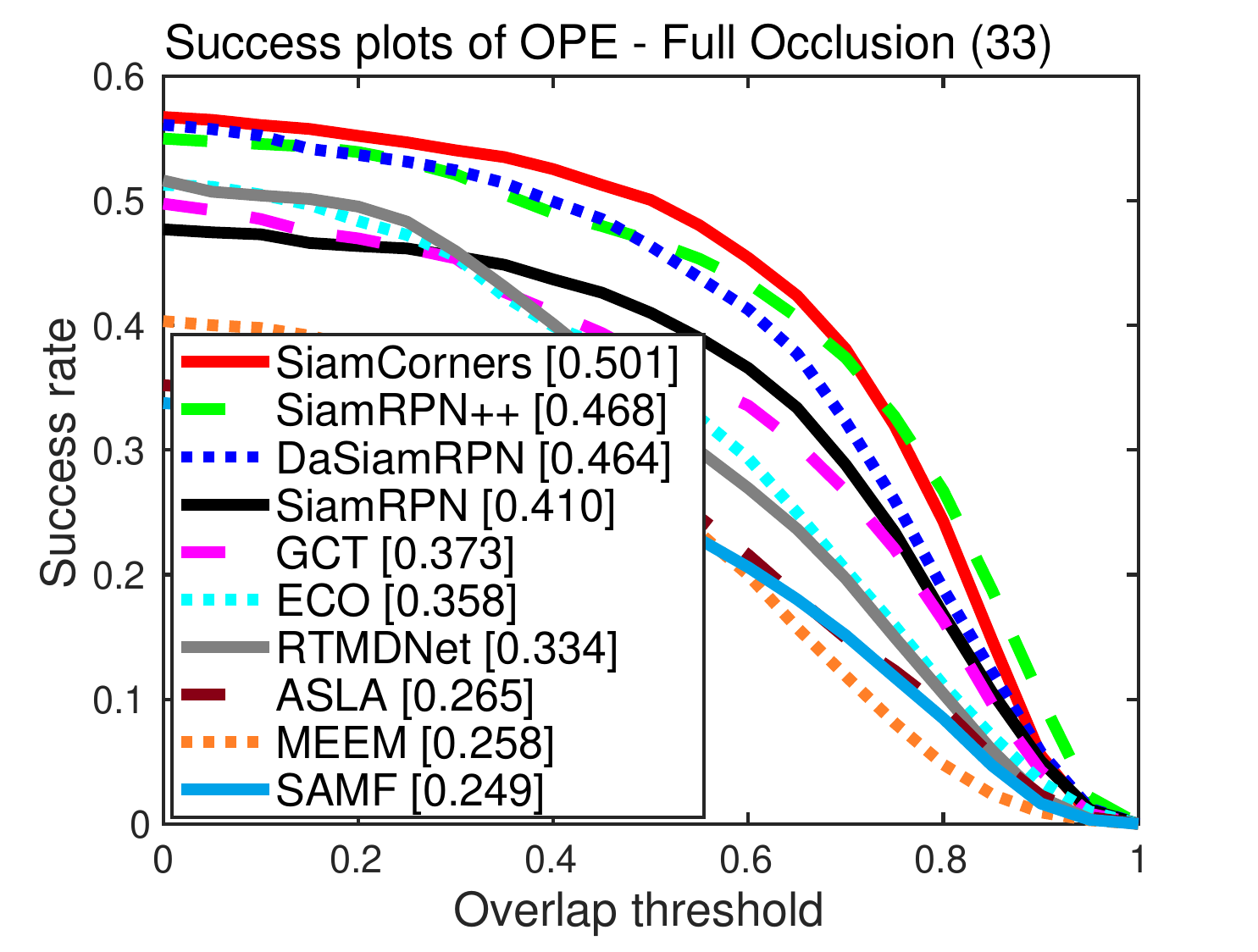}}
\subfigure[partial occlusion]{\includegraphics[width=0.245\textwidth]{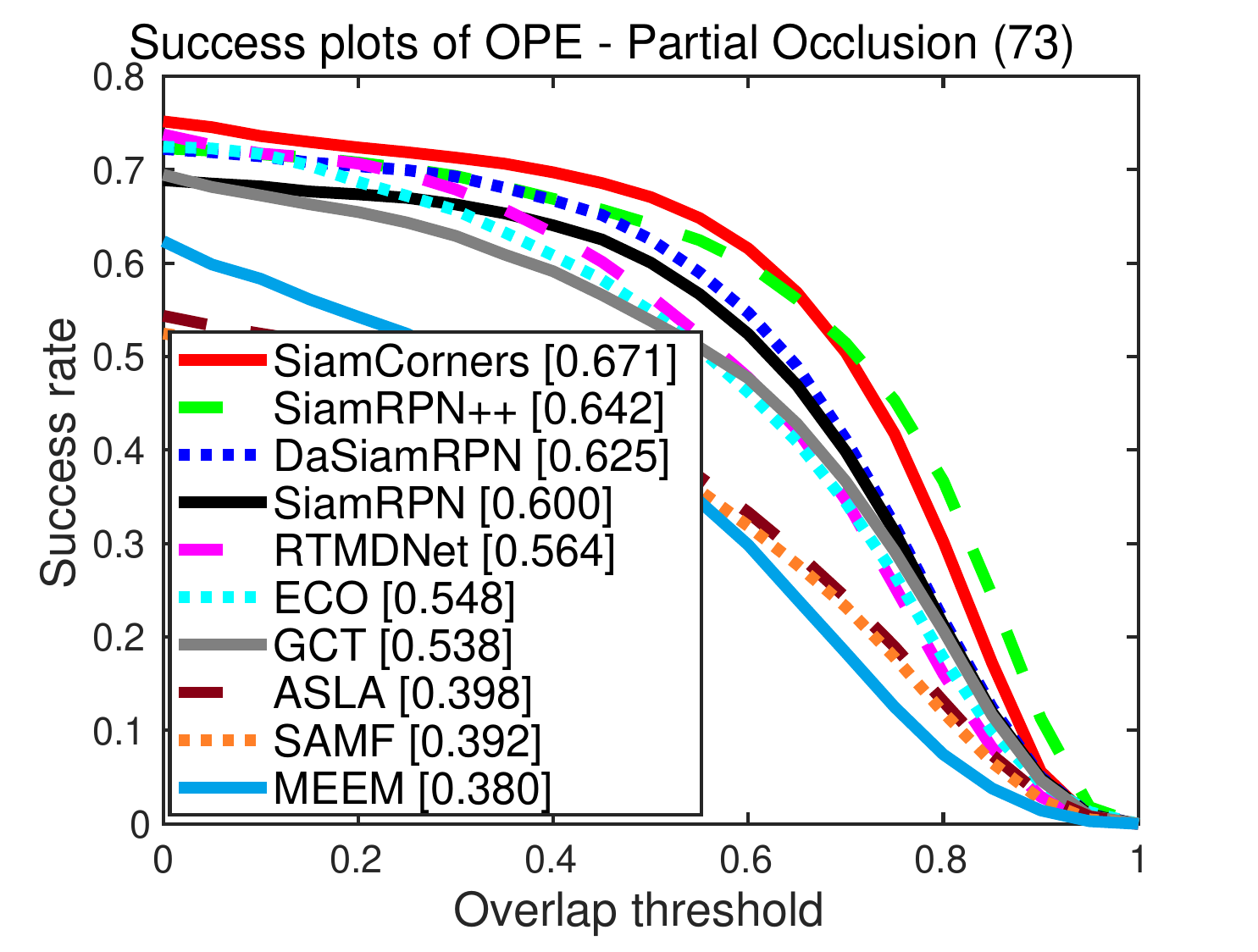}}
\subfigure[out-of-view]{\includegraphics[width=0.245\textwidth]{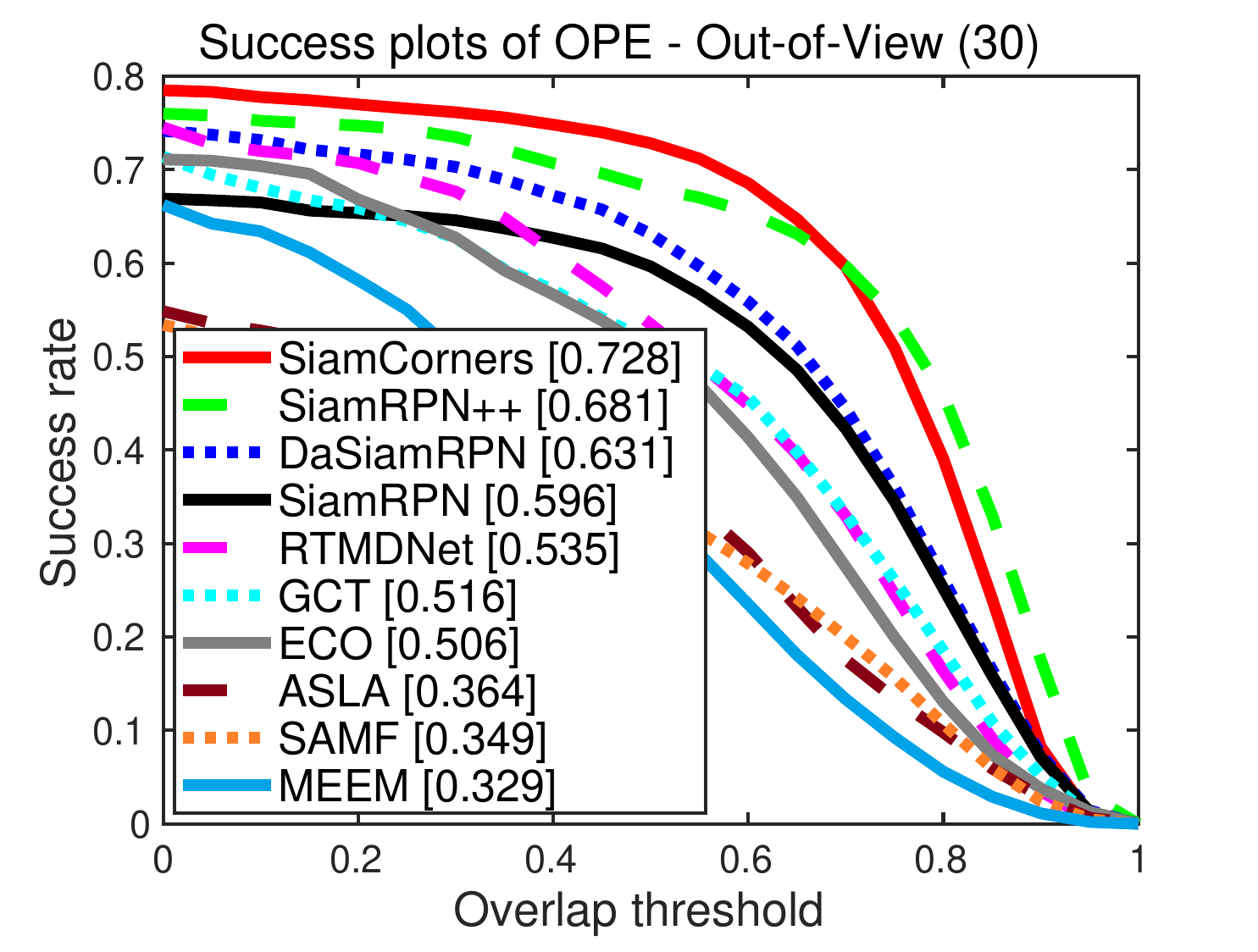}}
\subfigure[background clutter]{\includegraphics[width=0.245\textwidth]{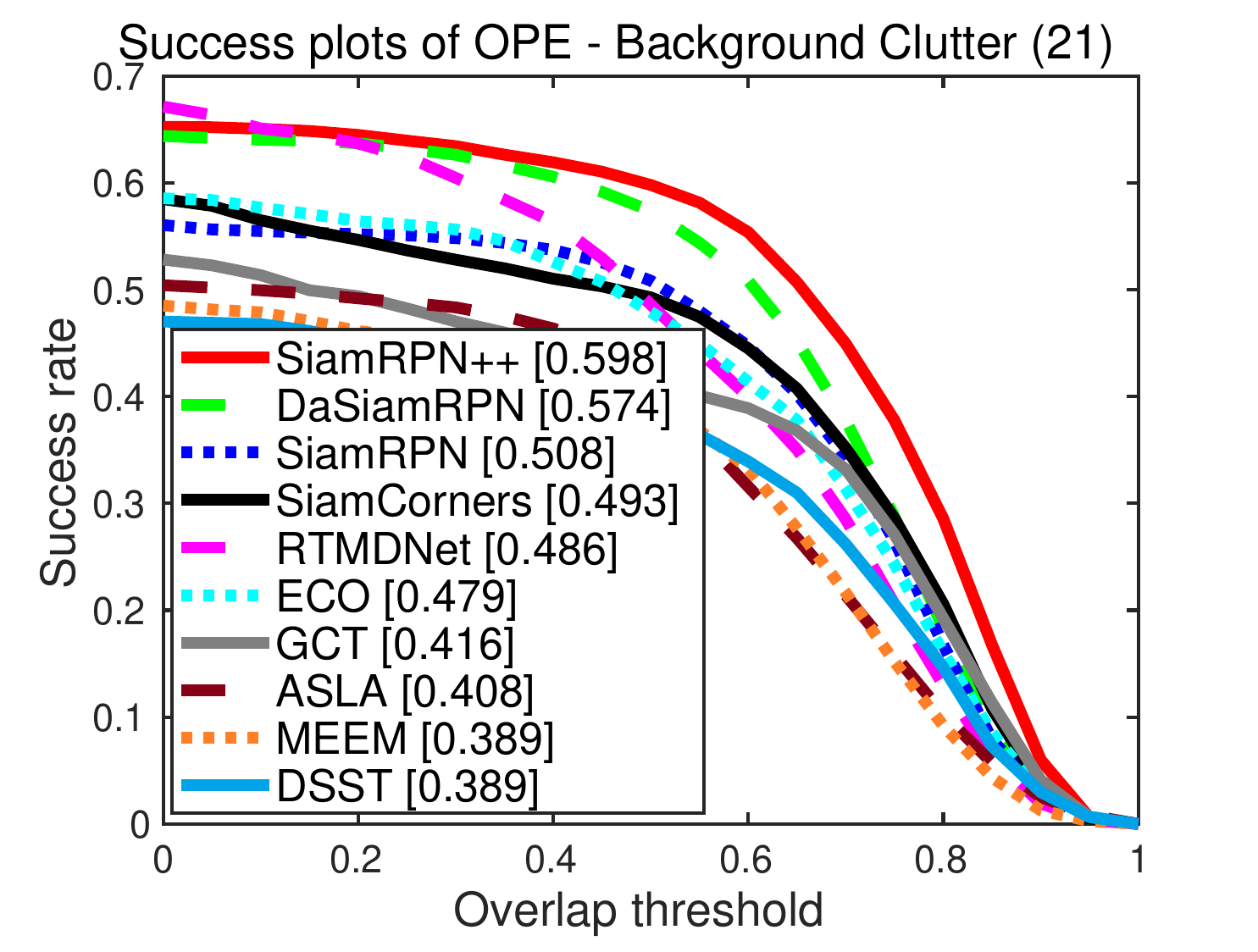}}
\subfigure[illumination variation]{\includegraphics[width=0.245\textwidth]{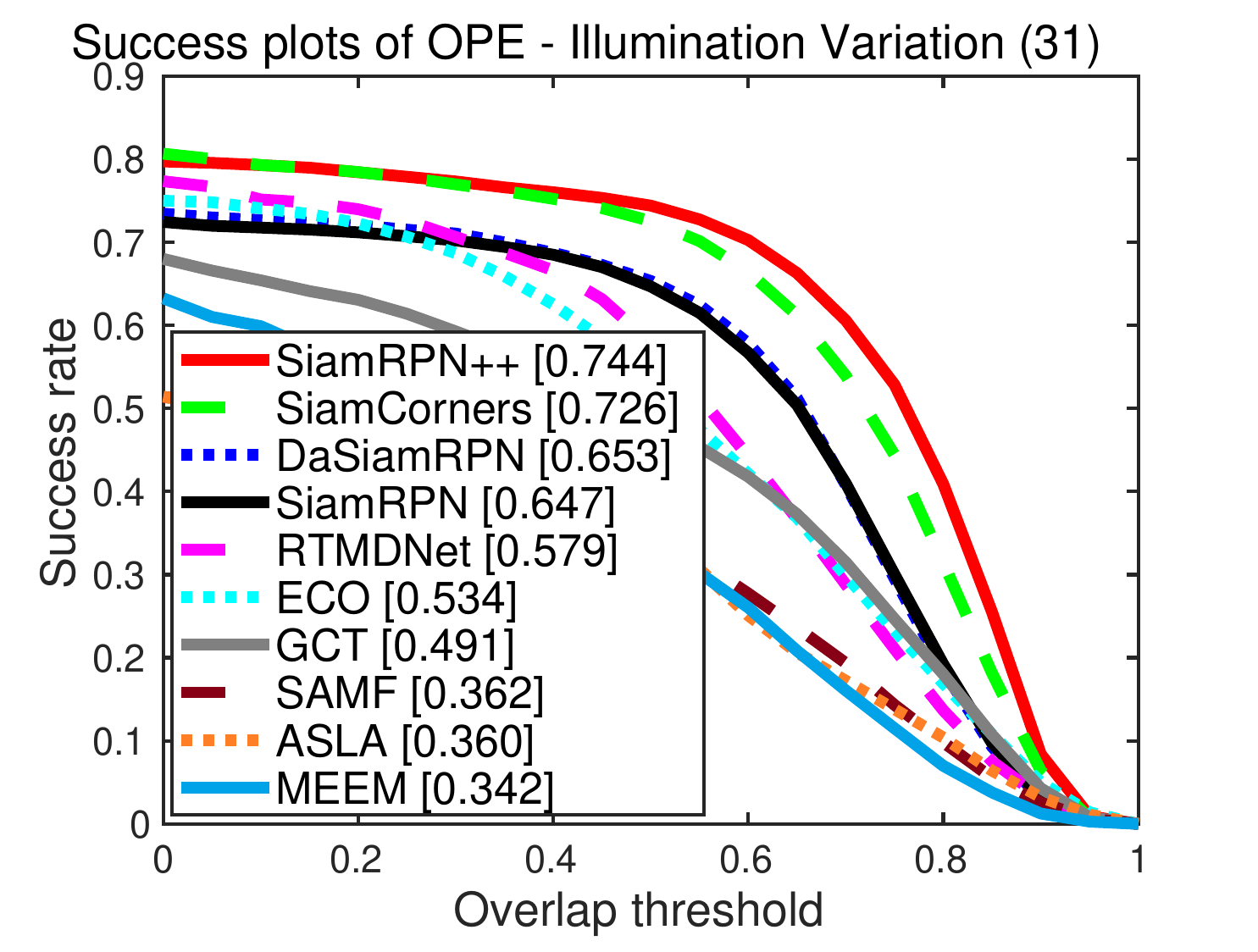}}
\subfigure[viewpoint change]{\includegraphics[width=0.245\textwidth]{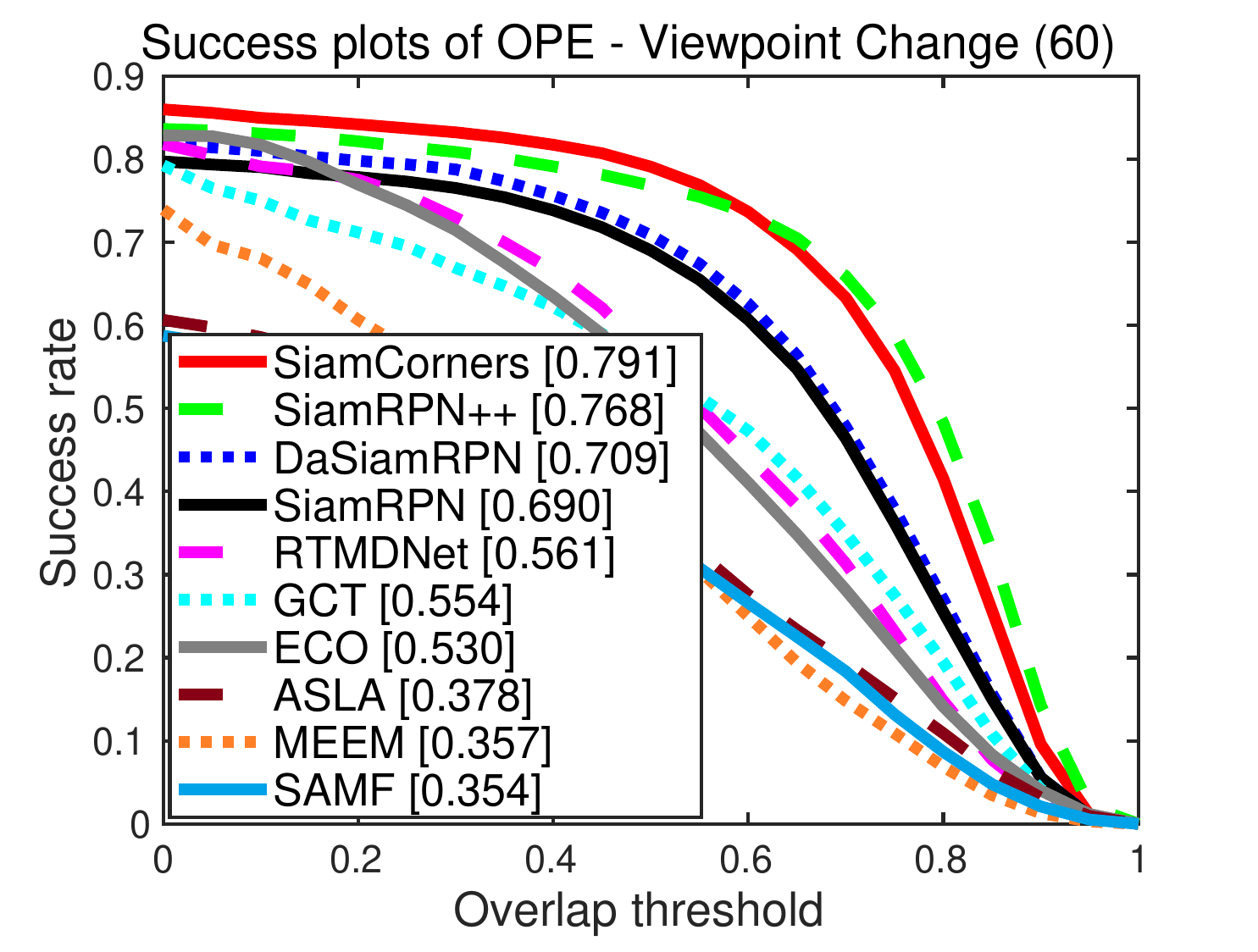}}
\subfigure[camera motion]{\includegraphics[width=0.245\textwidth]{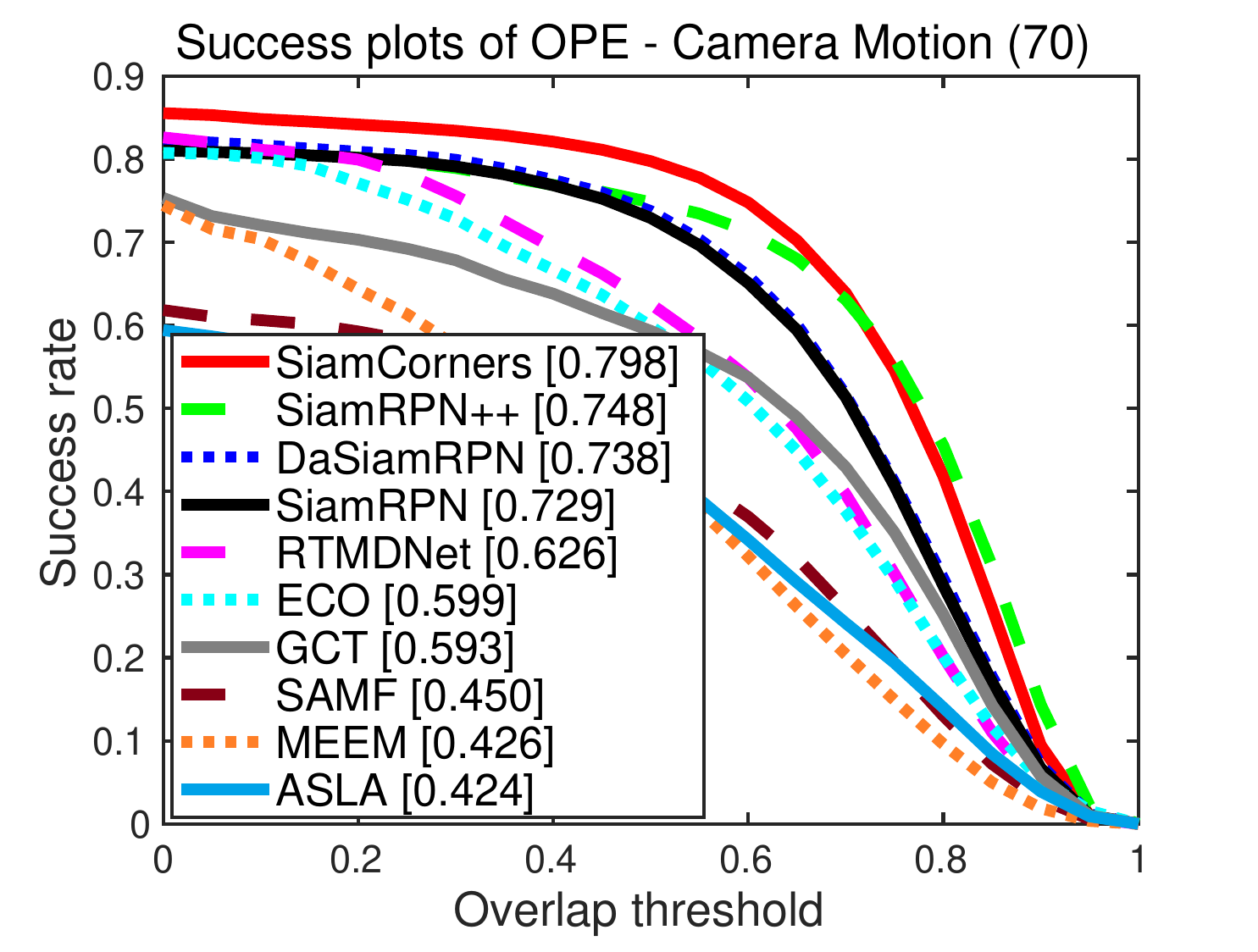}}
\subfigure[similar object]{\includegraphics[width=0.245\textwidth]{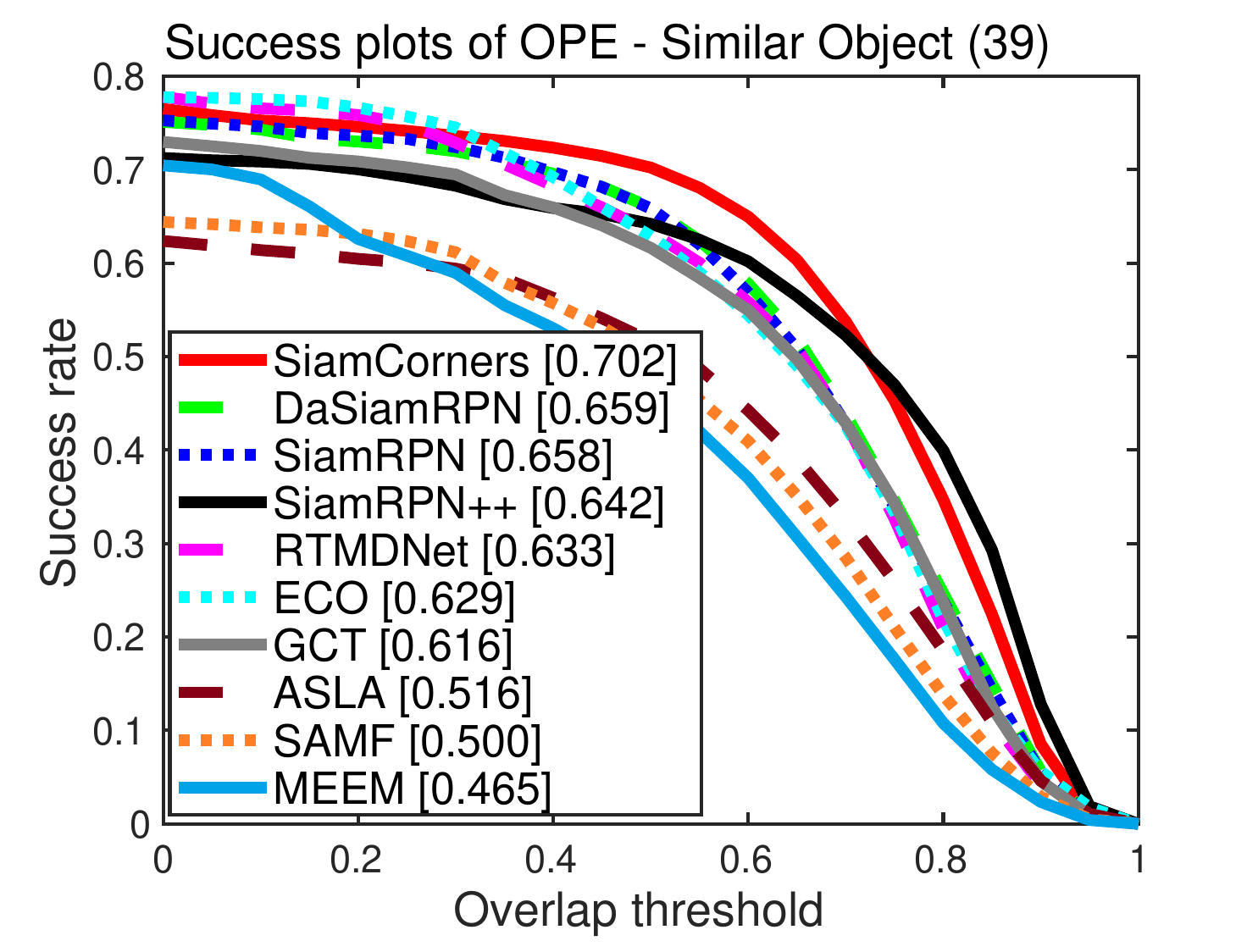}}
\caption{The success plots on UAV123 dataset for 12 challenging factors: scale variation,  aspect ratio change, low resolution, fast motion, full occlusion, partial occlusion, out-of-view, background clutter, illumination variation, viewpoint change, camera motion, and similar object. The legend shows the AUC score for each method.}
\label{uav123}
\end{figure*}

\begin{figure*}[!t]
\centering
\renewcommand\arraystretch{1.1}
\subfigure{\includegraphics[width=0.164\textwidth,height=1.635cm]{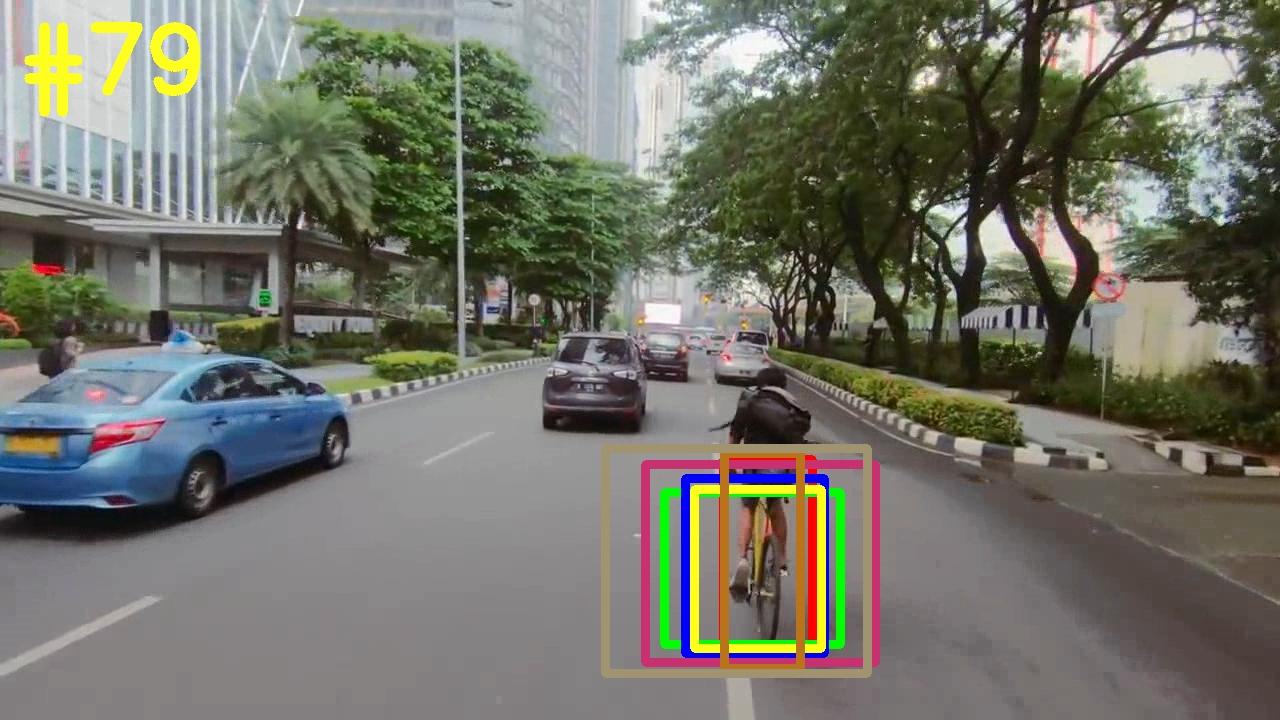}}\hspace{-0.027in}
\subfigure{\includegraphics[width=0.164\textwidth,height=1.635cm]{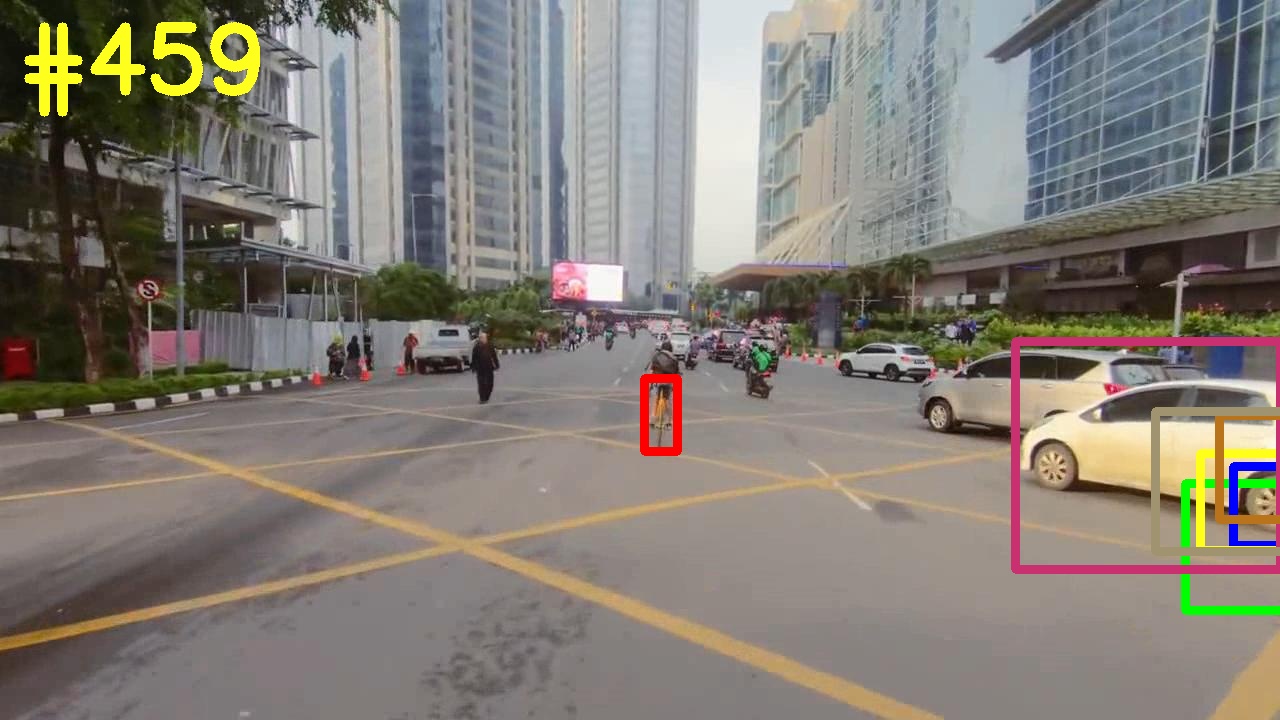}}\hspace{-0.027in}
\subfigure{\includegraphics[width=0.164\textwidth,height=1.635cm]{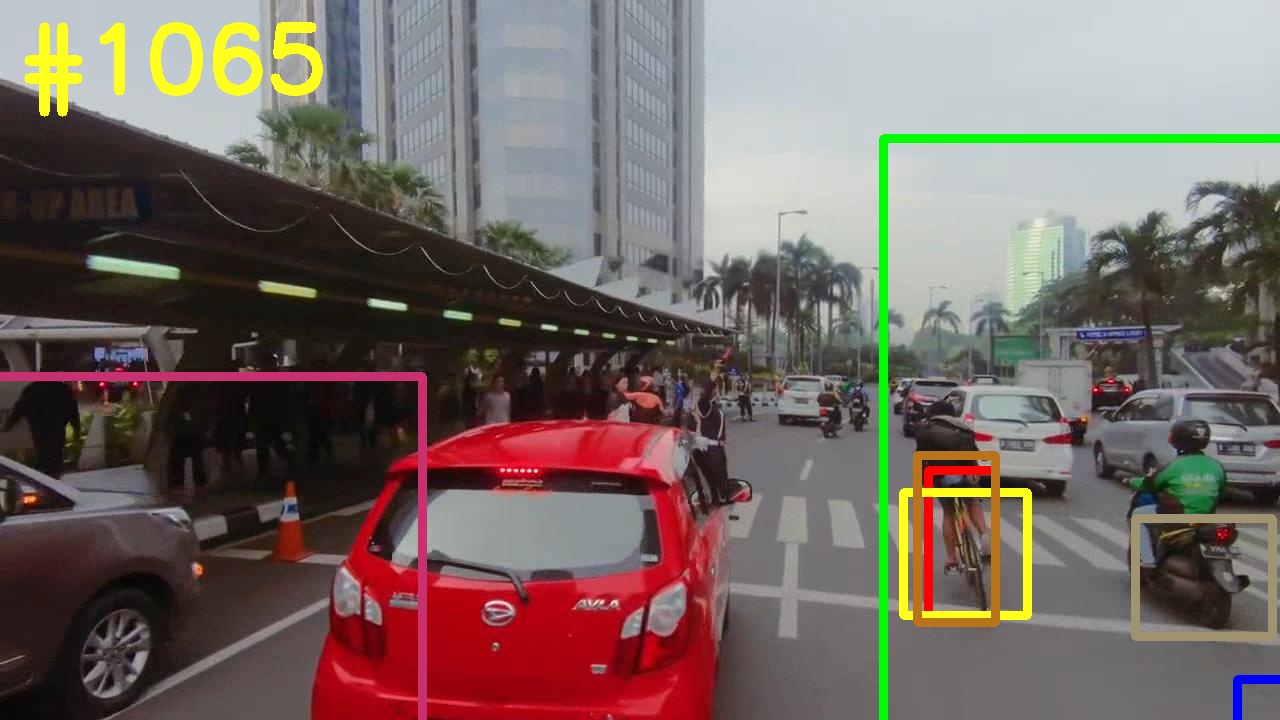}}\hspace{-0.027in}
\subfigure{\includegraphics[width=0.164\textwidth,height=1.635cm]{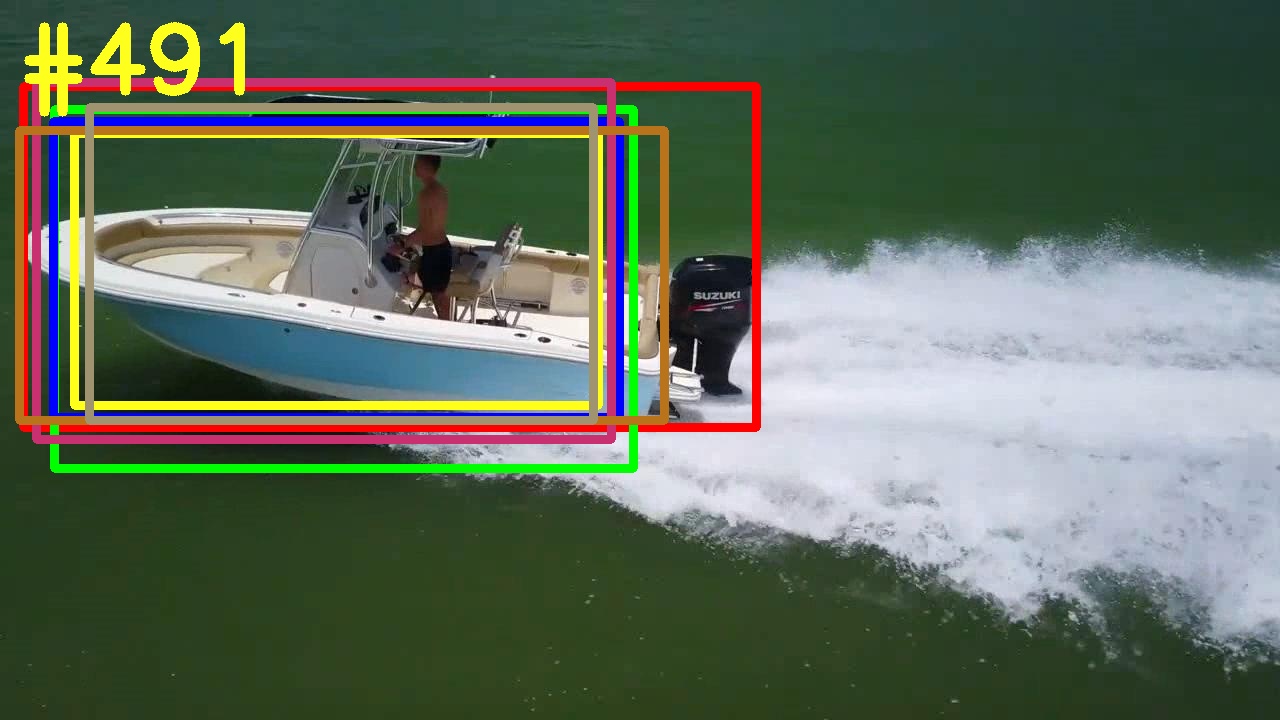}}\hspace{-0.027in}
\subfigure{\includegraphics[width=0.164\textwidth,height=1.635cm]{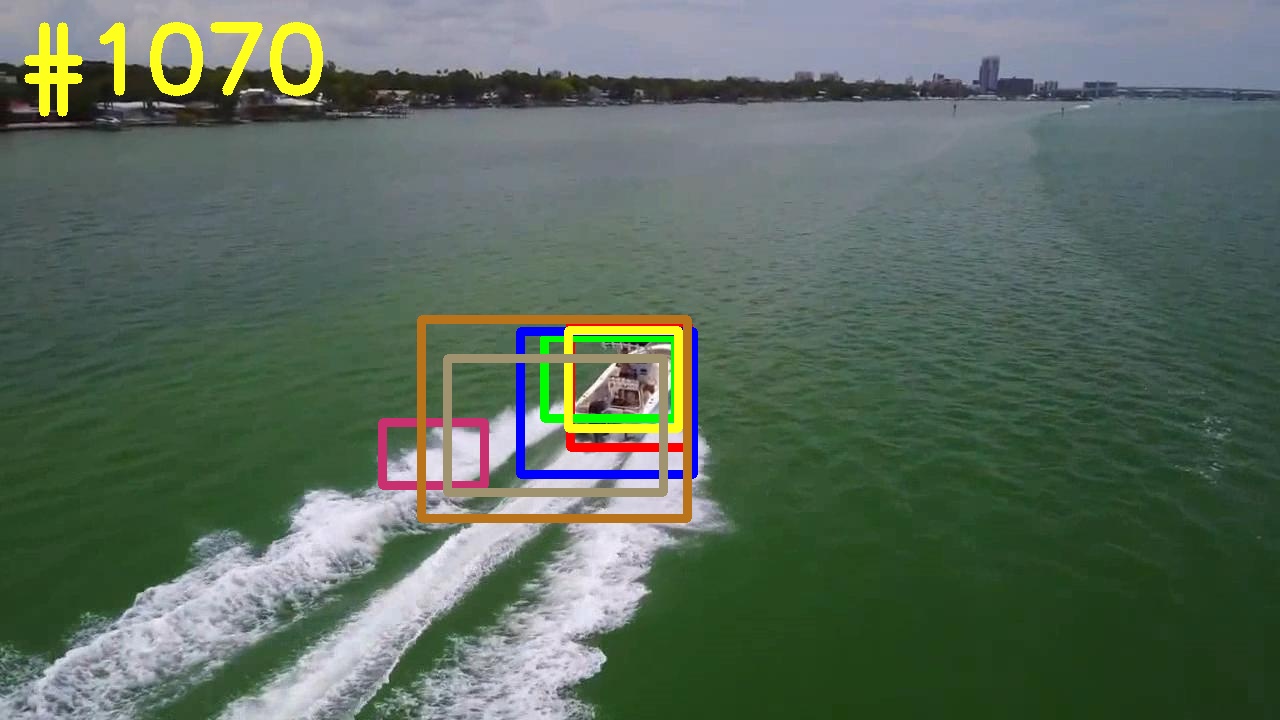}}\hspace{-0.027in}
\subfigure{\includegraphics[width=0.164\textwidth,height=1.635cm]{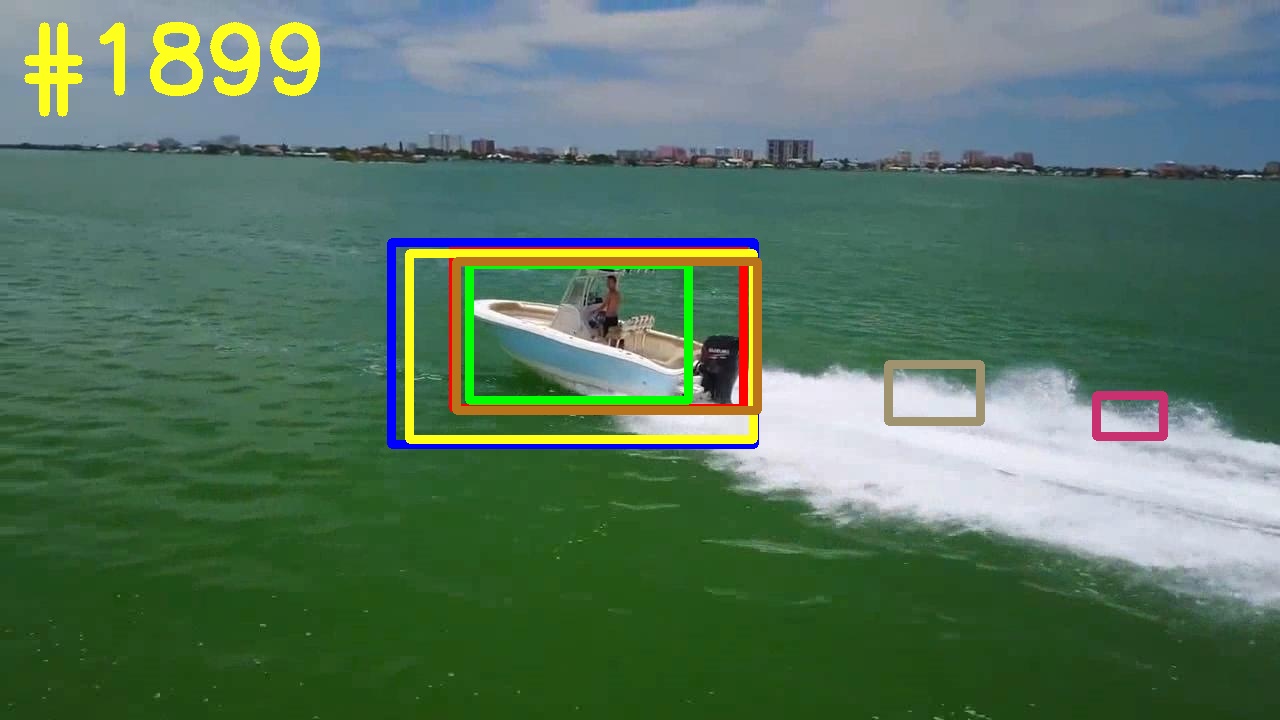}}\hspace{-0.027in}\vspace{-0.117in}
\subfigure{\includegraphics[width=0.164\textwidth,height=1.635cm]{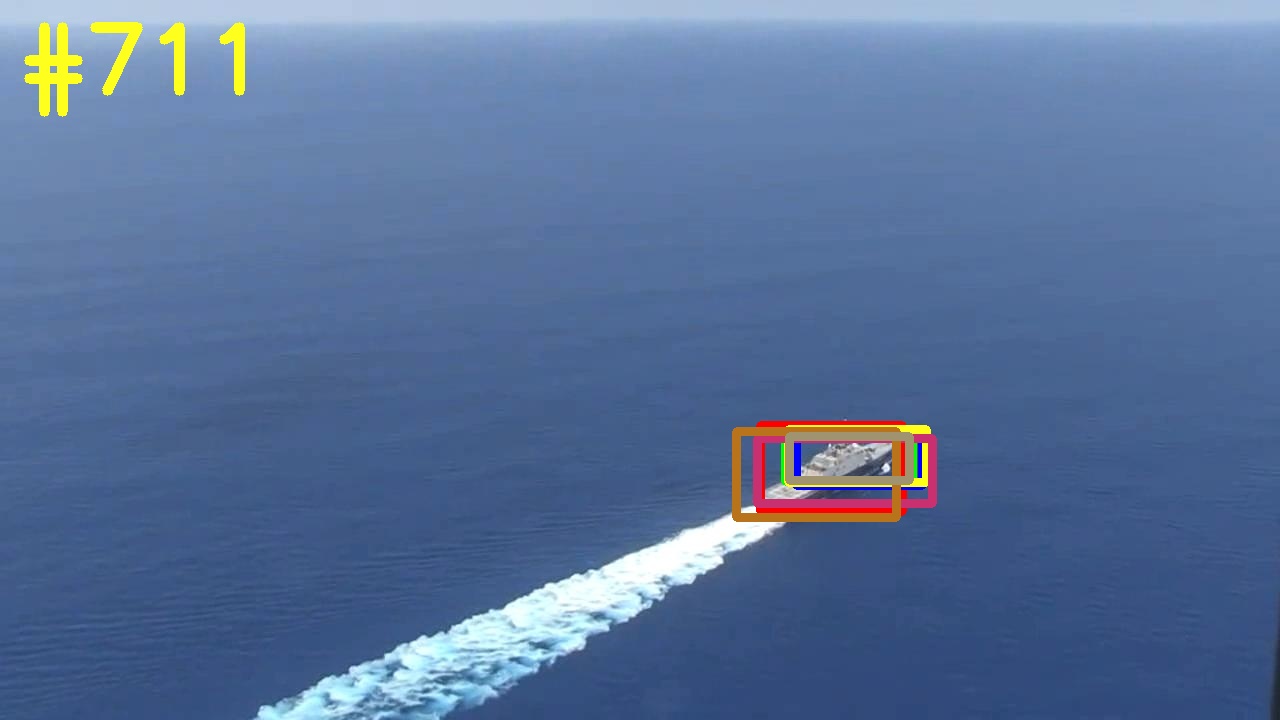}}\hspace{-0.027in}
\subfigure{\includegraphics[width=0.164\textwidth,height=1.635cm]{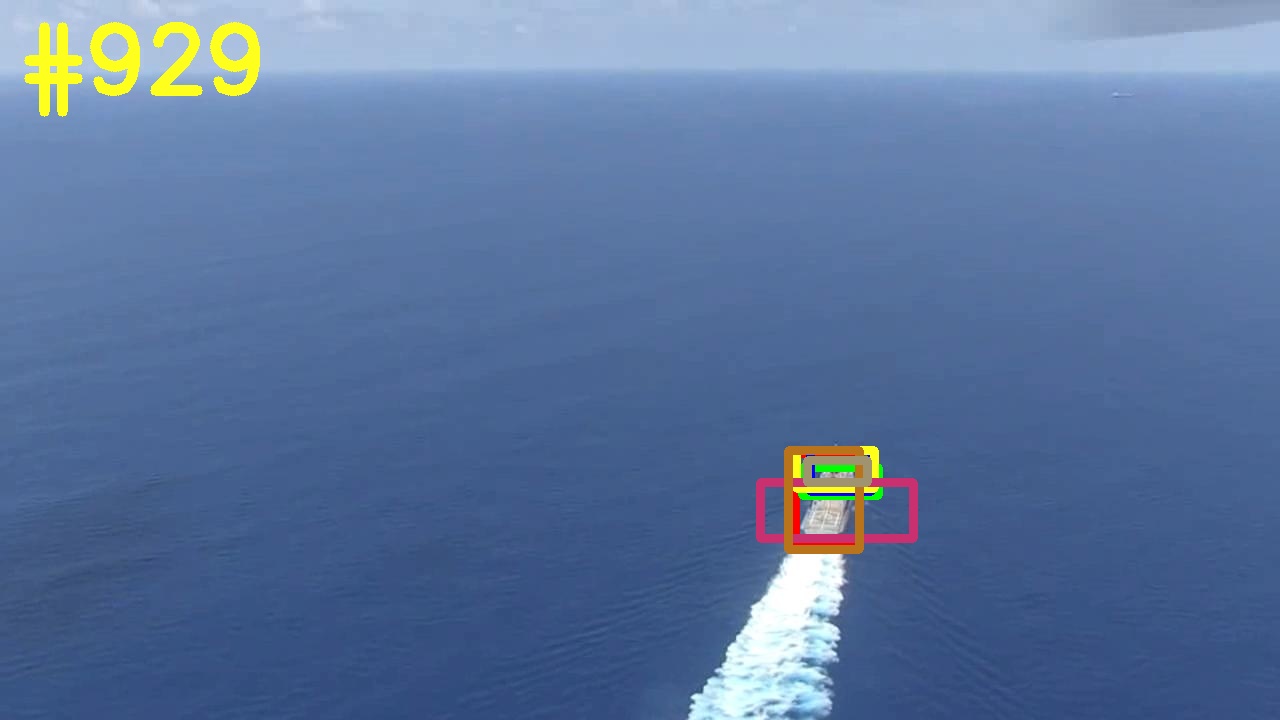}}\hspace{-0.027in}
\subfigure{\includegraphics[width=0.164\textwidth,height=1.635cm]{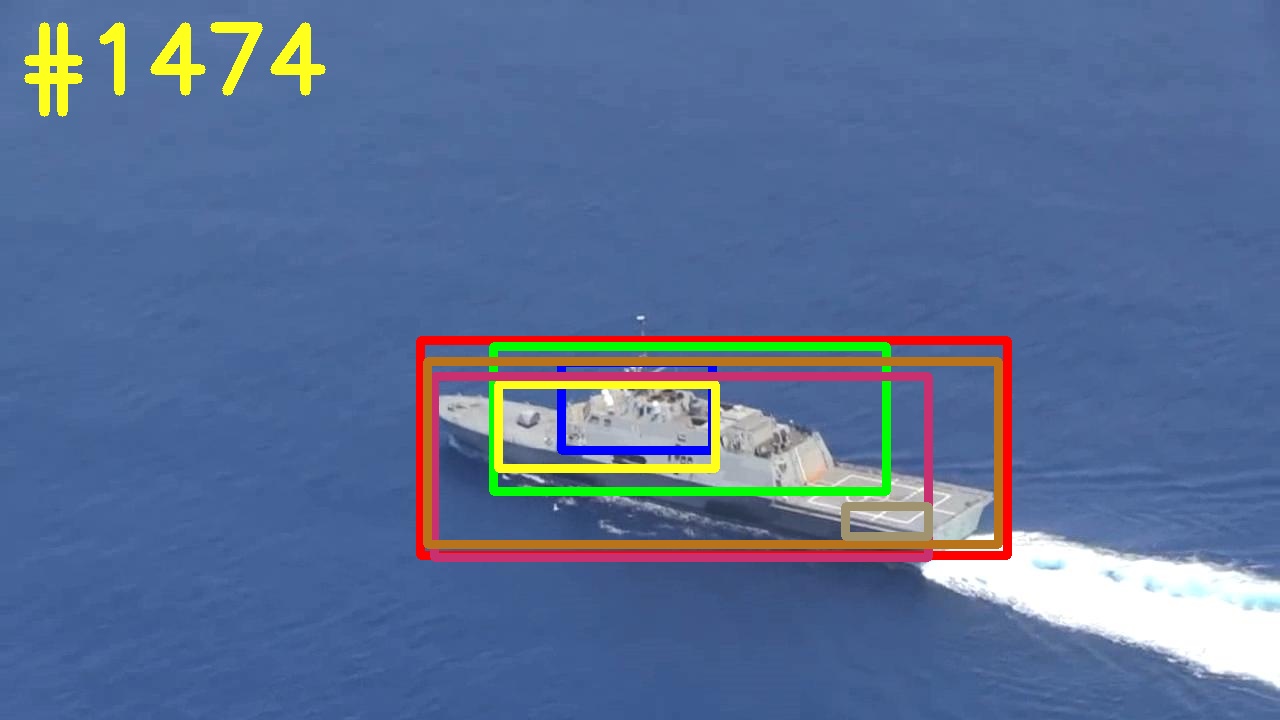}}\hspace{-0.027in}
\subfigure{\includegraphics[width=0.164\textwidth,height=1.635cm]{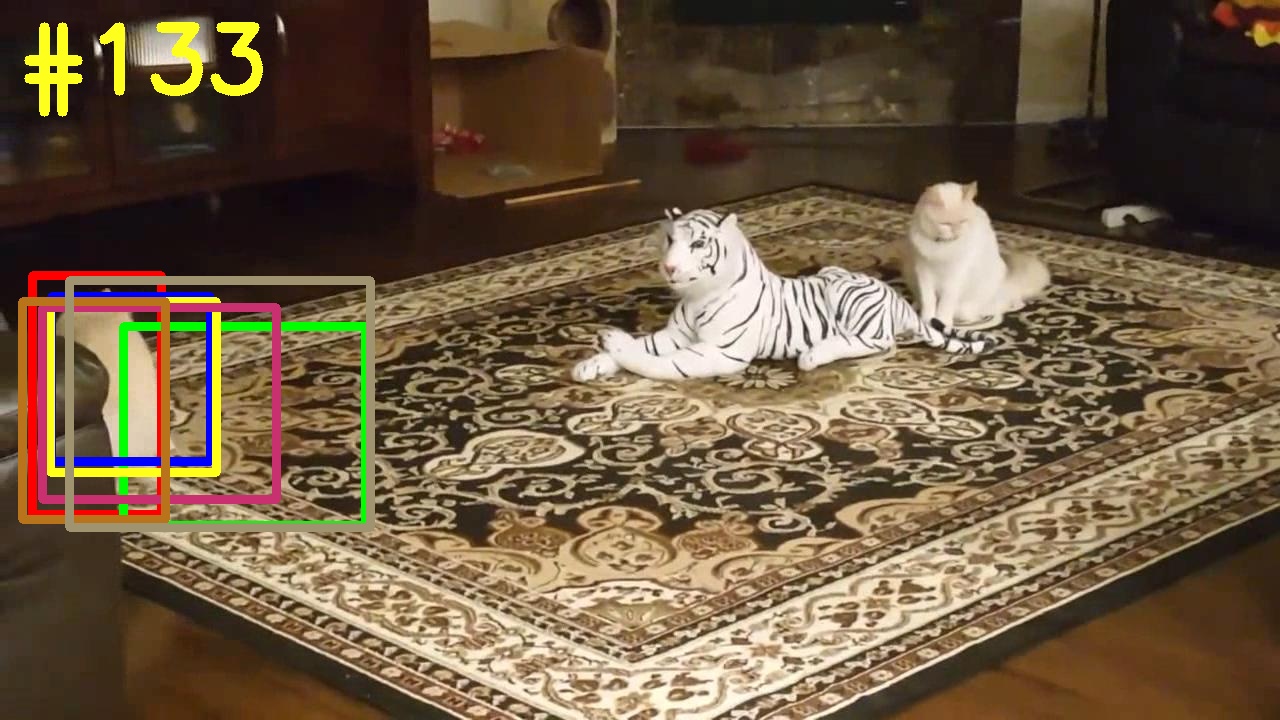}}\hspace{-0.027in}
\subfigure{\includegraphics[width=0.164\textwidth,height=1.635cm]{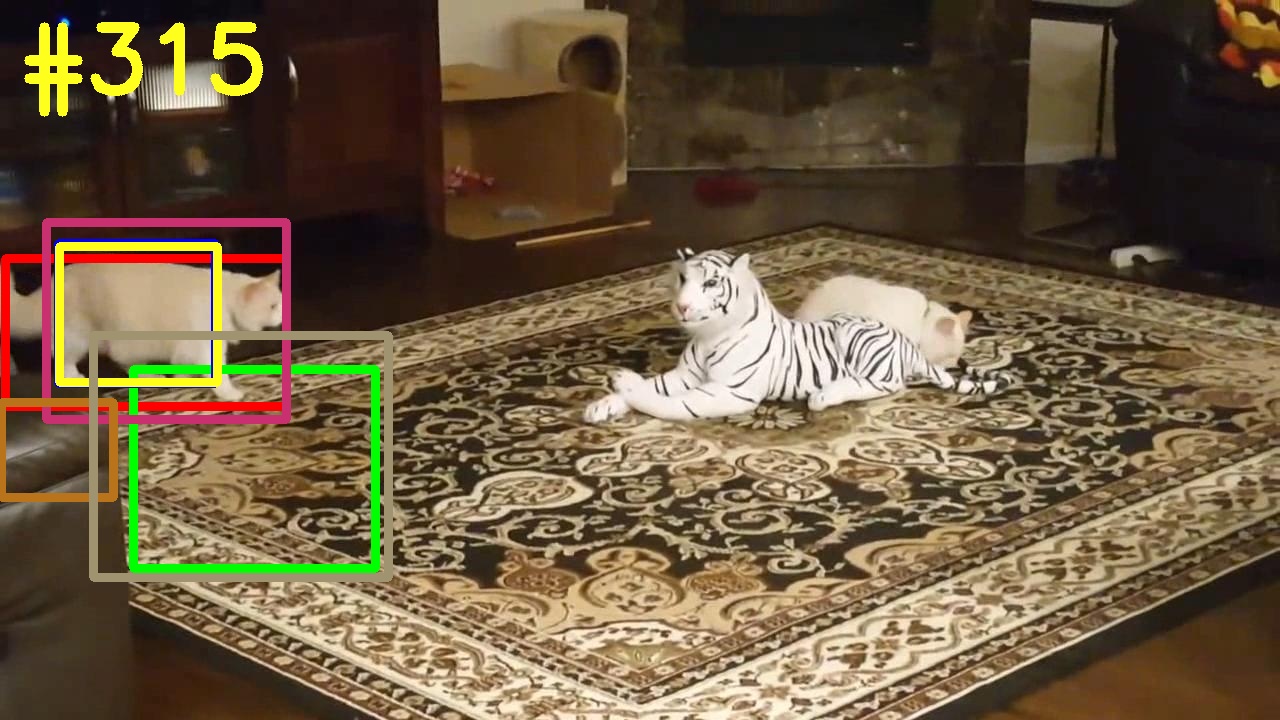}}\hspace{-0.027in}
\subfigure{\includegraphics[width=0.164\textwidth,height=1.635cm]{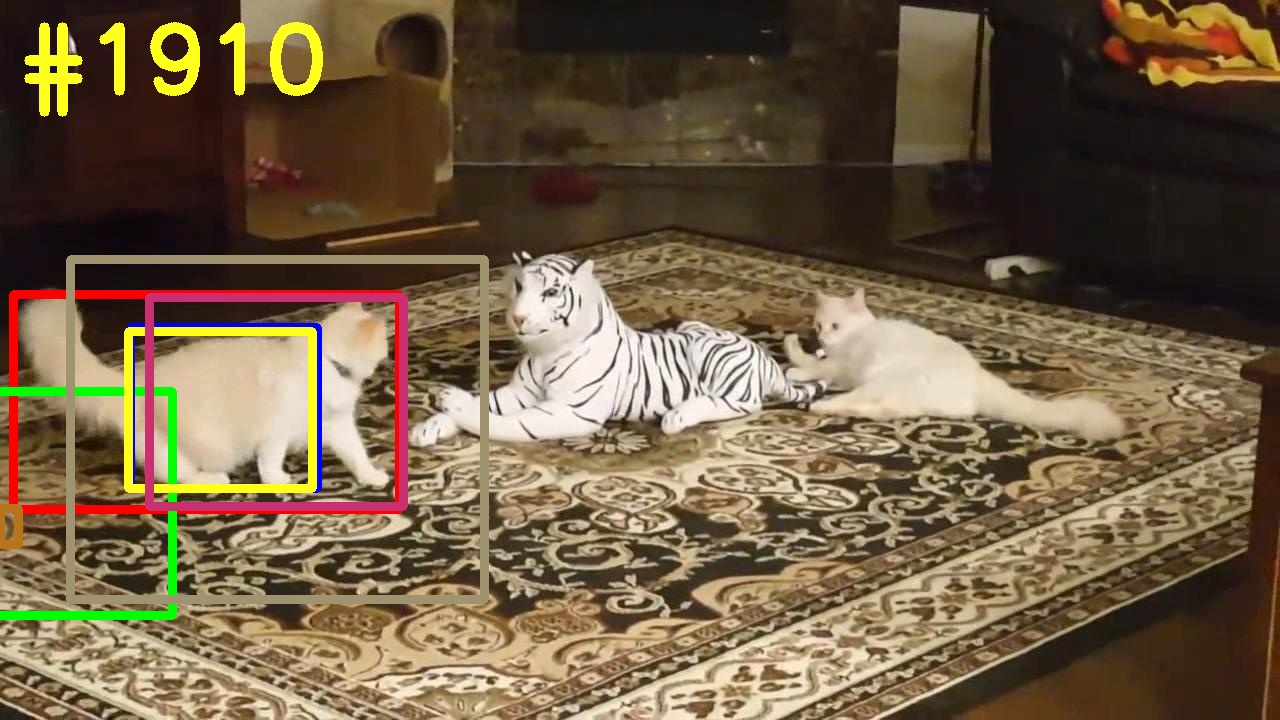}}\hspace{-0.027in}\vspace{-0.117in}
\subfigure{\includegraphics[width=0.164\textwidth,height=1.635cm]{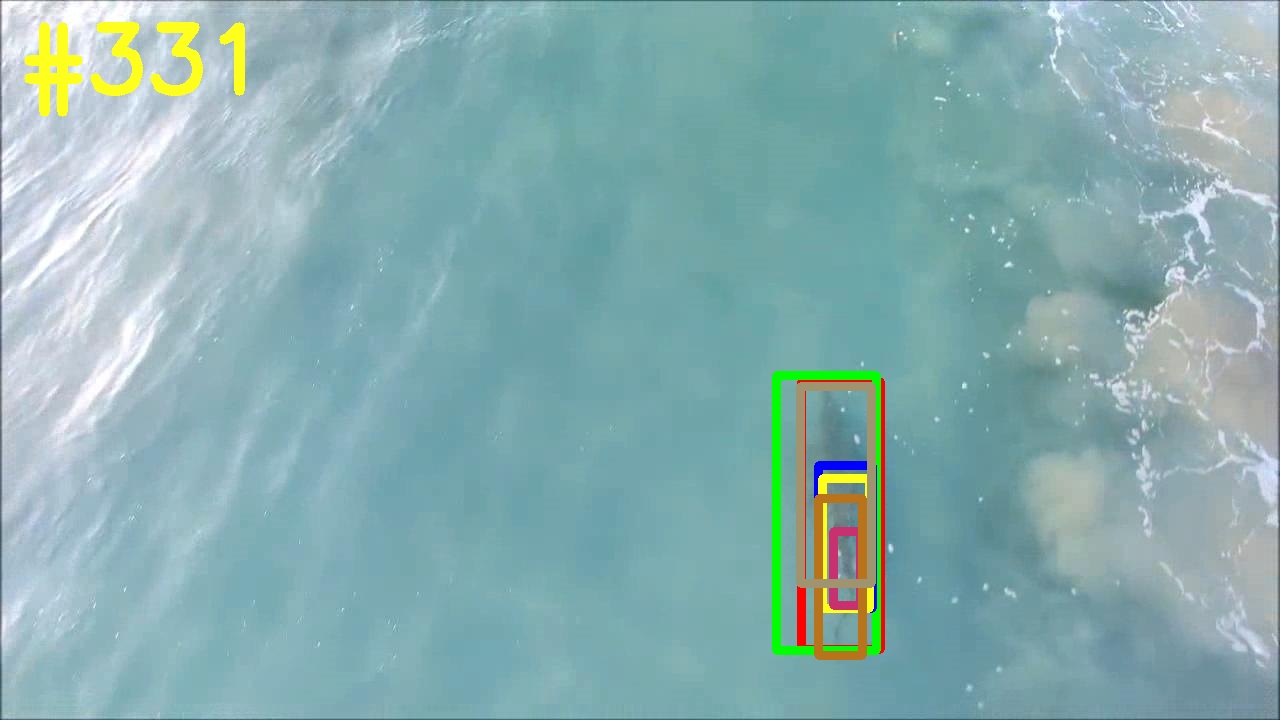}}\hspace{-0.027in}
\subfigure{\includegraphics[width=0.164\textwidth,height=1.635cm]{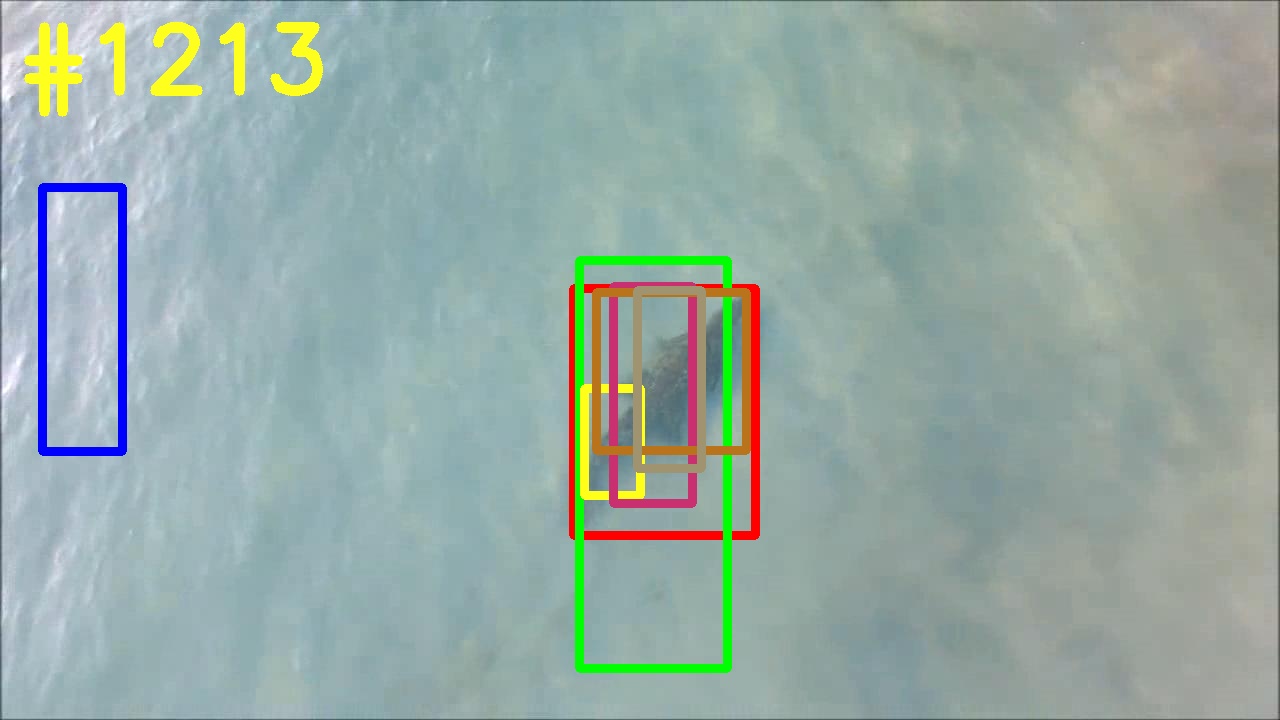}}\hspace{-0.027in}
\subfigure{\includegraphics[width=0.164\textwidth,height=1.635cm]{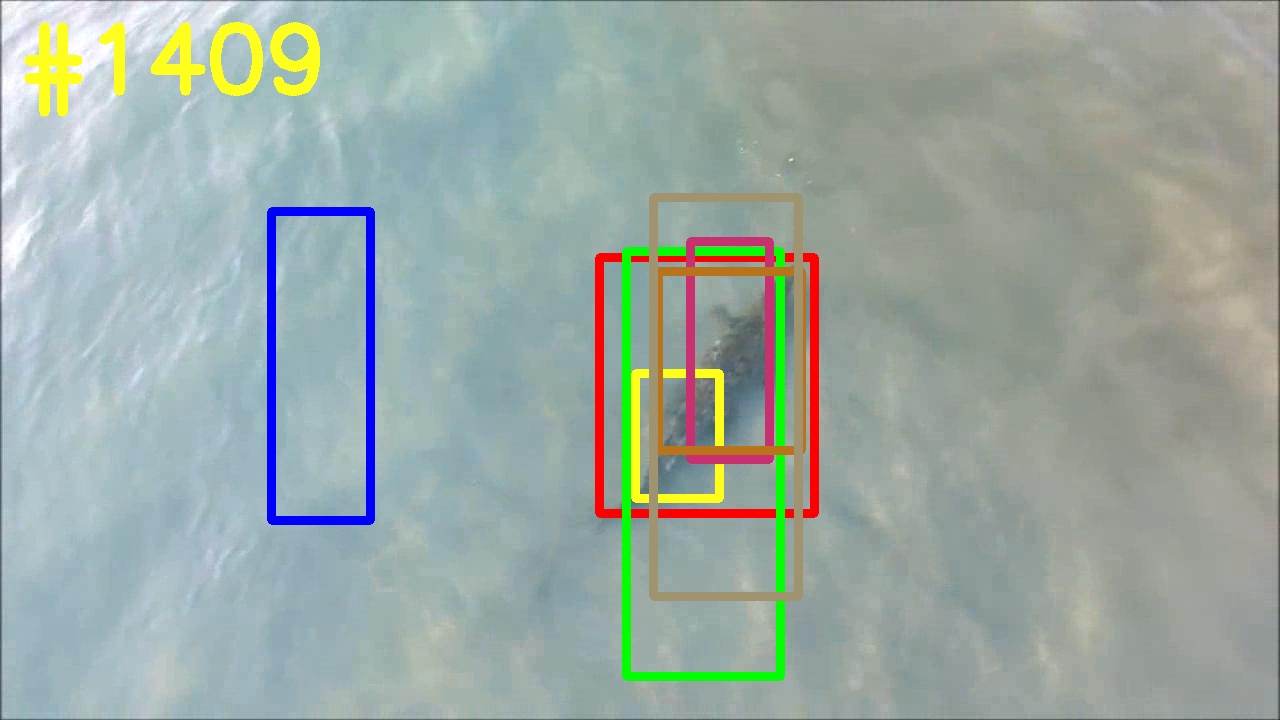}}\hspace{-0.027in}
\subfigure{\includegraphics[width=0.164\textwidth,height=1.635cm]{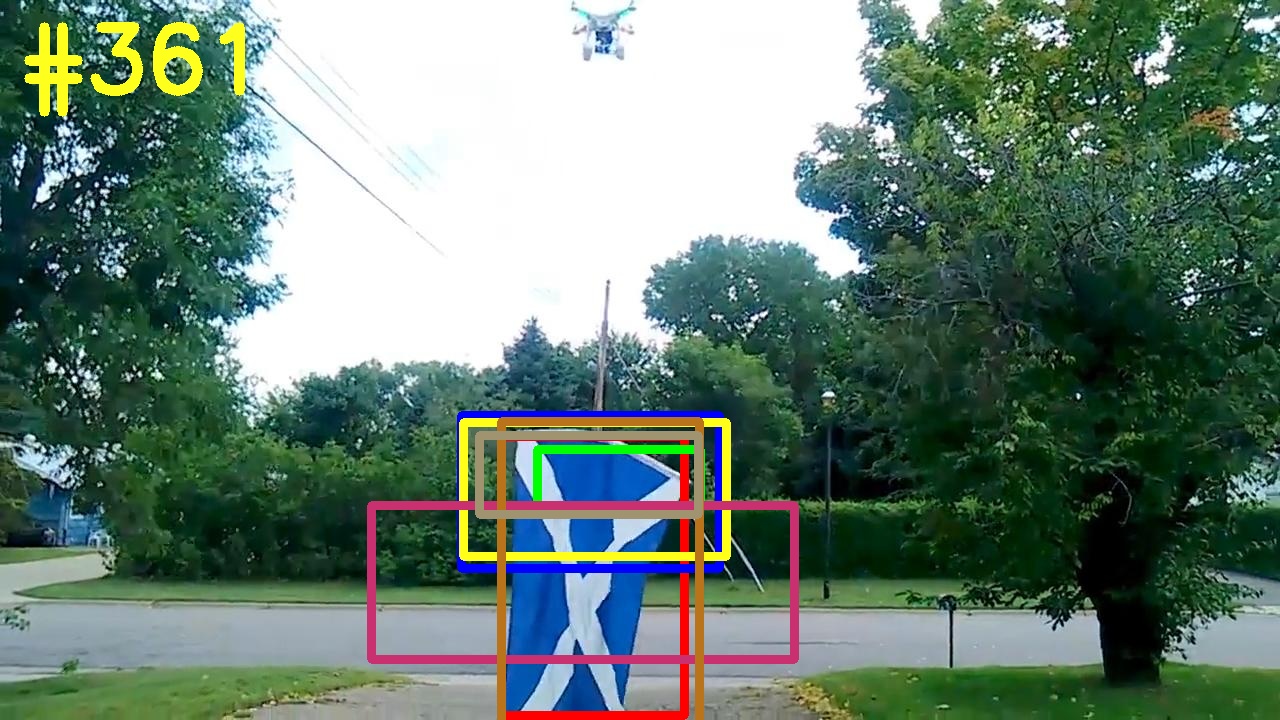}}\hspace{-0.027in}
\subfigure{\includegraphics[width=0.164\textwidth,height=1.635cm]{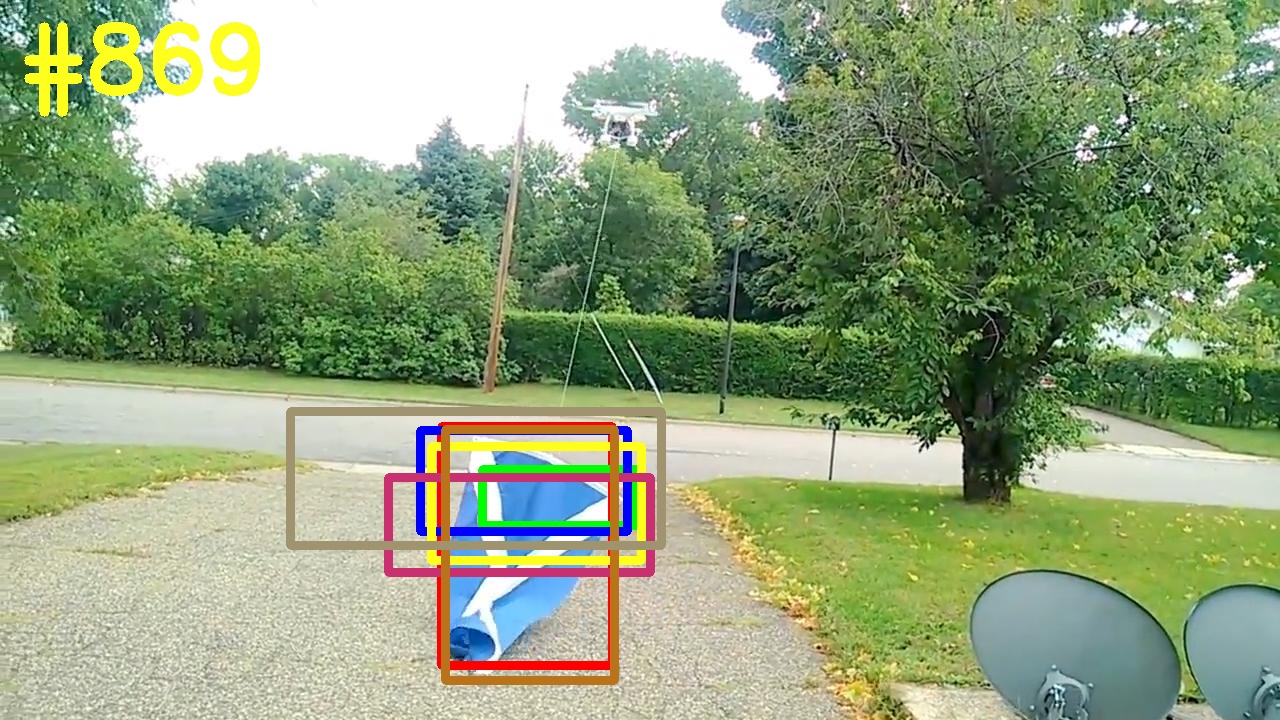}}\hspace{-0.027in}
\subfigure{\includegraphics[width=0.164\textwidth,height=1.635cm]{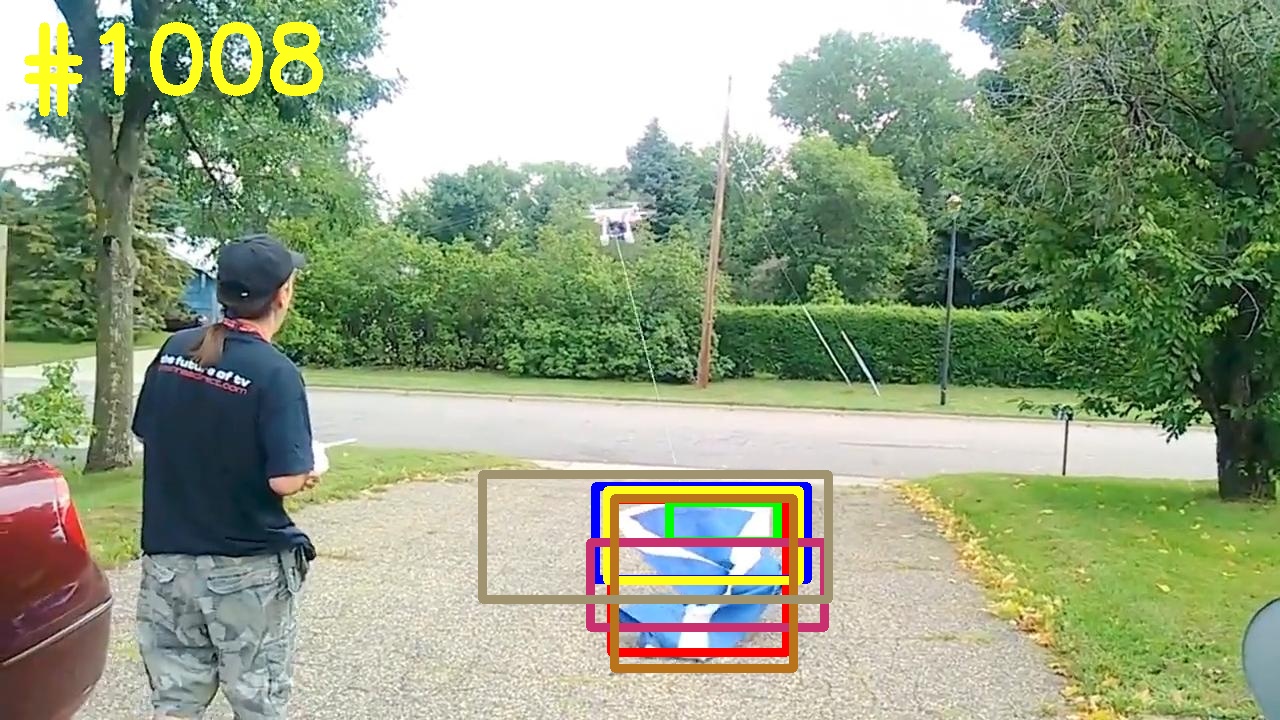}}\hspace{-0.027in}\vspace{-0.117in}
\subfigure{\includegraphics[width=0.164\textwidth,height=1.635cm]{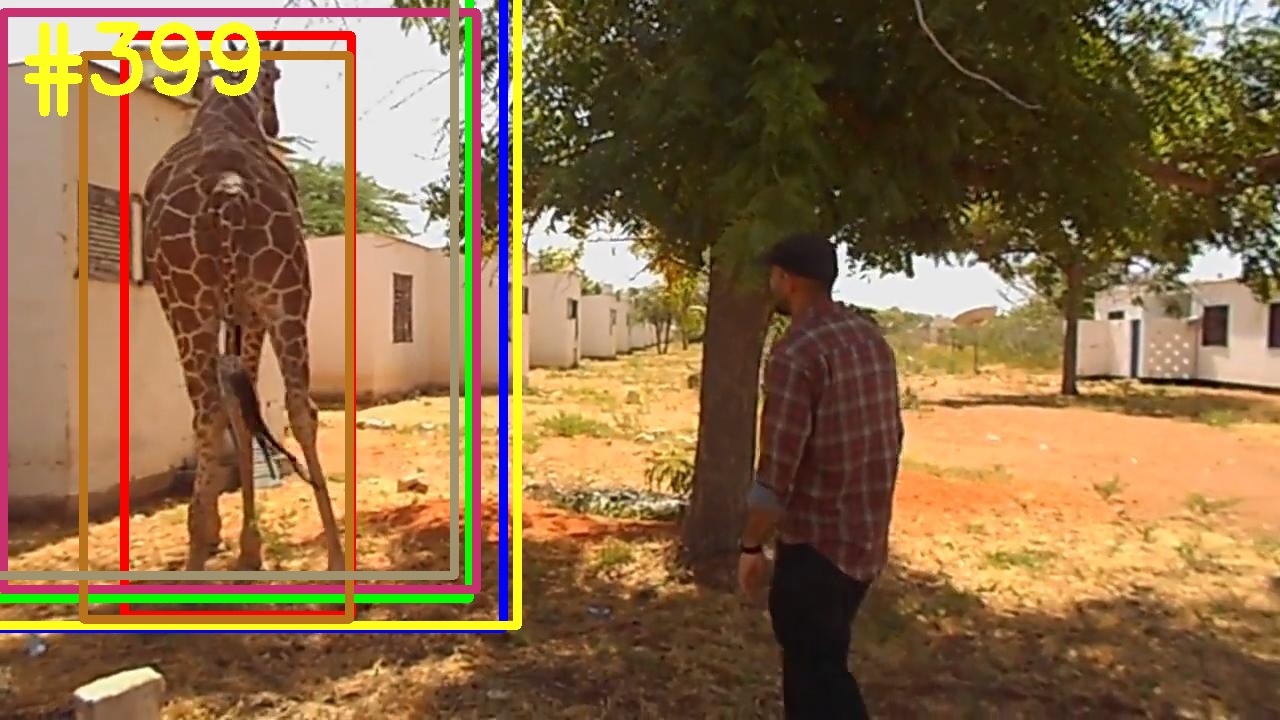}}\hspace{-0.027in}
\subfigure{\includegraphics[width=0.164\textwidth,height=1.635cm]{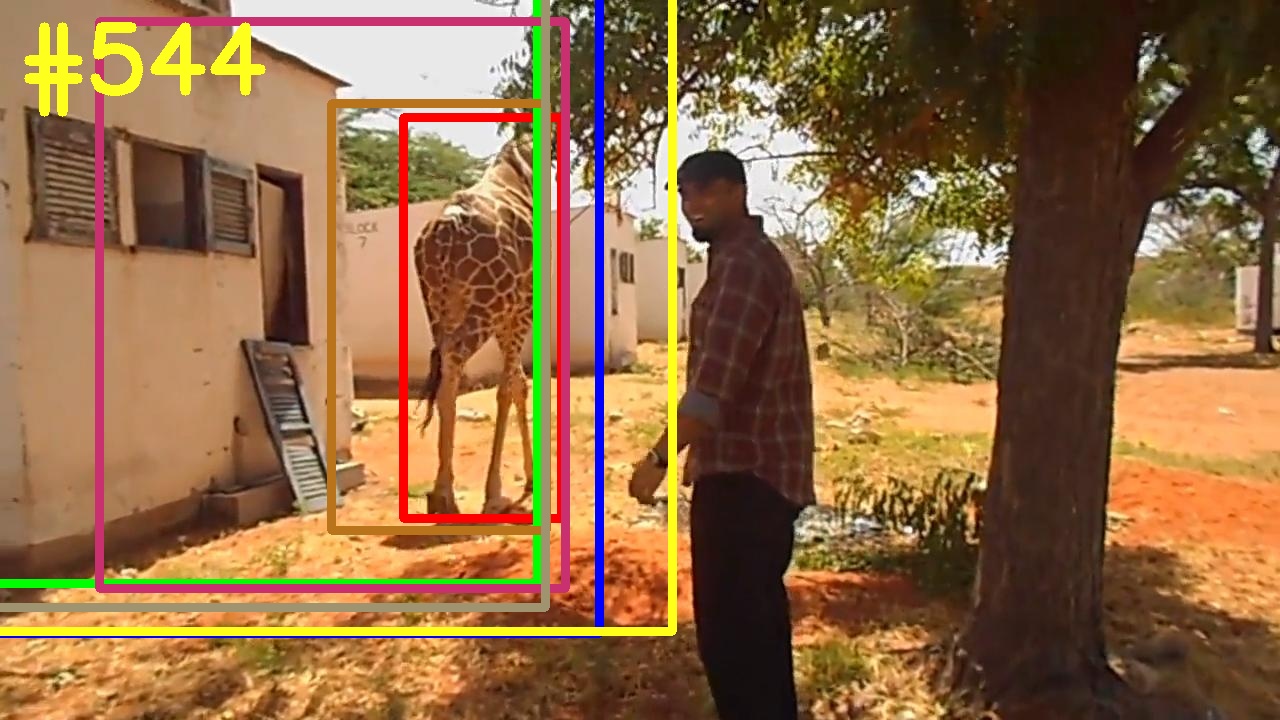}}\hspace{-0.027in}
\subfigure{\includegraphics[width=0.164\textwidth,height=1.635cm]{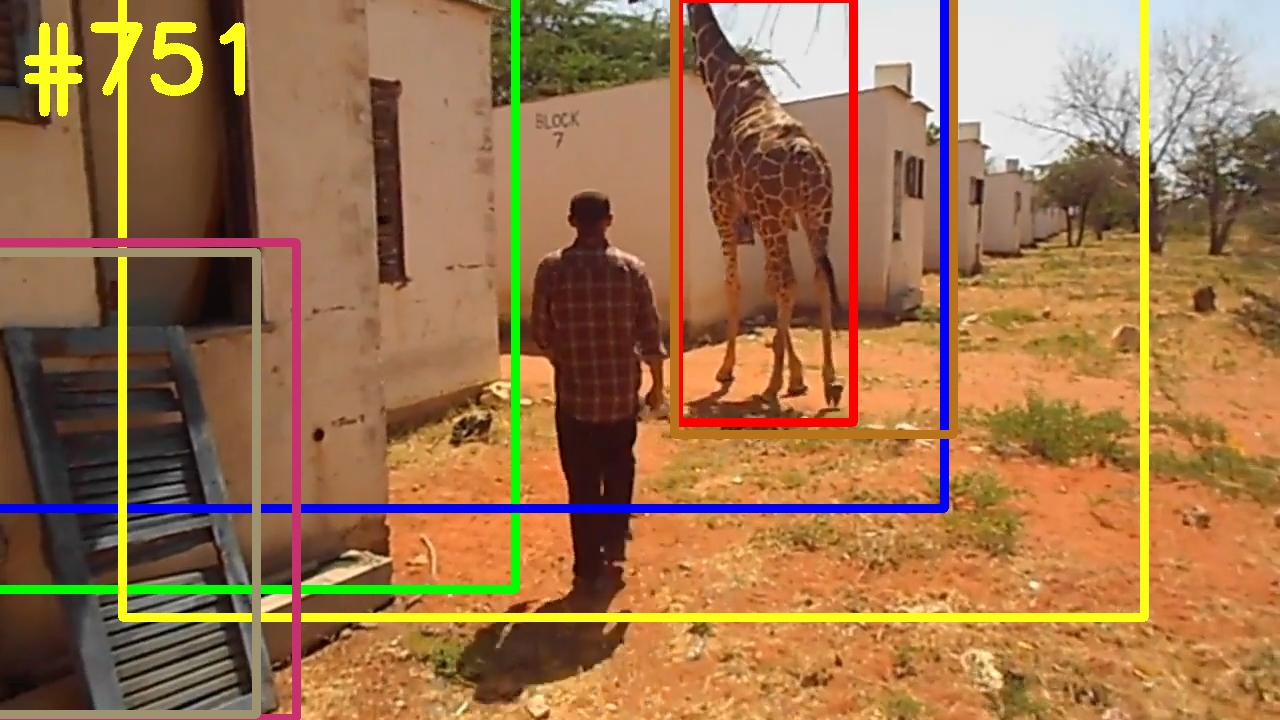}}\hspace{-0.027in}
\subfigure{\includegraphics[width=0.164\textwidth,height=1.635cm]{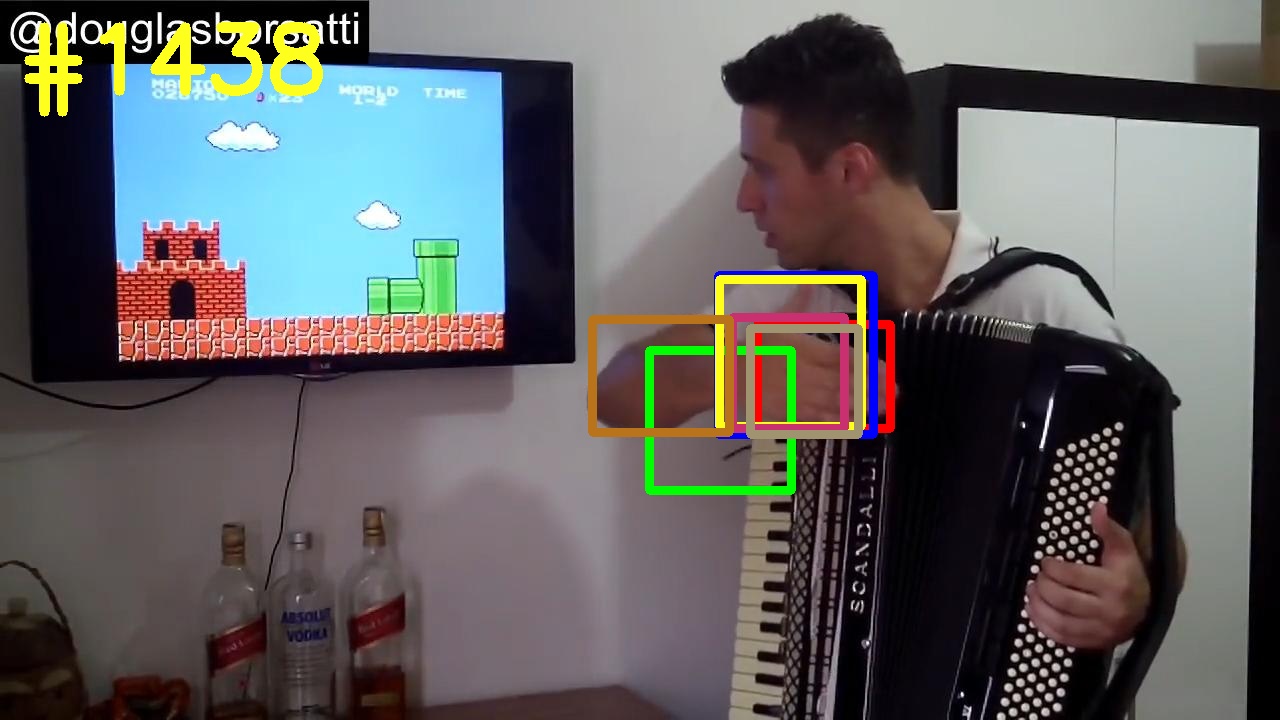}}\hspace{-0.027in}
\subfigure{\includegraphics[width=0.164\textwidth,height=1.635cm]{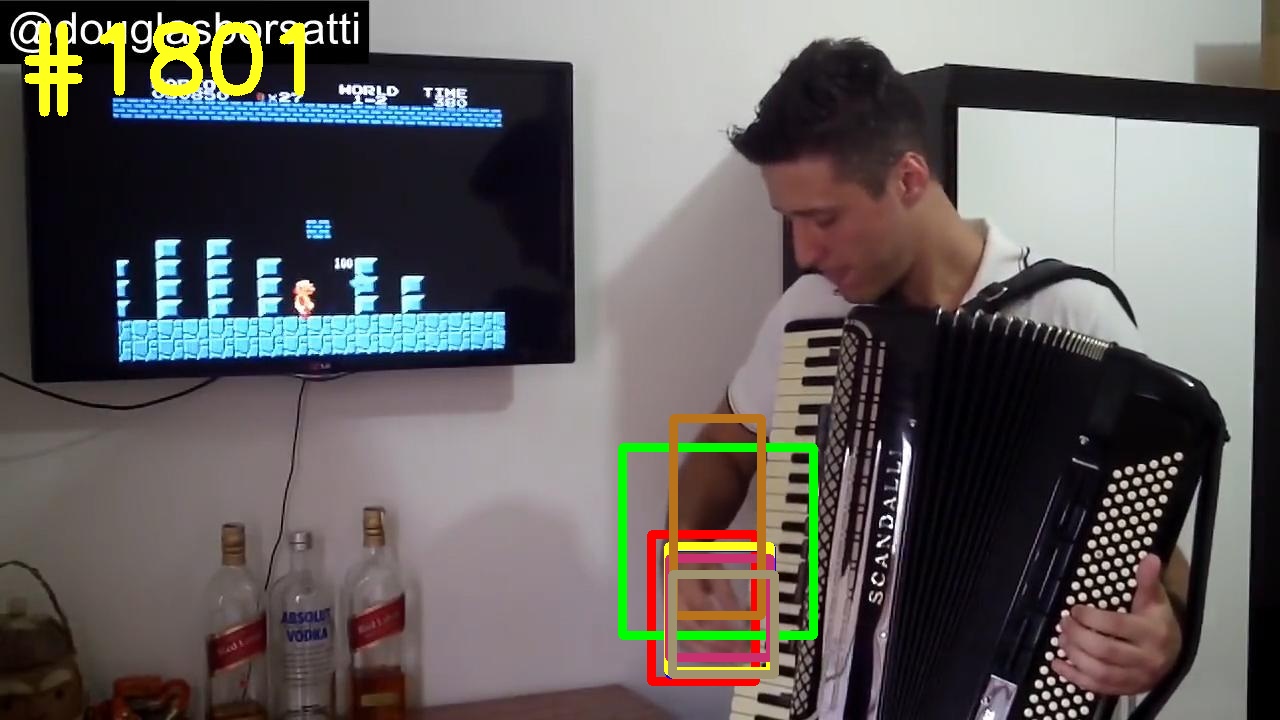}}\hspace{-0.027in}
\subfigure{\includegraphics[width=0.164\textwidth,height=1.635cm]{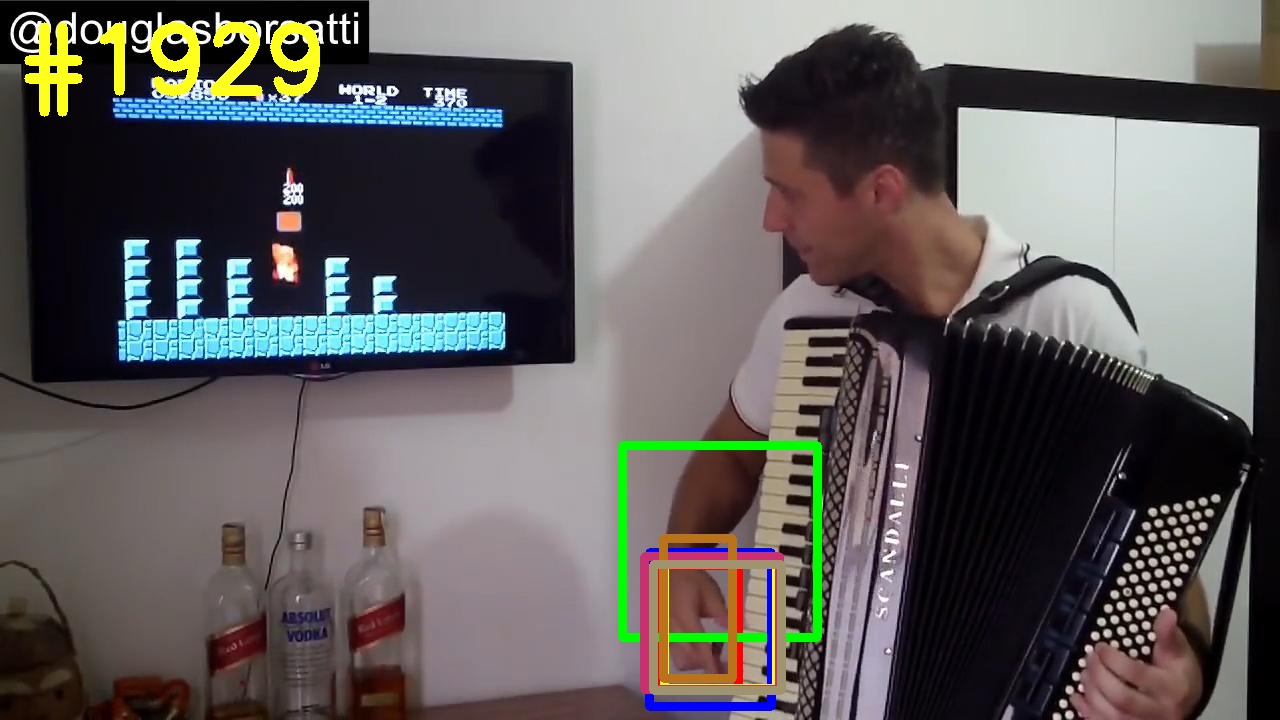}}\hspace{-0.027in}\vspace{-0.095in}
\subfigure{\includegraphics[width=0.9\textwidth,height=0.405cm]{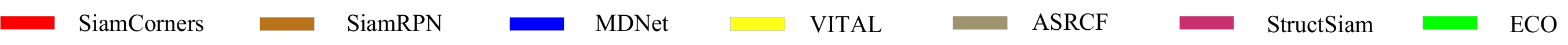}}
\caption{Visualization results of different methods on LaSOT dataset (from left to right and top to bottom: \emph{bicycle-9}, \emph{boat-4}, \emph{boat-17}, \emph{cat-3}, \emph{crocodile-10}, \emph{flag-5}, \emph{giraffe-10} and \emph{hand-16}).}
\label{fig_visual}
\end{figure*}

\textbf{OTB100 \cite{wu2015object}:} OTB100 is one of the most widely used benchmark datasets in tracking, which includes 100 fully labeled video sequences. The Area Under Curve (AUC) of success plot is employed to compare SiamCorners and nine state-of-the-art trackers including TADT \cite{li2019target}, GCT \cite{gao2019graph}, CIResNet-22-RPN \cite{zhang2019deeper}, RT-MDNet \cite{jung2018real}, SiamRPN \cite{li2018high}, DaSiamRPN \cite{zhu2018distractor}, FCAF \cite{han2019fully}, Ocean \cite{Ocean_2020}, ATOM \cite{danelljan2019atom} and DIMP \cite{bhat2019learning}. As shown in Fig. \ref{OTB}, our method performs better than the top tracker, such as ATOM, and slightly lower than its optimized version, DIMP. Our method outperforms the anchor-based trackers CIResNet22-RPN and DaSiamRPN with relative gains of 0.6\% and 1.3\% in terms of success score, respectively. Further, our method achieves a gain of 2.2\% in AUC over FCAF, while on par with offline anchor-free tracker Ocean. This shows that SiamCorners can achieve the state-of-the-art results by eliminating the anchor boxes.

\newcommand{\tabincell}[2]{\begin{tabular}{@{}#1@{}}#2\end{tabular}}
\begin{table*}[!t]
\centering
\caption{A comparison between the SiamCorners and state-of-the-art trackers on TrackingNet dataset in terms of AUC score, precision score (P) and normalized precision score (P$_{norm}$) }.

\fontsize{9pt}{9pt}\selectfont
\begin{tabular}{ccccccccccc}
\hline
\toprule  %Ìí¼Ó±í¸ñÍ•²¿´ÖÏß
 &\tabincell{c}{ECO \\ \cite{danelljan2017eco}}& \tabincell{c}{SiamFC \\ \cite{bertinetto2016fully}}&\tabincell{c}{CFNet \\ \cite{valmadre2017end}}& \tabincell{c}{MDNet \\ \cite{nam2016learning}}&\tabincell{c}{GFS-DCF \\ \cite{xu2019joint}}&\tabincell{c}{UPDT \\ \cite{bhat2018unveiling}}& \tabincell{c}{DaSiamRPN \\ \cite{zhu2018distractor}}&\tabincell{c}{UpdateNet \\ \cite{zhang2019learning}}&\tabincell{c}{ATOM \\ \cite{danelljan2019atom}}&\tabincell{c}{SiamCorners}\\
\midrule  %Ìí¼Ó±í¸ñÖÐºáÏß
AUC(\%)& 55.4  & 57.1  & 57.8  &60.6  &60.9  &61.1  &63.8  &67.7 &70.3  &69.5\\
P(\%)& 49.2& 53.3& 53.3&56.5  &56.6  &55.7&59.1 &62.5 &64.8&64.7\\
P$_{norm}$(\%)& 61.8&66.3& 65.4& 70.5 &71.8  &70.2&73.3 &75.2 &77.1&76.3\\
\bottomrule %Ìí¼Ó±í¸ñµ×²¿´ÖÏß
\end{tabular}
\label{trackingnet}
\end{table*}

\begin{table*}[!t]
\centering
\caption{A comparison between the SiamCorners and state-of-the-art trackers on NFS dataset.}
\fontsize{9pt}{9pt}\selectfont
\begin{tabular}{cccccccccccc}
\hline
\toprule  %Ìí¼Ó±í¸ñÍ•²¿´ÖÏß
&\tabincell{c}{BACF \\ \cite{kiani2017learning}}  &\tabincell{c}{SRDCF\\ \cite{lukevzivc2018discriminative}}  &\tabincell{c}{FCNT\\ \cite{wang2015visual}}  &\tabincell{c}{DaSiam-\\RPN \cite{Zhu_2018_ECCV}}  &\tabincell{c}{HDT\\ \cite{qi2016hedged}}  &\tabincell{c}{MDNet\\ \cite{nam2016learning}}  &\tabincell{c}{MKCF\\ \cite{tang2018high}}  &\tabincell{c}{ECO\\ \cite{danelljan2017eco}}  &\tabincell{c}{CCOT\\ \cite{danelljan2016beyond}} &\tabincell{c}{SiamRPN++\\ \cite{li2019siamrpn++}}  &SiamCorners\\
\midrule  %Ìí¼Ó±í¸ñÖÐºáÏß
AUC(\%)&34.2  & 35.3  & 39.3  &39.5  &40.0  &42.5  &45.5  &47.0  &49.2  &49.2  &53.7\\
FPS &35.3  & 13.0  &3.0  &160.0  &-  &1.0  &30.0  &8.0  &0.3 &35.0  &42.0\\

\bottomrule %Ìí¼Ó±í¸ñµ×²¿´ÖÏß
\end{tabular}
\label{NFS}
\end{table*}

\begin{figure}[!t]
\centering
\makeatletter\def\@captype{figure}\makeatother
\subfigure{
   \label{fig_Success_Plot}
   \includegraphics[width=4.1 cm]{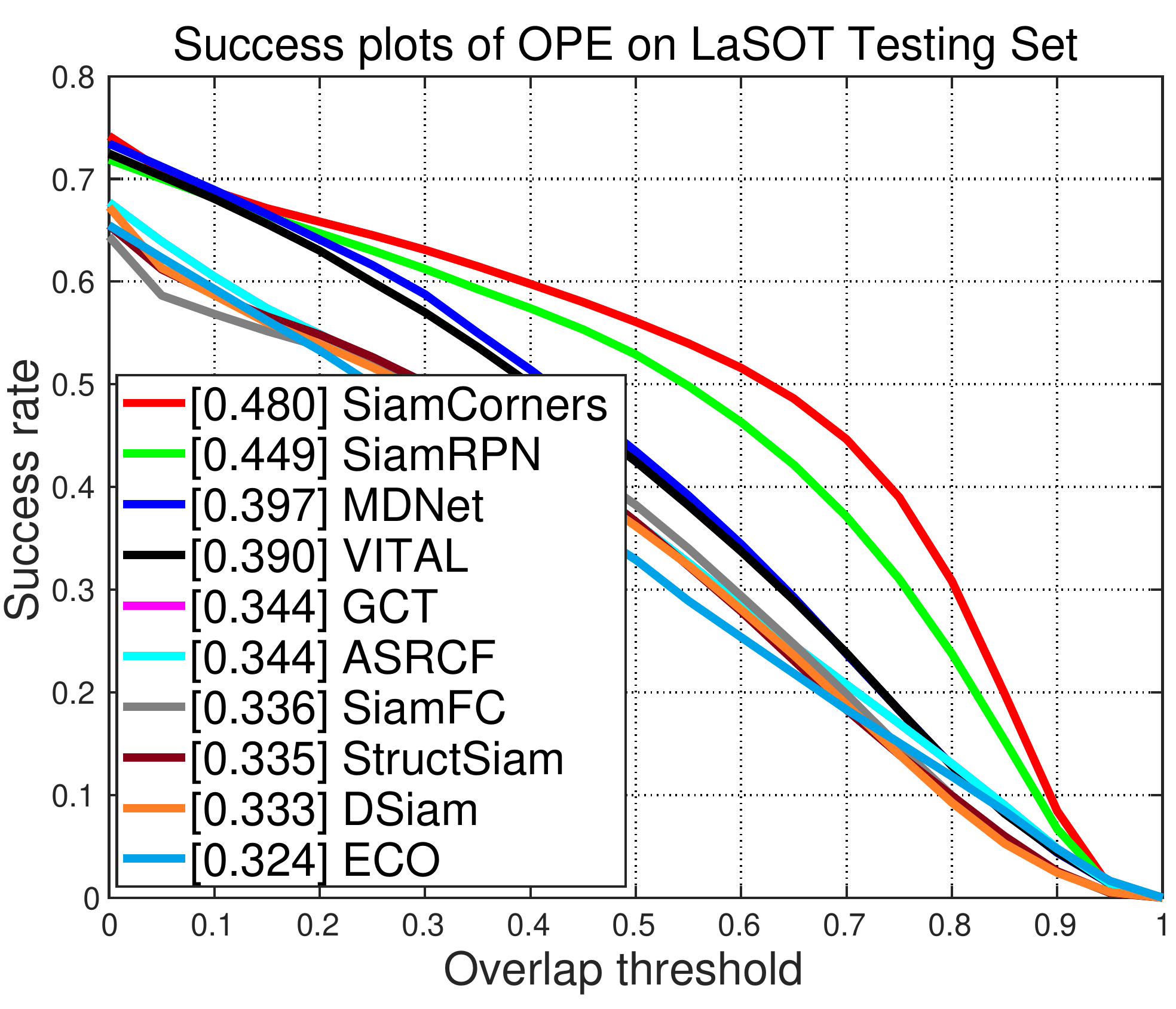}}
\hspace{0.005 in}
\subfigure{
   \label{fig_Precision_Plot}
   \includegraphics[width=4.1 cm]{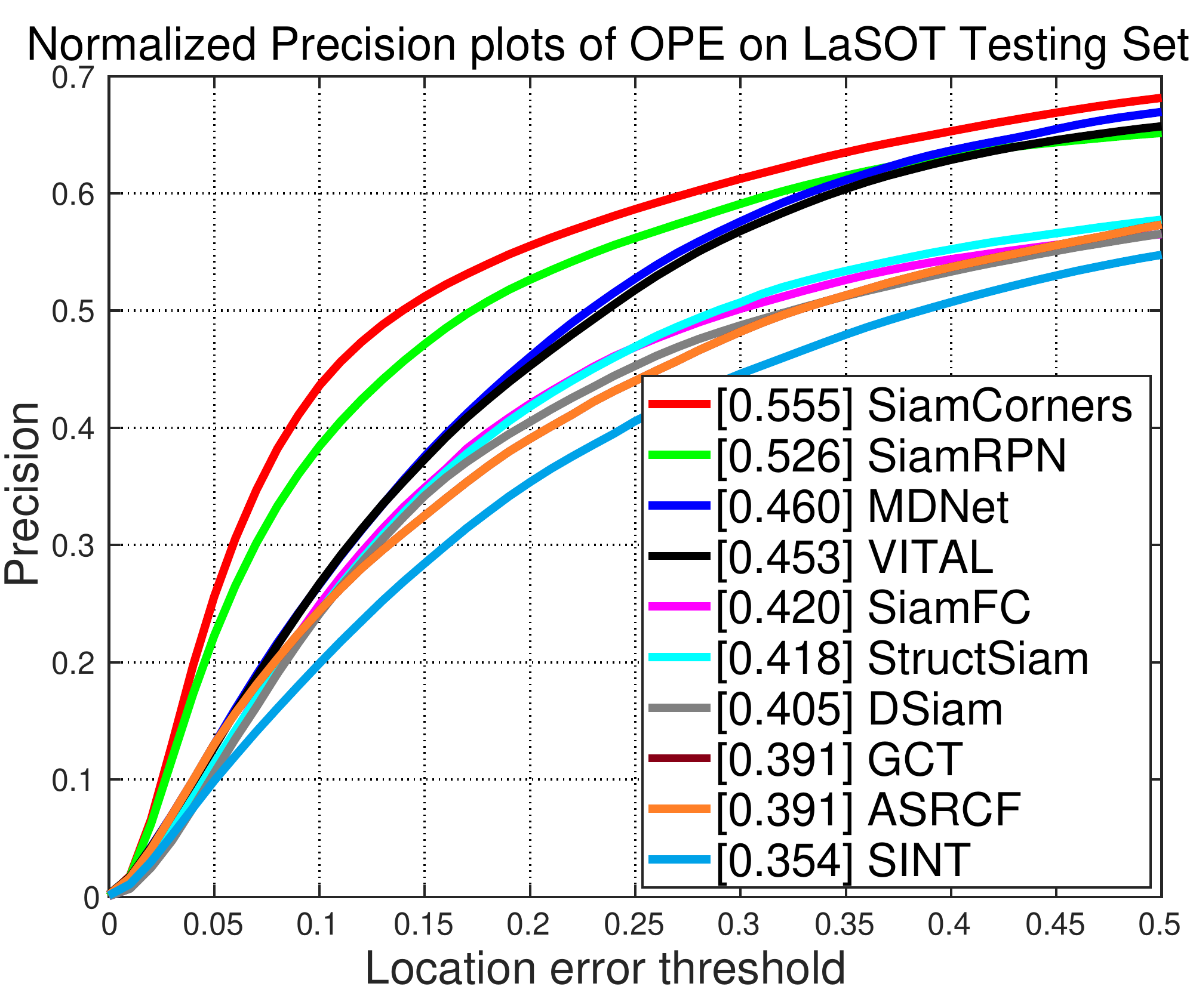}}
\caption{Experimental results of the methods on the LaSOT dataset.}
\label{LaSOT}
\end{figure}

\textbf{UAV123 \cite{mueller2016benchmark}:} We evaluate the proposed SiamCorners method on the UAV123 dataset in Fig. \ref{UAV}. The dataset is collected from a low-altitude UAV, which contains 123 video sequences and more than 110K frames in total. In this dataset,  we compare our method with recent trackers, \emph{i.e.} SiamRPN++ \cite{li2019siamrpn++}, DaSiamRPN \cite{zhu2018distractor}, SiamRPN \cite{li2018high}, RTMDNet \cite{jung2018real}, ECO \cite{danelljan2017eco}, GCT \cite{gao2019graph}, ASLA \cite{jia2012visual}, SAMF \cite{li2014scale}, MEEM \cite{zhang2014meem}, and DSST \cite{danelljan2014accurate}. In terms of 
the success score and precision score, SiamRPN++ achieves scores of 61.2\% and 80.6\%, respectively. Our approach outperforms the top tracker SiamRPN++ with relative gains of 0.2\% and 1.3\% in terms of the success score and precision score, respectively. It shows that our method achieves superior accuracy when compared with the recent methods based on anchor boxes.

In Fig. \ref{uav123}, we analyze the performance of the trackers under 12 challenging factors in tracking scenarios, such as lighting changes, scale variation, and aspect ratio change. These attributes are used to evaluate the performance of the trackers against different aspects. For clarity, the legend only shows the top 10 AUC scores in each plot. We observe that the proposed SiamCorners tracker handles well in scale variation, aspect ratio change, low resolution, fast motion, full occlusion, partial occlusion, out-of-view, viewpoint change, camera motion, and similar object. In particular, the proposed SiamCorners tracker has a 7.8\% higher AUC score than DaSiamRPN when tracking targets suffer from fast motion. These results show that the introduced SiamCorners method has the potential to solve the fast motion problem.

\textbf{TrackingNet \cite{muller2018trackingnet}:} TrackingNet is a recently proposed large-scale dataset in the wild, which provides more than 30K videos and 14 million bounding box annotations. We evaluate SiamCorners on its test split with 511 videos. Table \ref{trackingnet} shows the results of several state-of-the-art trackers, which are obtained by an online evaluation server. Among the compared approaches, ATOM achieves the best tracking result with an AUC score of 70.3\%. Our SiamCorners achieves an AUC score of 69.5\%, competitive with the other state-of-the-art trackers.

\textbf{LaSOT \cite{fan2019lasot}:} To further validate the effectiveness of the SiamCorners tracker, we evaluate our tracker on LaSOT \cite{fan2019lasot} dataset. LaSOT is a recently proposed tracking benchmark dataset, which contains 1400 sequences with an average video length of 2500 frames. We evaluate SiamCorners on its test split containing 280 videos. Furthermore, we compare our SiamCorners method with some representative trackers, that is,  SiamRPN \cite{li2018high}, MDNet \cite{nam2016learning}, VITAL \cite{song2018vital}, GCT \cite{gao2019graph}, ASRCF \cite{dai2019visual}, SiamFC \cite{bertinetto2016fully},
StructSiam \cite{zhang2018structured}, DSiam \cite{guo2017learning},  SINT \cite{tao2016siamese1}, and ECO \cite{danelljan2017eco}. As shown in Fig. \ref{LaSOT}, SiamCorners achieves the best performance in both success plot and normalized precision plot. Our approach outperforms MDNet with relative gains of 8.0\% and 9.3\% in terms of success score and normalized precision score, respectively.

Figure \ref{fig_visual} shows the visualization results of the trackers on the LaSOT dataset. Although these methods generally track targets accurately, they do not perform well under some challenging factors. The ECO and ASRCF based on correlation filters do not perform well in testing sequences with aspect ratio change (bicycle-9, cat-3, and giraffe-10). The MDNet and VITAL trackers are all built on classification component of CNN, which are prone to drift off the target when tracking targets undergo illumination variation (bicycle-9), partial occlusion (giraffe-10), and aspect ratio change (crocodile-10).  The SiamRPN and StructSiam trackers that follow the common Siamese architecture are less effective in addressing sequences with deformation (hand-16) and scale variation (boat-4, cat-3, and giraffe-10) attributes. In particular, the tracker SiamRPN based on anchor boxes cannot handle the scale variation problem well since it is difficult to handle the scale variation of the target by setting fixed ratio anchor boxes. Overall, the tracking results of the proposed SiamCorners performs well on most sequences. This qualitatively demonstrates the superiority of SiamCorners compared with other state-of-the-art methods.

\textbf{NFS30 \cite{kiani2017need}:} The NFS30 dataset contains 100 videos and a length of 380K frames in total, which are mainly captured by high frame rate cameras in real-world scenarios. We evaluate the SiamCorners tracker on the 30 FPS version of this dataset, which contains some challenging factors for fast-moving targets. From Table \ref{NFS}, we can observe that SiamCorners achieves $4.5\%$ higher than SiamRPN++ in terms of the AUC score. Furthermore, our method achieves a speed gain of 7 FPS compared with SiamRPN++, which verifies that the speed gain can be obtained by discarding anchors to reduce network parameters.

\subsection{Ablation Study}
\begin{table*}
\renewcommand\arraystretch{1.5}

	\centering
	\caption{Ablation study of the SiamCorners tracker on UAV123 and LaSOT.}
	\fontsize{9.7pt}{9.7pt}\selectfont

	\begin{tabular}{c|ccc|c|cc|cc|cc|c|c}
	    \hline
        Backbone & L3 & L4 & L5 & Offset & \tabincell{c}{MCP} & OCP & PPF &OPF &FRT &OUT &UAV123 & LaSOT \\
		
		\hline

		\cline{2-13}
		
		\multirow{6}{*}&\checkmark &  &  &\checkmark  &\checkmark  &&\checkmark&&\checkmark&&0.527  &0.380 \\
		
		~ &  &\checkmark  &  &\checkmark  &\checkmark  &&\checkmark&&\checkmark&&0.587 &0.432\\
		
		~ &  &  &\checkmark &\checkmark  &\checkmark   &&\checkmark&&\checkmark&&0.523  &0.373\\
		
		~ &\checkmark &\checkmark  &  &\checkmark  &\checkmark  &&\checkmark&&\checkmark&&0.556  &0.401\\
		
		~ &\checkmark &   &\checkmark &\checkmark  &\checkmark   &&\checkmark&&\checkmark&&0.539  &0.391\\
		
		~ {ResNet-50} &  &\checkmark &\checkmark &\checkmark  &\checkmark   &&\checkmark&&\checkmark& &0.601  &0.447\\
		
		\cline{2-13}
		
		\multirow{3}{*}{} &\checkmark &\checkmark   &\checkmark  &\checkmark  &  &\checkmark &\checkmark&&\checkmark& &0.593\  &0.455\\
		
		~ &\checkmark  &\checkmark &\checkmark &\checkmark  &\checkmark   &&&\checkmark&\checkmark& &0.600 &0.463\\
		
		~ &\checkmark  &\checkmark &\checkmark &\checkmark  & &\checkmark  &\checkmark&&&\checkmark &0.551  &0.411\\

		\cline{2-13}
		\multirow{2}{*}{} &\checkmark &\checkmark   &\checkmark  &\  &\checkmark  &&\checkmark&&\checkmark& &0.599\  &0.441\\
		
		~ &\checkmark  &\checkmark &\checkmark &\checkmark  &\checkmark   &&\checkmark&&\checkmark& &0.614  &0.480\\
		
		\hline
	\end{tabular}
\newline
\newline

\begin{tablenotes}

\item[*] Note: L3, L4, and L5 represent the \emph{conv3}, \emph{conv4}, and \emph{conv5} outputs of ResNet-50, respectively. Offset represents w-

hether to add an offset network to the convolutional network to predict the corner offsets. 
\end{tablenotes}

\label{ablation}
\end{table*}

To verify the effectiveness of each component in the proposed approach, we perform an ablation study on the UAV123 and LaSOT datasets in terms of AUC score of success plot. To be fair, we use the same training datasets and parameter settings in the training step for different components.

\textbf{Backbone Architecture.} Different convolutional layers of the backbone network produce different feature representations and parameters, which will affect the speed, memory usage, and performance of the tracker. As shown in Table \ref{ablation}, we empirically observe that \emph{con4} branch achieves the best AUC score in a single-level feature, while shallower or deeper layers obtain performance drops. Furthermore, the performance will be further improved by combining \emph{conv4} branch and \emph{conv5} branch, while the performance is not improved when combining the other two branches. Finally, SiamCorners achieves the best result that produces a 0.614 AUC score on UAV123 through aggregating all three layers, which is 2.7\% higher than the single-layer baseline.

\textbf{Offset.} We investigate the impact of the offset network by excluding it from our SiamCorners framework. As shown in Table \ref{ablation}, the proposed offset operation gains 1.5\% improvement on UAV123 and 3.9\% improvement on LaSOT, which demonstrates the importance of offset network. This happens because the introduced offset network can alleviate the impact of the network stride and thus improve the performance.

\textbf{Corner Pooling.} We investigate the impact of the original corner pooling (OCP) and the modified corner pooling (MCP) on the final tracking results. Note that the original corner pooling operation only performs corner prediction on a single feature map. We feed the feature maps of the top and left boundary images to the top-left corner pooling of OCP and the feature maps of the bottom and right boundary images to the bottom-right corner pooling of OCP in the ablation experiment. Table \ref{ablation} shows that introducing a MCP gains 2.1\% improvement on UAV123 and 2.5\% improvement on LaSOT. This shows that the MCP operation is a key component in the SiamCorners framework. We further investigate the proposed penalty function (PPF) and the original penalty function (OPF) on the tracking performance. Using a PPF to select the final tracking box yields 1.4\% improvement on UAV123 and 1.7\% improvement on LaSOT. Finally, we analyze the impact of the four respective templates (FRT) and one unified template (OUT) on the tracking performance. We use the OCP operation since OUT only generates a single feature map as input for corner pooling. Further, FRT obtains a 6.3\% improvement on UAV123 and a 6.9\% improvement on LaSOT. This suggests the FRT adds some performance improvements to our tracker.

\textbf{Component-wise Time.} We investigate the component-wise computation time of the SiamCorners tracker on the UAV123 dataset, shown in Table \ref{ablation2}. The last residual block \emph{conv5} of ResNet-50 takes the most computational time. Since the number of parameters on the feature extractor will be increased by using a deeper network, which increases the running time. It is worth noting that the calculation time of corner pooling ranks third. In the following research, we can design a more efficient pooling operation to reduce the running time.
\begin{table}[!t]

\centering
\caption{Component-wise computation time of the SiamCorners tracker on UAV123 dataset}

\fontsize{9pt}{9pt}\selectfont
\begin{tabular}{ccccccccc}
\hline
\toprule  %Ìí¼Ó±í¸ñÍ•²¿´ÖÏß
 Component
 &\tabincell{c}{C3\\}
 &\tabincell{c}{C4\\}
 &\tabincell{c}{C5\\}
 &\tabincell{c}{Off\\}
 &\tabincell{c}{Cor\\}
 &\tabincell{c}{Dec\\}
 &\tabincell{c}{Pos\\}\\
\midrule  %Ìí¼Ó±í¸ñÖÐºáÏß
Time (ms)
&\footnotesize 2.30
&\footnotesize 4.43
&\footnotesize 5.69
&\footnotesize 0.63
&\footnotesize 4.38
&\footnotesize 3.29
&\footnotesize 0.82 \\
\bottomrule %Ìí¼Ó±í¸ñµ×²¿´ÖÏß
\end{tabular}
\label{ablation2}
\newline
\newline
\footnotesize{\leftline{Note: C3, C4, and C5 represent  calculation time that the testing images are}}

\footnotesize{\leftline{fed to \emph{conv3}, \emph{conv4}, and \emph{conv5} of ResNet-50, respectively. Off and Cor rep-}}

\footnotesize{\leftline{resent the offset predictions of the corners and corner pooling operations, r-}}

\footnotesize{\leftline{espectively. Dec and Pos represent decoding operations and post-processing}}

\footnotesize{\leftline{operations that correspond to Section \ref{decoding1} and Section \ref{fusion}, respectively.}}

\end{table}

\section{Conclusions}
In this paper, we propose a novel tracking architecture, SiamCorners, which does not require multi-scale testing or pre-defined anchor boxes with different aspect ratios. SiamCorners directly tracks the targets as a pair of corners. Our architecture integrates layer-wise features in deep networks. By exploring a new penalty function, we can constrain the variation of center, width, and height of the bounding boxes in adjacent frames. We evaluate SiamCorners on OTB100, UAV123, NFS30, LaSOT, and TrackingNet datasets, which demonstrates its effectiveness and efficiency. We hope that our work will lead to further exploration for anchor-free methods in visual tracking.
\section*{Acknowledgment}

This research was supported in part by Special Research project on COVID-19 Prevention and Control of Guangdong Province (Grant No. 2020KZDZDX1227), in part by the Shenzhen Research Council (Grant No. JCYJ20170413104556946 and No. JCYJ20170413105929681), in part by the Natural Science Foundation of China under Grant No. 62006060, Grant No. U2013210,  Grant No. 62002241 and Grant No. 61972112, in part by the Guangdong Basic and Applied Basic Research Foundation under Grant No. 2021B1515020088. This work was also supported in part by funding (Grant No. 2019-INT021) from Shenzhen Institute of Artificial Intelligence and Robotics for Society (AIRS).

\bibliographystyle{IEEEtran}
\bibliography{IEEEabrv,mybibfile}
\begin{IEEEbiography}[{\includegraphics[width=1in,height=1.25in,clip,keepaspectratio]{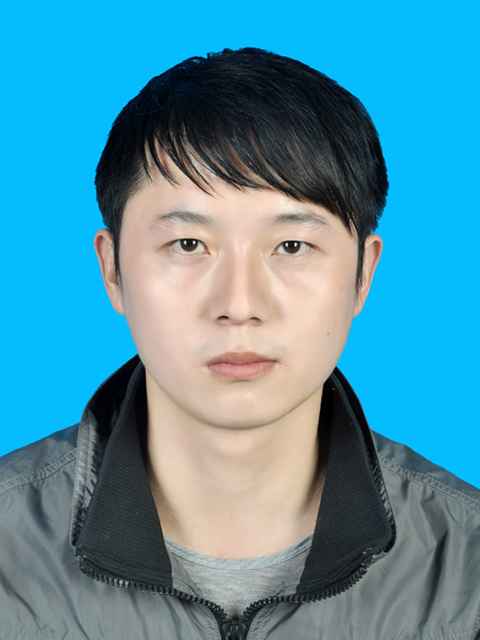}}]{Kai Yang}  is now pursuing the Ph.D. degree with
the Department of Computer Science and Technology, Harbin Institute of Technology, Shenzhen, China. His interests include deep learning, machine learning and visual tracking.
\end{IEEEbiography}

\begin{IEEEbiography}[{\includegraphics[width=1in,height=1.25in,clip,keepaspectratio]{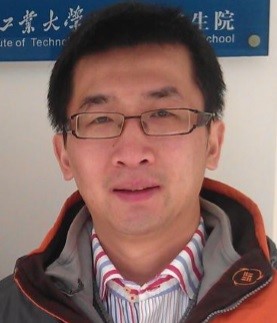}}]{Zhenyu He} received his Ph.D. degree from the Department of Computer Science, Hong Kong Baptist University, Hong Kong, in 2007. From 2007 to 2009, he worked as a postdoctoral researcher in the department of Computer Science and Engineering, Hong Kong University of Science and Technology. He is currently a full professor in the School of Computer Science and Technology, Harbin Institute of Technology, Shenzhen, China. His research interests include machine learning, computer vision, image processing and pattern recognition.
\end{IEEEbiography}

\begin{IEEEbiography}[{\includegraphics[width=1in,height=1.25in,clip,keepaspectratio]{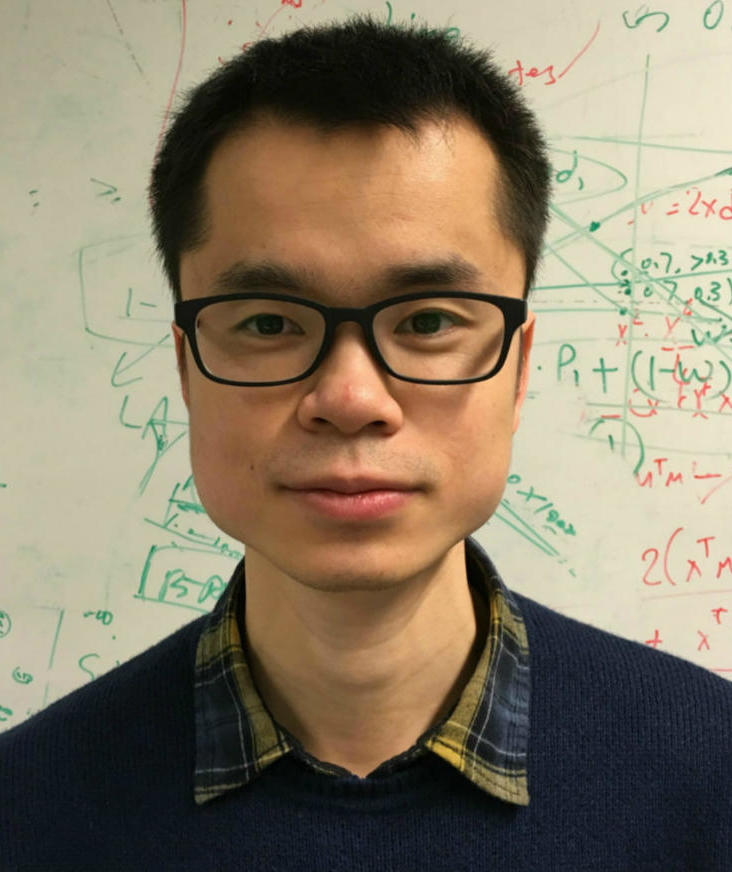}}]{Wenjie Pei}
is currently an Assistant Professor with the Harbin Institute of Technology, Shenzhen, China. He received the Ph.D. degree from the Delft University of Technology, the Netherlands in 2018, working with Dr. Laurens van der Maaten and Dr. David Tax. Before joining Harbin Institute of Technology, he was a Senior Researcher on Computer Vision at Tencent Youtu X-Lab. In 2016, he was a visiting scholar with the Carnegie Mellon University. His research interests lie in Computer Vision and Pattern Recognition including sequence modeling, deep learning, video/image captioning, etc.
\end{IEEEbiography} 

\begin{IEEEbiography}[{\includegraphics[width=1in,height=1.25in,clip,keepaspectratio]{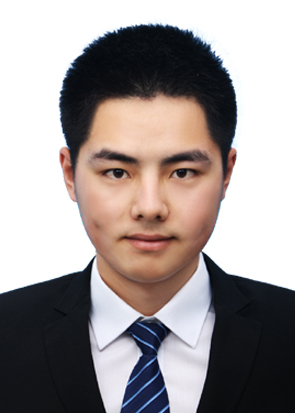}}]{Zikun Zhou}
received his Master degree from Harbin Institute of Technology in 2018. He is currently a Ph.D. candidate in school of Computer Science and Technology at Harbin Institute of Technology, Shenzhen. His research interests include computer vision and machine learning.
\end{IEEEbiography} 

\begin{IEEEbiography}[{\includegraphics[width=1in,height=1.25in,clip,keepaspectratio]{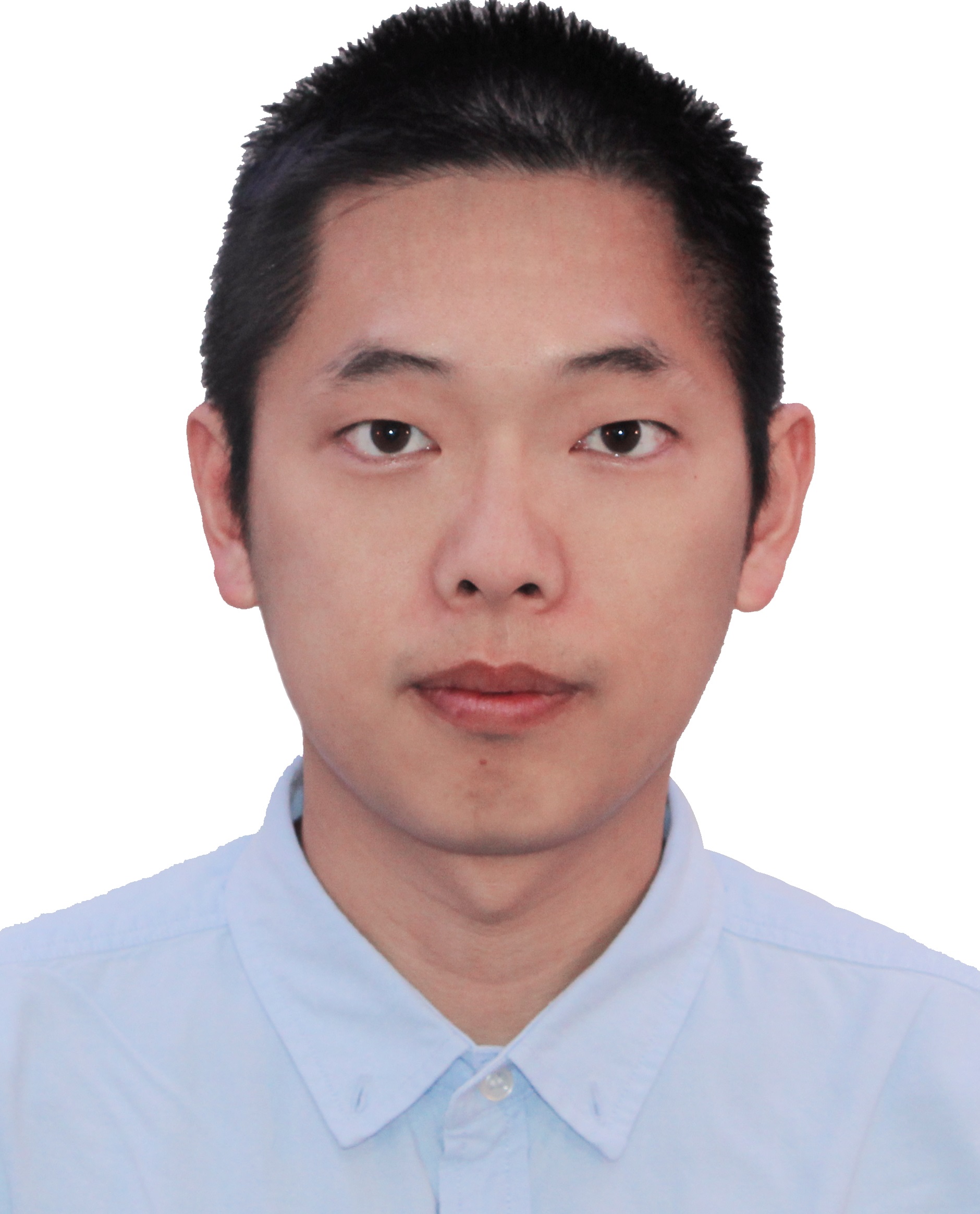}}]{Xin Li}
is a postdoc researcher with the AI center at the Peng Cheng Laboratory. He received the Ph.D. degree in computer applied technology from Harbin Institute of Technology, Shenzhen, China in 2020. His research interests include visual tracking, machine learning, and computer vision.
\vspace{-5mm}
\end{IEEEbiography} 

\begin{IEEEbiography}[{\includegraphics[width=1in,height=1.25in,clip,keepaspectratio]{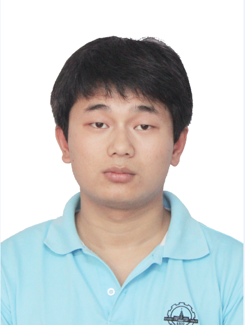}}]{Di Yuan}
received the M.S. degree in Applied Mathematics from Harbin Institute of Technology, Shenzhen, China, in 2017, where he is currently pursuing the Ph.D. degree in Computer Science and Technology. His current research interests include object tracking, image processing, and self-supervised learning.
\vspace{-5mm}
\end{IEEEbiography}

\begin{IEEEbiography}[{\includegraphics[width=1in,height=1.25in,clip,keepaspectratio]{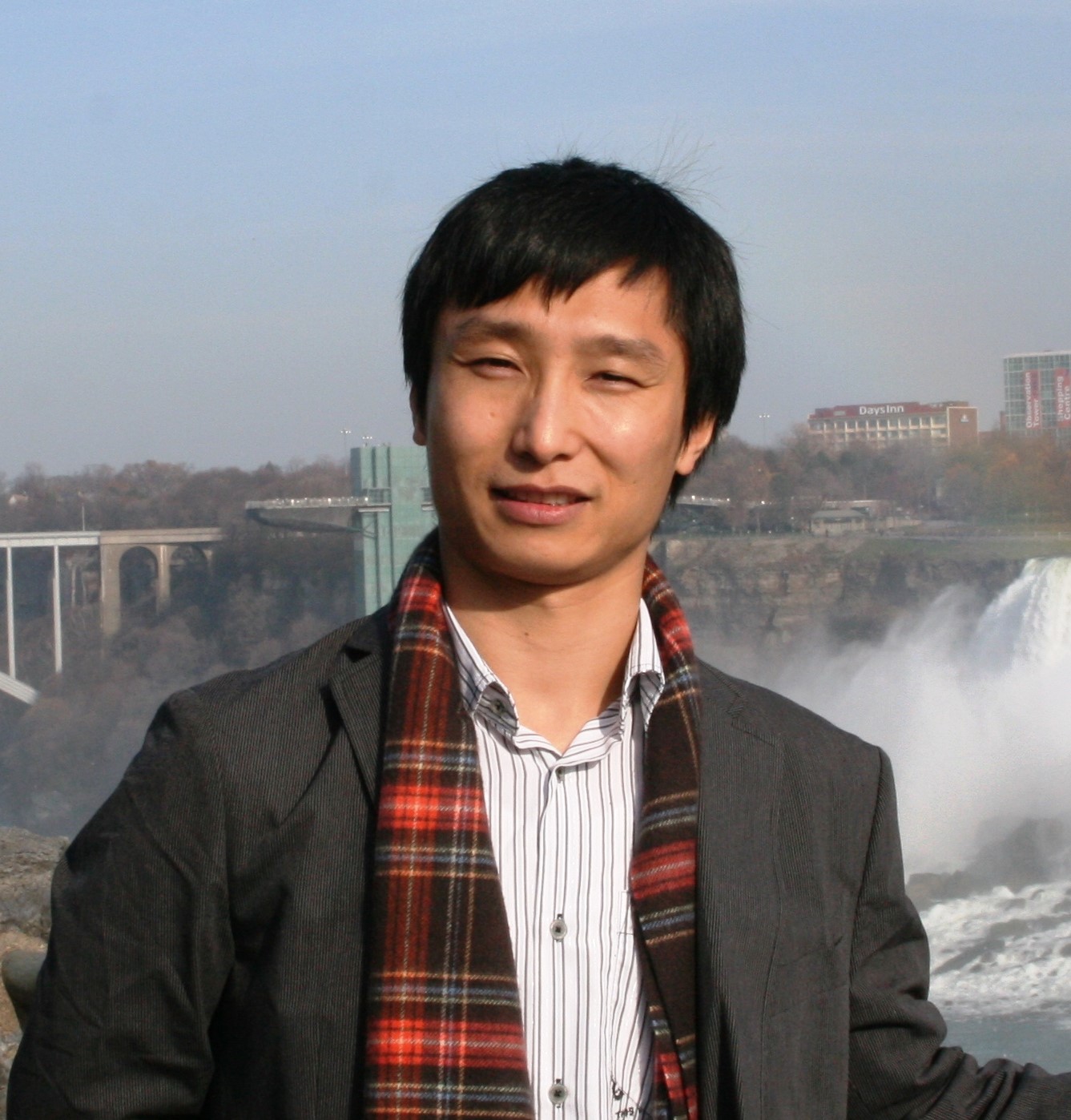}}]{Haijun Zhang (M'13)}
received the B.Eng. and Master's degrees from Northeastern University, Shenyang, China, and the Ph.D. degree from the Department of electronic Engineering, City University of Hong Kong, Hong Kong, in 2004, 2007, and 2010, respectively. He was a Post-Doctoral Research Fellow with the Department of Electrical and Computer Engineering, University of Windsor, Windsor, ON, Canada, from 2010 to 2011. Since 2012, he has been with the Shenzhen Graduate School, Harbin Institute of Technology, China, where he is currently a Professor of Computer Science. His current research interests include multimedia data mining, machine learning, computational advertising, and service computing. Prof. Zhang is currently an Associate Editor of Neurocomputing, Neural Computing and Applications, and Pattern Analysis and Applications.
\vspace{-5mm}
\end{IEEEbiography} 
\end{document}